\documentclass[pdflatex,sn-mathphys-num]{sn-jnl}

\usepackage{graphicx}%
\usepackage{multirow}%
\usepackage{amsmath,amssymb,amsfonts}%
\usepackage{amsthm}%
\usepackage{mathrsfs}%
\usepackage[title]{appendix}%
\usepackage{xcolor}%
\usepackage{textcomp}%
\usepackage{manyfoot}%
\usepackage{booktabs}%
\usepackage{algorithm}%
\usepackage{algorithmicx}%
\usepackage{algpseudocode}%
\usepackage{listings}%
\usepackage{times}
\usepackage{epsfig}
\usepackage{graphicx}
\usepackage{amsmath}
\usepackage{amssymb}
\usepackage{bm}
\usepackage{tabularx}
\usepackage{mathrsfs}
\usepackage{booktabs}
\usepackage{enumitem}
\usepackage{multirow}
\usepackage{array}
\usepackage{float}
\usepackage{stfloats}
\usepackage{caption}
\usepackage{graphicx} 
\usepackage{subcaption} 
\usepackage{hyperref}
\usepackage{cleveref}  
\usepackage{xcolor}

\theoremstyle{thmstyleone}%
\theoremstyle{thmstyletwo}%

\theoremstyle{thmstylethree}%

\begin{document}

\title[OpenEarthSensing: Large-Scale Fine-Grained Benchmark \\ for Open-World Remote Sensing]{OpenEarthSensing: Large-Scale Fine-Grained Benchmark \\ for Open-World Remote Sensing}


\author*[1,2]{\fnm{Xiang} \sur{Xiang}}\email{xex@hust.edu.cn}

\author[1]{\fnm{Zhuo} \sur{Xu}}

\author[1]{\fnm{Yao} \sur{Deng}}

\author[1]{\fnm{Qinhao} \sur{Zhou}}

\author[1]{\fnm{Yifan} \sur{Liang}} 

\author[2]{\fnm{Ke} \sur{Chen}}

\author[2]{\fnm{Qingfang} \sur{Zheng}}

\author[2]{\fnm{Yaowei} \sur{Wang}}

\author[3]{\fnm{Xilin} \sur{Chen}}

\author[2]{\fnm{Wen} \sur{Gao}}

\affil[1]{\orgname{Huazhong University of Science and Technology}, \orgaddress{\city{Wuhan}, \country{China}}}

\affil[2]{\orgname{Peng Cheng Laboratory}, \orgaddress{\city{Shenzhen}, \country{China}}}

\affil[3]{\orgname{Chinese Academy of Sciences}, \orgaddress{\city{Beijing}, \country{China}}}






\abstract{
The advancement of remote sensing, including satellite systems, facilitates the continuous acquisition of remote sensing imagery globally, introducing novel challenges for achieving open-world tasks. Deployed models need to continuously adjust to a constant influx of new data, which frequently exhibits diverse shifts from the data encountered during the training phase. To effectively handle the new data, models are required to detect semantic shifts, adapt to covariate shifts, and continuously update their parameters without forgetting learned knowledge, as has been considered in works on a variety of open-world tasks. However, existing studies are typically conducted within a single dataset to simulate realistic conditions, with a lack of large-scale benchmarks capable of evaluating multiple open-world tasks. In this paper, we introduce \textbf{OpenEarthSensing (OES)}, a large-scale fine-grained benchmark for open-world remote sensing. OES includes 189 scene and object categories, covering the vast majority of potential semantic shifts that may occur in the real world. Additionally, to provide a more comprehensive testbed for evaluating the generalization performance, OES encompasses five data domains with significant covariate shifts, including two RGB satellite domains, one RGB aerial domain, one multispectral RGB domain, and one infrared domain. We evaluate the baselines and existing methods for diverse tasks on OES, demonstrating that it serves as a meaningful and challenging benchmark for open-world remote sensing. The proposed dataset OES is available at \url{https://haiv-lab.github.io/OES}.}

\keywords{Open-world benchmark, Out-of-distribution detection, Incremental learning}



\maketitle

\section{Introduction}\label{sec1}

Remote sensing imagery provides a wealth of physical information about the real world. Interpreting these images can support various downstream applications, including disaster monitoring, resource management, and land use assessment \cite{rahnemoonfar2021floodnet}. Recent advances in deep learning have significantly improved remote sensing image interpretation. However, these methods are often trained and tested in constrained environments with fixed semantic categories, leading to challenges when deployed in open-world scenarios.
\begin{figure*}[t]  
\centering
\includegraphics[width=\textwidth]{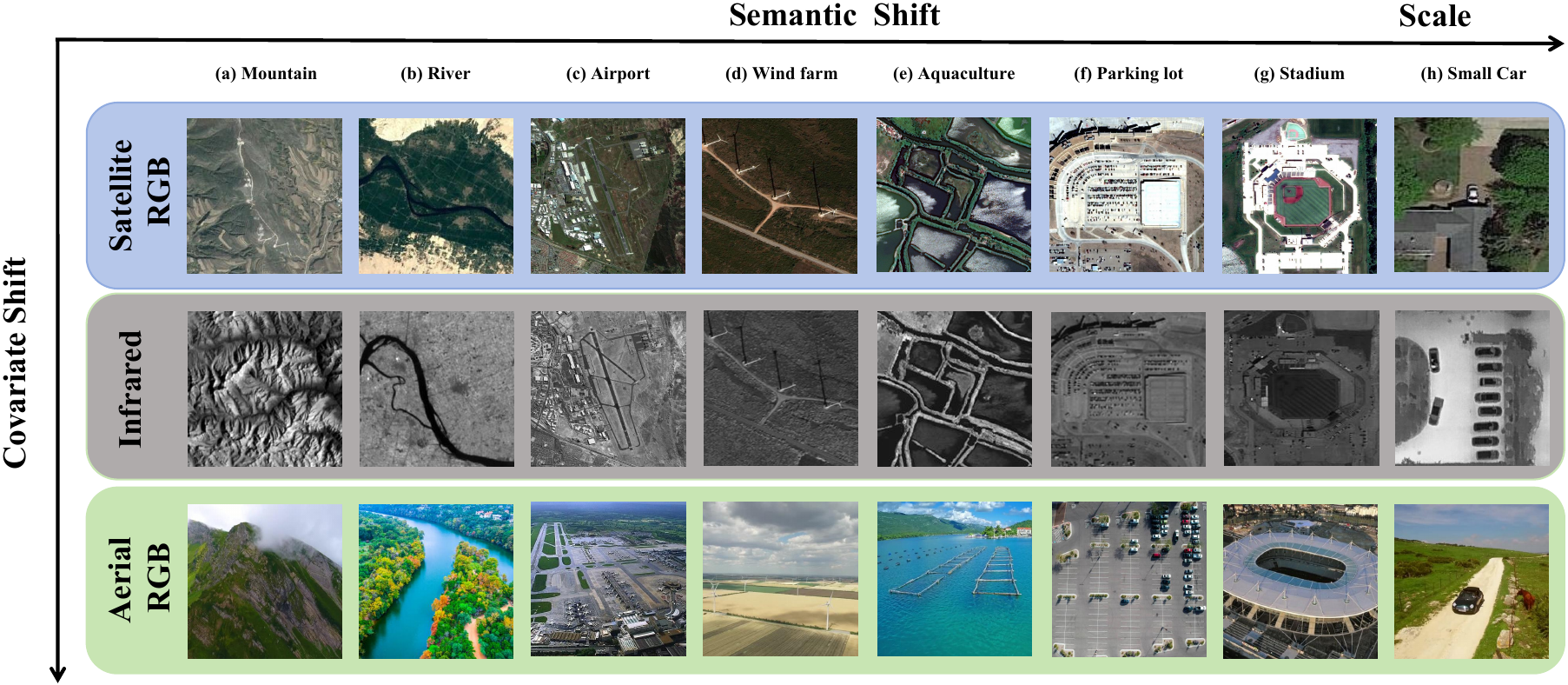}  
\caption{Data structure and examples in OpenEarthSensing, which incorporate common semantic shifts and covariate shifts in the open world. From the perspective of semantic shifts, OpenEarthSensing includes 189 remote sensing categories that encompass a variety of scales and contain diverse semantic information. From the perspective of covariate shifts, OpenEarthSensing includes five data domains with significant covariate shifts (three most representative data domains— satellite RGB, aerial RGB, and infrared images—are illustrated).}
\label{setting}
\end{figure*}
In the open-world scenarios, deployed models encounter distribution shifts during test time, encompassing both semantic shifts and covariate shifts. When faced with samples that exhibit semantic shifts, models must be able to effectively recognize unknown categories - a focus of research in open-set recognition (OSR) \cite{openmax16cvpr} and out-of-distribution (OOD) detection \cite{hendrycks17baseline}. Additionally, when presented with testing samples that display covariate shifts, models need to adapt to these changes. Related tasks include domain adaptation (DA) \cite{ben2010theory} and domain generalization (DG) \cite{li2018learning}. Furthermore, as new samples continuously emerge in the environments, models should continuously update - a challenge explored in incremental learning (IL) \cite{li2017learning}.

Existing works in open-world remote sensing often rely on single, simple classification datasets, which lack sufficient scale and diversity. These datasets typically show less variation within categories across different domains and follow an independent and identically distributed pattern during training and testing, failing to reflect real-world distribution differences caused by complex factors like shooting angles, geographical variations, and sensor types.
Furthermore, most of the latest methods demonstrate high accuracy on these datasets for open-world tasks, making their comparisons uninformative. Metadatasets \cite{Triantafillou2020Meta-Dataset}, which compile multiple datasets for greater scale and diversity, have been  created for remote sensing classification tasks \cite{roberts2023satin,dimitrovski2023current}. However, these metadatasets merely consist of a simple aggregation of multiple related subsets serving specific tasks, making them inadequate for the rigorous requirements of the open world. Therefore, establishing a more challenging, realistic, and large-scale benchmark for interpreting remote sensing imagery has become a critical priority in the field.

In this paper, we introduce \textbf{\textit{OpenEarthSensing (OES)}}, a large-scale, fine-grained benchmark for open-world remote sensing. OES features a metadataset that comprises five sub-datasets across \textbf{five domains} and \textbf{three modalities}, as illustrated in Fig.~\ref{setting}. It includes two RGB satellite imagery datasets, one RGB drone aerial imagery dataset, one multispectral RGB (MSRGB) dataset, and one infrared (IR) dataset. These five sub-datasets share the same categories but contain different covariate shifts, providing a more comprehensive testbed for evaluating the generalization ability of  models. In total, OES possesses \textbf{157,674 images} and comprises \textbf{10 \textit{coarse-grained} categories} with \textbf{189 \textit{fine-grained} categories}, containing scenes and objects with various ranges of scale.

Based on the five sub-datasets included in OES, we benchmark multiple mainstream open-world tasks and construct settings for each task that align with practical applications in remote sensing. To evaluate the adaptability of open-world models to semantic shifts, we choose \textbf{\textit{semantic shift OOD detection}} and \textbf{\textit{open-set recognition (OSR)}} as representative tasks. For assessing the generalization ability of open-world models to covariate shifts, we select \textbf{\textit{covariate shift OOD detection and generalization}} as key tasks. 
To evaluate the model's capacity for continuous updating, we benchmark \textbf{\textit{class-incremental learning (CIL)}} \cite{li2017learning}, \textbf{\textit{domain-incremental learning (DIL)}} \cite{wang2022s} and \textbf{\textit{coarse-to-fine few-shot class-incremental learning (C2FSCIL)}} \cite{xiang2022coarse} as representative tasks. Additionally, we evaluate the model's capabilities in \textbf{\textit{closed-set classification}} and \textbf{\textit{zero-shot classification}} on OES, which serve as the performance upper and lower bounds for the mentioned open-world tasks.
In each setting, we conduct performance evaluations of baselines as well as publicly available mainstream approaches, highlighting the significant challenges presented. To further advance the development of the  remote sensing field, we will make the proposed benchmark open source.
We summarize our contributions as follows.
\begin{itemize}
   \item 
        We introduce the OpenEarthSensing metadataset, a large-scale, fine-grained multi-modal dataset featuring 189 categories across five distinct domains and three modalities.
    \item   
        We benchmark essential visual tasks that are representative of open-world remote sensing and align with practice.
    \item 
        A comprehensive analysis of the experimental results contribute to both the research and development of open-world remote sensing.

\end{itemize} 
\label{sec:intro}

\section{Related Works}
\subsection{Remote Sensing Datasets}
Recently, EarthNets \cite{xiong2024earthnets} conducted a comprehensive review of over 500 publicly available remote sensing datasets. Among these, classification and detection datasets comprise the majority. This provides valuable support for the construction of open-world remote sensing datasets. For classification datasets, some early works focus on patch-level classification of satellite images, such as UCM Land Use \cite{yang2010bag} and BigEarthNet \cite{sumbul2019bigearthnet}. However, these datasets often face limitations in real-world applications due to constraints in scale and resolution. To address these challenges, researchers have introduced larger-scale and more diverse datasets, including NWPU-RESISC45 \cite{cheng2017remote}, fMoW \cite{christie2018functional}, RSD46-WHU \cite{xiao2017high}, millionAID \cite{long2021creating}, as well as detection-oriented benchmarks like DOTA \cite{xia2018dota} and FAIR1M \cite{sun2022fair1m}.  These datasets significantly expand the scope of remote sensing analysis, enabling more robust model training and evaluation. 
Beyond labeled datasets, several large-scale collections of globally distributed imagery—such as GeoLifeCLEF \cite{botella2023geolifeclef}, Satlas \cite{bastani2023satlaspretrain}, and RS5M \cite{zhang2024rs5m}—have emerged. While these datasets may lack fine-grained category annotations, their diversity, geographic coverage, and scalability make them invaluable for pretraining, self-supervised learning, and cross-domain adaptation studies. The growing availability of resources underscores the rapid evolution of remote sensing data infrastructure, paving the way for more generalized and adaptable remote sensing AI systems in Earth observation.

\begin{figure*}
\hsize=\textwidth 
\centering
\includegraphics[width=1.0\textwidth]{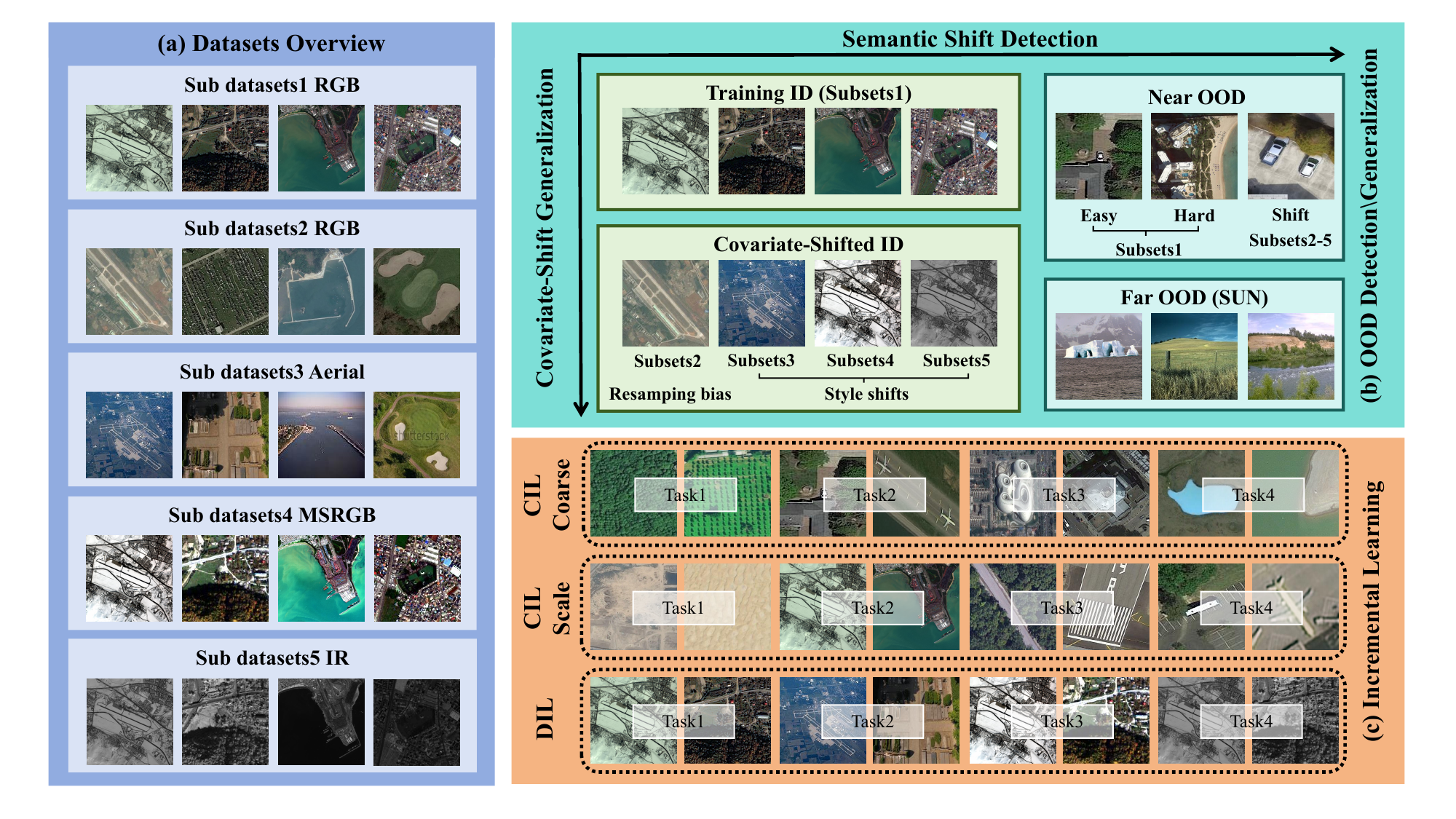}

\caption{Overview of the OpenEarthSensing datasets and the corresponding open-world tasks: (a) Includes five sub-datasets across five domains; (b) Evaluation protocol for OOD detection and generalization; (c) Evaluation protocol for incremental learning.
}

\label{pip}
\end{figure*}

\subsection{Remote Sensing Benchmark}
For natural images, a well-established ecosystem of open-world classification benchmarks has emerged to facilitate rigorous evaluation of algorithms, models, and systems. Notable examples include OpenOOD \cite{yang2022openood, zhang2024openood, yang2023full} for out-of-distribution detection, Dassl \cite{zhou2022domain} for domain adaptation studies, and PyCIL \cite{zhou2023pycil_new} for continual learning scenarios. These comprehensive benchmarks have significantly advanced methodological development in computer vision.  However, the remote sensing community has seen relatively limited progress in developing comparable open-world evaluation frameworks.  Current efforts include DPN-RS \cite{inkawhich2022improving} and \cite{gawlikowski2022advanced}, which investigate out-of-distribution detection using established datasets like AID \cite{xia2017aid}, UCM LandUse \cite{yang2010bag}, and MLRSNet \cite{qi2020mlrsnet}. For incremental learning tasks, CLRS \cite{li2020clrs} offers a 30-class remote sensing classification dataset, while SATIN \cite{roberts2023satin} provides a meta-dataset encompassing 27 satellite image datasets for vision-language model evaluation.
Existing remote sensing benchmarks still face key limitations: (1) their scale remains far smaller than that of natural image benchmarks, and (2) they focus on narrow tasks rather than open-world scenarios. This gap challenges the community to develop more expansive benchmarks for earth observation models.

\section{OpenEarthSensing Overview}

\renewcommand{\arraystretch}{1.2}

\renewcommand{\arraystretch}{1.0}

\subsection{Datasets Construction}

The OpenEarthSensing dataset is a large-scale, fine-grained open-world remote sensing image classification dataset, containing 157,674
images from 189 fine classes across 5 domains and 3 modalities. Each domain corresponds to a sub-dataset in Fig. \ref{pip} (a). All the data are filtered and retrieved from publicly available datasets and web data sources. Although many sources provide multi-spectral images with more than three channels, for the primary component of the OES dataset, we utilize three-channel visible light imagery rather than full-spectrum data. This decision is based on three key considerations:  (1) ensuring compatibility with pretrained three-channel models (e.g., Vision-Language Models), (2) maintaining consistency with established open-world methods that rely on standard RGB data (including out-of-distribution detection and incremental learning), and (3) addressing limitations in publicly available dataset, where existing 13-band collections either suffer from insufficient resolution (e.g., BigEarthNet \cite{sumbul2019bigearthnet}) or inconsistent spectral coverage (e.g., 4-8 bands in fMoW \cite{christie2018functional}).
 
\noindent\textbf{Merged and selected from public classification datasets.} In the compilation of OES, we first merge several publicly available remote sensing classification datasets, including WHU-RS19 \cite{xia2010structural}, NWPU-RESISC45 \cite{cheng2017remote}, RSD46-WHU \cite{xiao2017high}, AID \cite{xia2017aid}, MillionAID \cite{long2021creating}, RSI-CB256 \cite{li2020rsi}, BigEarthNet \cite{sumbul2019bigearthnet}, fMoW \cite{christie2018functional}, TreeSatAI \cite{ahlswede2022treesatai}, FGSC-23 \cite{2021FGSC23AL}, FGSCR-42 \cite{di2019public}, NaSC-TG2 \cite{zhou2021nasc},  MRSSC2.0 \cite{liu2022remote}, USTC SmokeRS \cite{ba2019smokenet}, MLRSNet \cite{qi2020mlrsnet}, UCM LandUse \cite{yang2010bag}, and RSI-CB128 \cite{li2020rsi}. We filter and merge the overlapping data with the same semantic categories to ensure the rationality of the included categories. 

\begin{algorithm}[h]
\caption{Image Cropping Algorithm}
\label{alg:image_cropping}
\begin{algorithmic}[1]
\State \textbf{Input:} Detection dataset images and XML files
\State \textbf{Output:} Cropped images by class

\For{$i = 1$ to $P$} \Comment{Total images}
    \State Inspect classes, count locations, store in list
    \For{each class}
        \State Take top-left target as $S_p$
        \State Compute distances from targets to $S_p$, sort
        \State Merge targets sequentially from $S_p$
        \If{No overlap with other classes}
            \State Merge into current target
        \Else
            \State Restart merging from current target
        \EndIf
    \EndFor
\EndFor
\State Save class bounding boxes, keep top 1000 by area
\end{algorithmic}
\end{algorithm}

\noindent\textbf{Cropped from public object detection datasets.} Although we have collected a substantial number of images across various categories from publicly available classification datasets, there remains a shortage of images specifically for object categories. Therefore, we also consider cropping bounding boxes from object detection datasets like FAIR1M \cite{sun2022fair1m} and VisDrone \cite{zhu2021detection} to generate object-level images. However, the images obtained by directly cropping the bounding boxes are often too tight and lack sufficient background information, leading to issues such as reduced category diversity and lower classification difficulty. Therefore, we create more diverse classification data using pixel expansion and merging similar objects. For each detection image, we statistically analyze the classes present and the locations of objects within each class. Starting from the top-leftmost object of each class, we perform merging operations to crop out suitable images as effectively as possible. The detailed pipeline is explained in Algorithm~\ref{alg:image_cropping} .

\noindent\textbf{Retrieval from web datasets.}
Most of the images collected from these public datasets are satellite images. To expand the diversity included in the dataset, we additionally retrieve semantically similar aerial images from large-scale web datasets, including RS5M \cite{zhang2024rs5m} and CC3M \cite{sharma2018conceptual}. We use the GeoRSCLIP \cite{zhang2024rs5m} to compute the visual-text similarity between candidate images and the label space of our collected data. For each category, we extract the top 100 most relevant images, which are then refined through a two-stage filtering process: (1) automated screening via a multi-modal large language model to eliminate low-quality or irrelevant samples, and (2) manual verification by human experts to ensure label consistency and visual fidelity. This curation mitigates noise from web data while preserving semantic alignment.

Through the above methods, we construct two satellite RGB datasets with overlapping semantic categories and domain shifts, along with one aerial dataset, one MSRGB dataset, and one infrared dataset. To meet the requirements of different evaluation tasks, we further categorize each subset into finer-grained partitions, where subscripts indicate modal/domain information(e.g. $R1$ for the first RGB domain) and superscripts specify task-related attributes. For example, $D_{R1}^{oode}$ refers to the 'Easy-OOD' split from the first RGB domain. All OES categories have been carefully reviewed by experts and designed to minimize overlap as much as possible. The statistics of the data sources, resolution, image sizes, and other information can be found in \emph{appendix}.
\begin{table*}[t]
\footnotesize
\centering
\caption{Statistical information for the different partitions of sub-datasets in OpenEarthSensing.}
\label{tab:data}  
\setlength{\tabcolsep}{1.4pt}
\begin{tabular}{c|ccccc|ccc|cccc|cccc|cccc}
    \hline
    \multirow{2}{*}{Datasets} & \multicolumn{5}{c|}{Sub Datasets1-RGB} & \multicolumn{3}{c|}{Sub Datasets2-RGB} & \multicolumn{4}{c|}{Sub Datasets3-Aerial} & \multicolumn{4}{c|}{Sub Datasets4-MSRGB} & \multicolumn{4}{c}{Sub Datasets5-IR} \\
    & $\mathcal{D}_{R1}^{all}$ & $\mathcal{D}_{R1}^{id}$ & $\mathcal{D}_{R1}^{oode}$ & $\mathcal{D}_{R1}^{oodh}$ & $\mathcal{D}_{R1}^{d}$ & 
    $\mathcal{D}_{R2}^{all}$ & $\mathcal{D}_{R2}^{id}$ & $\mathcal{D}_{R2}^{ood}$ & 
    $\mathcal{D}_{A}^{all}$ & $\mathcal{D}_{A}^{id}$ & $\mathcal{D}_{A}^{ood}$ & $\mathcal{D}_{A}^{d}$ & 
    $\mathcal{D}_{M}^{all}$ & $\mathcal{D}_{M}^{id}$ & $\mathcal{D}_{M}^{ood}$ & $\mathcal{D}_{M}^{d}$ & 
    $\mathcal{D}_{I}^{all}$ & $\mathcal{D}_{I}^{id}$ & $\mathcal{D}_{I}^{ood}$ & $\mathcal{D}_{I}^{d}$ \\
    \hline
    Class & 189 & 94 & 48 & 47 & 50 & 65 & 43 & 22 & 137 & 71 & 66 & 50 & 56 & 34 & 22 & 50 & 62 & 36 & 26 & 50 \\
    Images & 75707 & 40291 & 15962 & 18454 & 21053 & 26277 & 16699 & 9578 & 11037 & 5553 & 5484 & 3789 & 22153 & 14960 & 7193 & 20121 & 23374 & 15444 & 7930 & 20025 \\
    \hline
\end{tabular}
\end{table*}

\subsection{Dataset Analysis}
\begin{figure}[h!]

\begin{subfigure}{0.48\linewidth}
\centering
 \hspace{-3mm} 
\includegraphics[scale=0.30]{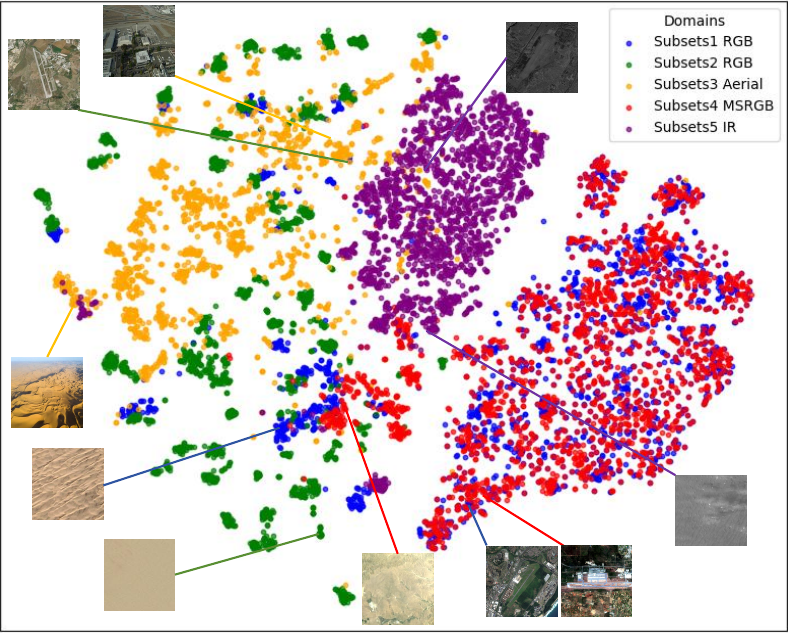}
\caption{t-SNE across 5 domains}
\label{fig:5domains}
\end{subfigure}
\begin{subfigure}{0.48\linewidth}
\centering
\includegraphics[scale=0.30]{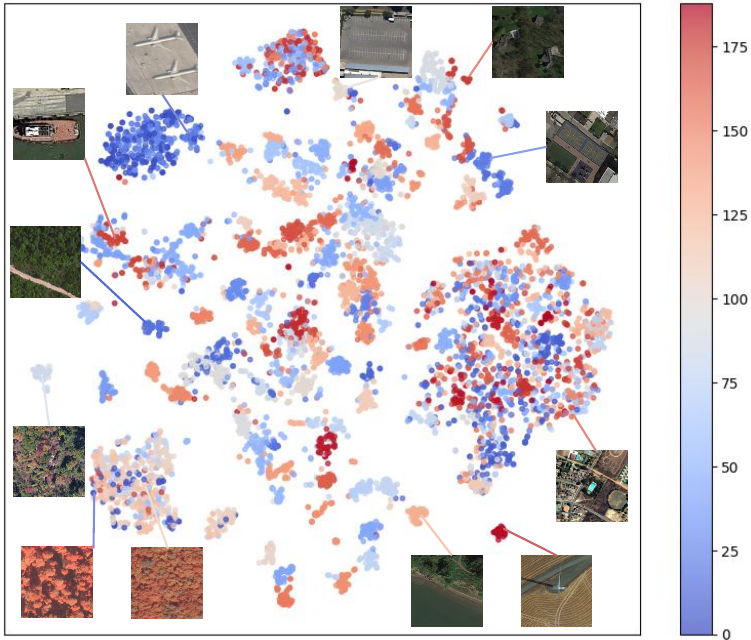}
\caption{t-SNE on domain 1}
\label{fig:domains}
\end{subfigure}


\caption{t-SNE visualization on OES.}
\label{fig:visualization}

\end{figure}

Compared to existing open-world remote sensing datasets, OES exhibits the following characteristics:

\noindent\textbf{Multiple and diverse domains.} OES comprises five sub-datasets with five distinct domains, enabling it to serve as a testbed for various generalization tasks. We randomly select 2,000 images from each domain and utilize GeoRSCLIP \cite{zhang2024rs5m} to extract features. The t-SNE visualization is presented in Fig.~\ref{fig:5domains}, with each color representing a different domain. Notably, even though sub-dataset 1 and 2 both originate from satellite imagery, there is a significant domain shift due to the varying capturing conditions. Furthermore, satellite, aerial, and infrared images display considerable differences as well. These domain shifts highlight the significant evaluation value and challenges in OES.

\noindent\textbf{Wide span of scales.} The evolution of remote sensing has led to a progressive enhancement in the resolution of imagery. Consequently, the demands in recognition have expanded beyond mere scene classification to encompass the identification of objects at finer scales. To accommodate the scale variations present in remote sensing images, all included categories in OES exhibit significant scale variations, ranging from broader scenes (e.g., construction site and wind farm) to specific objects (e.g., steel smelter and wind turbine). Among the 189 categories, there are 152 scenes and 37 objects. We deliberately separate scenes and objects, even though objects may be part of scenes (e.g., wind farms and wind turbines), to evaluate the model's robustness in distinguishing between different scales. We aim for the model to recognize not only large-scale scenes but also fine-grained, smaller-scale objects when the camera focuses on distinctive targets.

\begin{figure}[h!]

\begin{subfigure}{0.49\linewidth}
\centering
\includegraphics[scale=0.31]{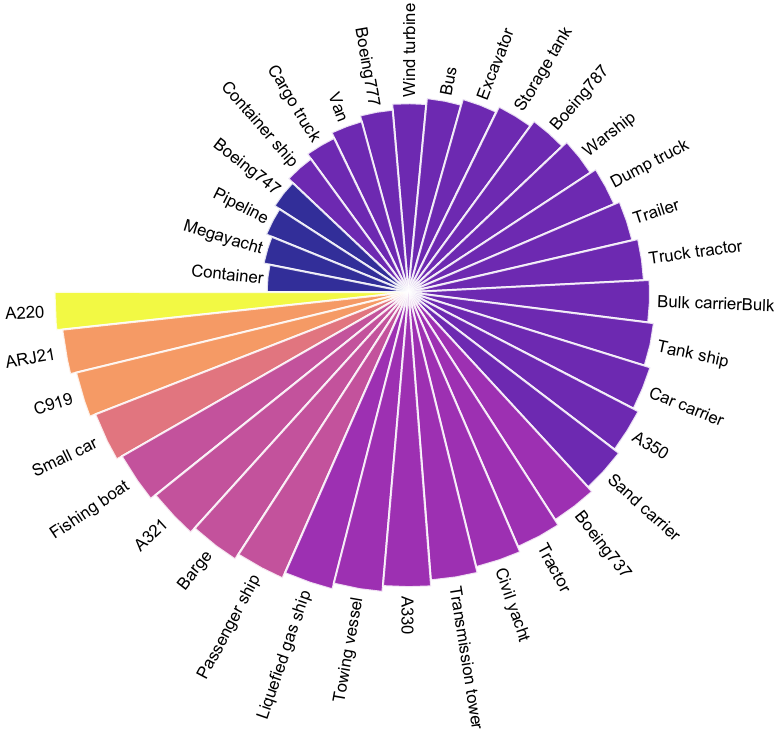}
\caption{Objects scale scores}
\label{fig:scale a}
\end{subfigure}
\begin{subfigure}{0.49\linewidth}
\centering
\includegraphics[scale=0.31]{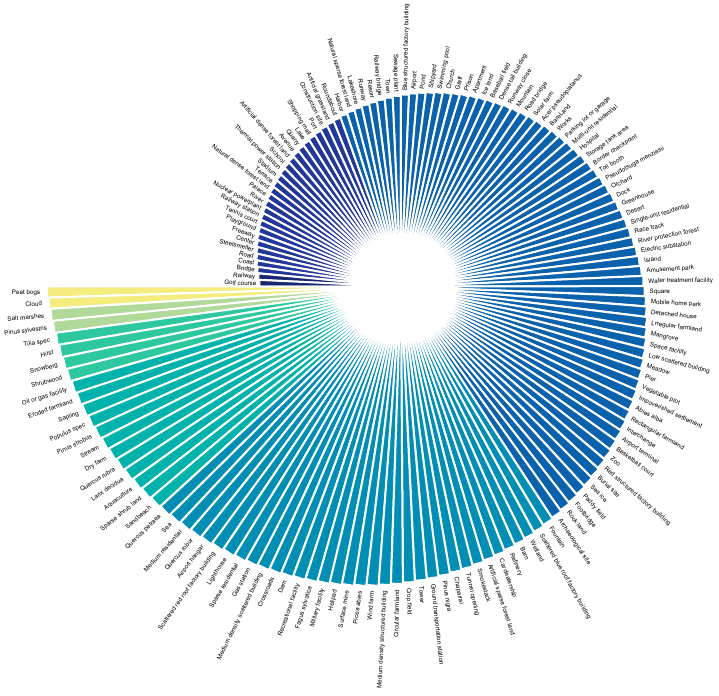}
\caption{Scenes scale scores}
\label{fig:scale b}
\end{subfigure}

\caption{Scale scores on different categories in OES.}

\label{fig:scale}

\end{figure}

To delve deeper into the intricacies of scale diversity within the OES dataset, \emph{Qwen-VL-chat} \cite{bai2023qwen} is employed to evaluate the image scales associated with both scene and object categories. We use multiple instructions to generate different results, such as: "This is a remote sensing image of [classname]. Please rate its scale on a 1-10 score range, where 10 represents large-format scenes and 1 indicates targets with smaller physical coverage." The distribution of OES across different scales is visually represented in Fig.~\ref{fig:scale}. The extensive spectrum of scale variations within the OES dataset introduces a novel challenge to the realm of remote sensing recognition.

\noindent\textbf{Multiple coarse categories.} OES comprises 10 coarse-grained categories, which cover the majority of scenarios encountered in remote sensing applications. Each coarse-grained category is further divided into 10 to 27 fine-grained subcategories, culminating in a total of 189 distinct classifications. For instance, the \textit{Infrastructure} coarse-grained category encompasses 26 fine subcategories, including but not limited to \textit{church} and \textit{palace}. In Fig.~\ref{fig:domains}, we visualize the feature distribution of certain categories within domain 1. The comprehensive information regarding all coarse-grained and fine-grained categories included in OES is available in the \emph{Appendix}.

Tab.~\ref{compare} presents a comparison of OES with other existing datasets used in open-world tasks, focusing on resolution, number of categories, and other characteristics. Compared to other datasets, OES offers a wider range of resolutions, a broader and more fine-grained set of categories, as well as supports for more modalities and domains, enabling it to facilitate diverse open-world tasks.

\begin{table*}[t]
    \centering
    \caption{Comparison of OpenEarthSensing with other datasets used in remote sensing open-world tasks.}
    \setlength{\tabcolsep}{4pt} 
    \begin{tabular}{cccccccccc} 
        \hline
        Datasets &Resolution & Classes & Scenes & Objects & Fine-grained
  & Modals & Domains & Hierarchy   \\  
        \hline
        EuroSAT \cite{helber2019eurosat}&10m &10 &10 &0 &$\times$ &\checkmark  &$\times$ &$\times$\\
        AID \cite{xia2017aid} &3m &30 & 30 & 0 &  \checkmark  &$\times$ & $\times$ &$\times$ \\
         NWPU-RESISC45 \cite{cheng2017remote} & 0.2-30m & 45 &43 & 2 & \checkmark & $\times$ &$\times$ & $\times$\\  
        UCMLandUse \cite{yang2010bag} & 0.3m & 21 & 20 & 1 & \checkmark & $\times$ &$\times$ & $\times$\\
        CLRS \cite{li2020clrs}&0.26-8.85m&47&47&0&\checkmark &$\times$ &$\times$ & \checkmark \\
        
         OES (Ours) &0.06-153m & 189 & 157 & 32& \checkmark & \checkmark & \checkmark & \checkmark \\

        \hline
    \end{tabular}
    \label{compare}
\end{table*}

 \subsection{Benchmarking Open-World Tasks}

We first benchmark the zero-shot and closed-set classification tasks on OES, representing the lower and upper bounds of model performance in open environments respectively. Specifically, we utilize the $\mathcal{D}_{R1}^{all}$, $\mathcal{D}_{A}^{all}$, $\mathcal{D}_{M}^{all}$, and $\mathcal{D}_{I}^{all}$ sub-datasets to assess classification capabilities across satellite RGB images, aerial RGB images, MSRGB images, and infrared images, respectively.
Then, we benchmark open-world tasks from the perspectives of adapting to covariate shifts, detecting semantic shifts, and incrementally learning from new shifted categories and domains rapidly. Unlike conventional benchmarks, OES establishes a unified benchmark for open-world remote sensing tasks, addressing two critical limitations: (1) eliminating redundant protocol development across small datasets for different tasks, and (2) overcoming the performance saturation in current methods.

\noindent\textbf{Semantic Shift OOD Detection \& OSR}. Recent work \cite{wang2024dissecting} highlights a strong correlation between OOD detection and OSR in both settings and performance. Both tasks detect new categories with shifted semantics, while OSR also requires maintaining in-distribution (ID) accuracy. We unify these tasks to evaluate a model's ability to handle semantic shifts. Unlike existing remote sensing benchmarks that randomly split ID and OOD samples, we consider the semantic shift degree between coarse and fine classes, aligning our setup with real-world deployment scenarios.
As shown in Tab.~\ref{tab:data}, for the 189 classes of satellite RGB image data, we designate 94 classes as ID samples $\mathcal{D}{R1}^{id}$; the remaining 95 classes are considered OOD samples with semantic shifts, with 48 classes categorized as OOD-Easy split $\mathcal{D}{R1}^{oode}$ and 47 classes as OOD-Hard split $\mathcal{D}_{R1}^{oodh}$. The OOD-Easy exhibits a significant semantic shift from ID, while the OOD-Hard shows a smaller semantic shift from ID.

\noindent\textbf{Covariate Shift OOD Detection \& Generalization.}
Covariate shift OOD detection emphasizes robustness to covariate shifts, also referred to as full-spectrum OOD detection \cite{yang2023full}, where the ID data remain semantically consistent, while covariates vary. Given the practical needs of remote sensing, we focus on the following shifts: (1) Resampling bias, requiring model generalization across varying acquisition parameters (angle, height, resolution, time) within the same modality; (2) Modal shift, demanding generalization across different modalities (satellites, aerial images) for the same semantic categories.
We utilize 94 classes of satellite RGB training data from $\mathcal{D}_{R1}^{id}$.  During testing, both ID and OOD data are drawn from the remaining sub-datasets, with ID/OOD labels assigned based on semantic alignment with the training categories.  For instance, in resampling bias testing, the test set of $\mathcal{D}_{R2}^{id}$ serves as ID data, while $\mathcal{D}_{R2}^{ood}$ is used as OOD data. This setup simulates real-world scenarios where models encounter covariate shifts.


\noindent\textbf{Class-Incremental Learning (CIL)}.
With the ever-evolving landscape of remote sensing technologies, copious amounts of high-quality images are captured daily across various scales and locations worldwide. Continual training of models is essential to incorporate and leverage this influx of data, enabling the recognition of novel classes in the open-world setting. 
However, existing deep learning methods often encounter a phenomenon known as catastrophic forgetting during Class-Incremental Learning, where the model progressively loses its ability to accurately recognize previously encountered classes.
Although there have been some CIL benchmarks in remote sensing, they suffer from the following problems: (1) limited category diversity, hindering the emulation of intricate real-world settings; (2) the scope of coarse-grained categories is constrained, particularly given the prevalence of specialized satellites dedicated to capturing data within specific coarse categories, which results in a lack of consideration for the continual processing of diverse coarse-grained categories; (3) prevalent uniformity in data scale, a departure from the diverse scales encountered in actual remote sensing operations influenced by factors like satellite orbits.

To address these limitations, we evaluate existing CIL methods using three benchmarks:
\textbf{Random}, \textbf{Coarse}, and \textbf{Scale}, utilizing $\mathcal{D}_{R1}^{all}$, which contains RGB images with 189 classes. In \textbf{Random}, we follow the widely-used CIL setting and randomly assign classes to 10 sessions equally. In \textbf{Coarse}, we set each session to contain fine classes of one coarse category to simulate the continuous learning from data captured by different types of dedicated satellites by the model. We divide all the classes into 10 coarse categories corresponding to 10 sessions, see the appendix for the division. In \textbf{Scale}, we aim to replicate the continual process from large to small scales. To establish the setting for scale transformation, we initially differentiate 37 small-scale objects from 152 relatively large-scale scenes manually. Subsequently, the scales of the object and scene categories are individually evaluated using the multimodal large model, leading to the scale distributions depicted in Fig.~\ref{fig:scale}. The 10 sessions are composed of evenly distributed categories based on a progression from large to small scales.

\noindent\textbf{Domain-Incremental Learning (DIL)}.
To evaluate models' adaptability to cross-domain data, we benchmark Domain-Incremental Learning tasks on OES. We select 50 categories containing the same semantic classes from RGB satellite, RGB aerial, MSRGB, and IR images, denoted as $\mathcal{D}_{R1}^{d}$, $\mathcal{D}_{A}^{d}$, $\mathcal{D}_{M}^{d}$ and $\mathcal{D}_{I}^{d}$, as shown in Tab.~\ref{tab:data}. In each task, models are trained on images from only one domain, while being evaluated across all previously learned domains during testing.



\noindent\textbf{Coarse-to-Fine Few-shot Class-Incremental Learning (C2FSCIL)}. 
In this task, we provide models with all training samples accompanied by coarse labels in the base session, including 10 coarse classes. In the subsequent incremental sessions, we introduce samples with fine labels for each of the 10 coarse classes, supplying only 5 samples per class at each session, which is consistent with the few-shot setting. 

   
\section{Experiments}
\label{sec:exp}

\subsection{Implementation Details}

To ensure sufficient training and testing data, we divide the data from the sub-dataset 3 into training and testing sets with a ratio of 6:4. For other sub-datasets, we use a ratio of 8:2.
All experiments are implemented using PyTorch on an NVIDIA RTX 4090 with 24 GB of memory.
The code for all unimodal OOD detection methods is derived from the OpenOOD benchmark. The code for all vision-language model (VLM) based OOD detection methods, as well as for zero-shot and closed-set classification, is sourced from Dassl. Additionally, the code for incremental learning methods is obtained from PyCIL. The detailed configs for evaluated methods are available in the \emph{appendix}.
\subsection{Closed-set \& Zero-shot Classification}
\begin{figure}[h]
\begin{subfigure}{0.48\linewidth}
\centering
\hspace{-4mm} 
\includegraphics[scale=0.195]{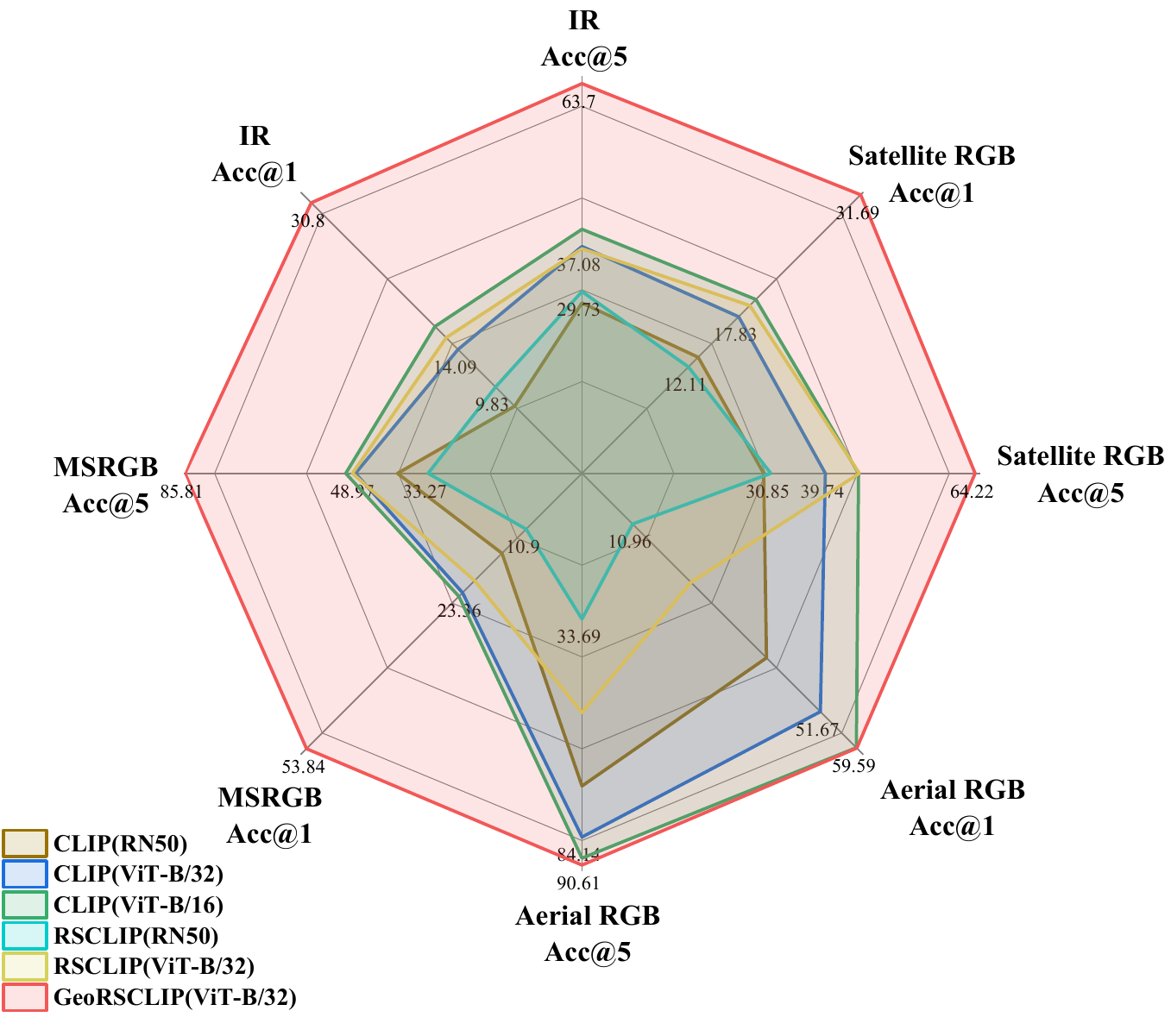}
\caption{Zero-shot classification.}
\label{fig:zs}
\end{subfigure}
\begin{subfigure}{0.48\linewidth}
\centering
\includegraphics[scale=0.195]{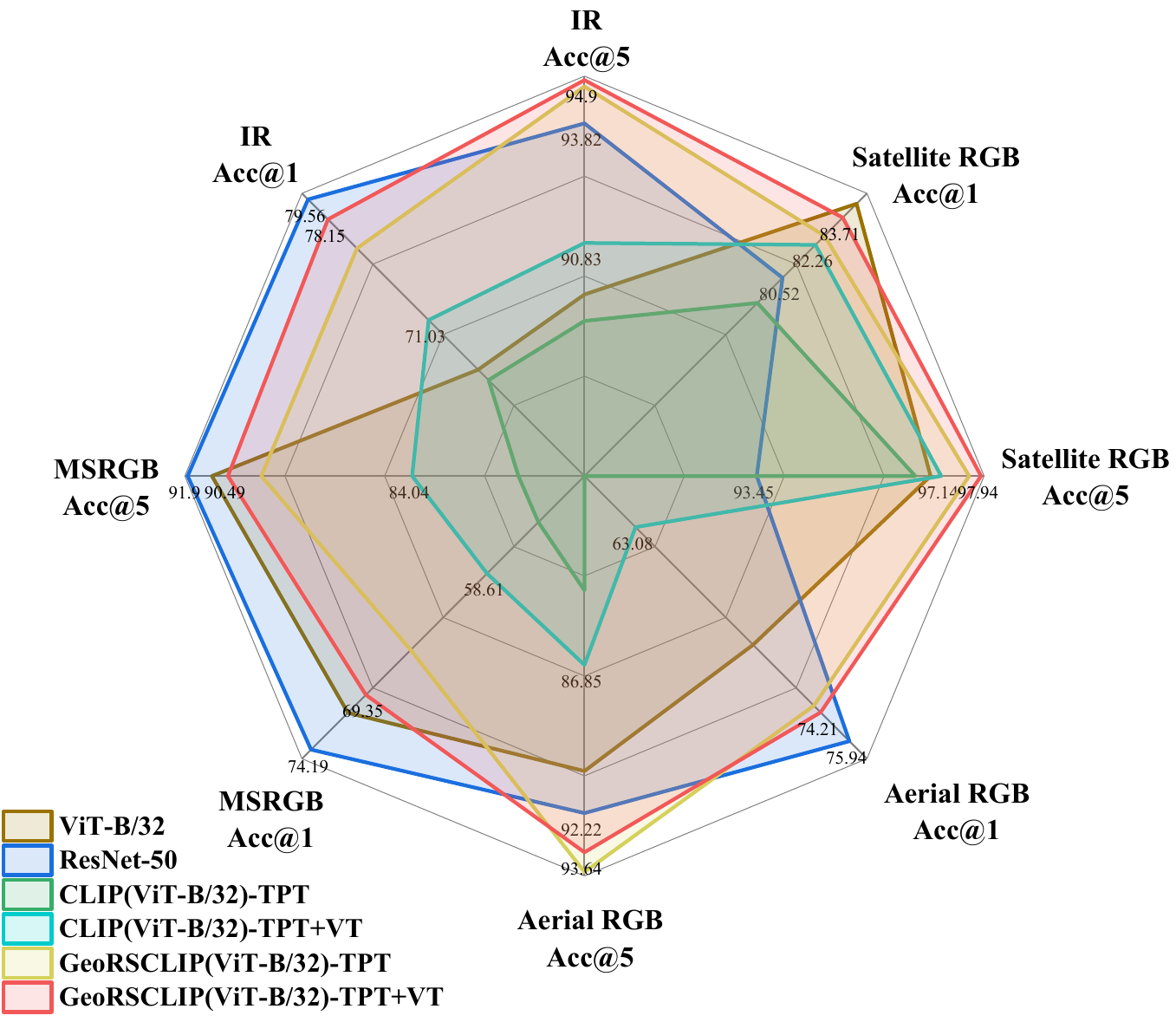}
\caption{Closed-set classification.}
\label{fig:cs}
\end{subfigure}


\caption{Performance boundary evaluation on OES.}
\label{fig:cszs}

\end{figure}

\textbf{Settings.} To test the upper and lower bounds of the open-world model's performance, we evaluate the closed-set classification and zero-shot classification capabilities on $\mathcal{D}_{R1}^{all}$, $\mathcal{D}_{R3}^{all}$, $\mathcal{D}_{R4}^{all}$, $\mathcal{D}_{R5}^{all}$ from  sub-datasets 1, 3, 4, and 5, reporting the top-1 and top-5  accuracies for each dataset. For closed-set classification, we evaluate the performance of different architectures of ResNet \cite{he2016deep}, ViT \cite{dosovitskiy2020vit}, and CLIP \cite{radford2021learning}. For CLIP, we  evaluate various finetuning methods, including Textual Prompt Tuning (TPT) and Visual Tuning (VT). TPT involves tuning the textual prompts, while VT involves an adapter following the visual encoder. For zero-shot classification, we evaluate different CLIP architectures. Tab.~\ref{fig:cszs} presents the zero-shot and closed-set classification performance of representative architectures on OES.

\begin{table*}[t]
    \centering
    \caption{OOD detection performance on \textbf{\textit{Standard}}, \textit{\textbf{Resampling Bias}}, \textit{\textbf{Modal-shift (Aerial)}},\textit{\textbf{ Modal-shift (MS)}}, and \textit{\textbf{Modal-shift (IR)}} OOD detection task. 'Near' represents the average AUROC for Near-OOD datasets, 'Far' indicates the average AUROC for Far-OOD datasets, and 'Acc' denotes the Top-1 ID classification accuracy.}
    \setlength{\tabcolsep}{0.8mm} 
    \small 
    \begin{tabular}{c|ccc|ccc|ccc|ccc|ccc}
        \toprule

        \multicolumn{1}{c}{Method} & \multicolumn{3}{c}{Standard} & \multicolumn{3}{c}{Resampling Bias} & \multicolumn{3}{c}{Modal-shift (Aerial)} & \multicolumn{3}{c}{Modal-shift (MS)} & \multicolumn{3}{c}{Modal-shift (IR)} \\
\multicolumn{1}{c}{} & \multicolumn{1}{c}{Near} & \multicolumn{1}{c}{Far} & \multicolumn{1}{c}{Acc} & \multicolumn{1}{c}{Near} & \multicolumn{1}{c}{Far} & \multicolumn{1}{c}{Acc} & \multicolumn{1}{c}{Near} & \multicolumn{1}{c}{Far} & \multicolumn{1}{c}{Acc} & \multicolumn{1}{c}{Near} & \multicolumn{1}{c}{Far} & \multicolumn{1}{c}{Acc} & \multicolumn{1}{c}{Near} & \multicolumn{1}{c}{Far} & \multicolumn{1}{c}{Acc} \\
\hline
     
\hline
\multicolumn{16}{l}{\textbf{CNN-based Methods}} \\
\hline

MSP \cite{hendrycks17baseline} & 87.90 & 95.25 & 92.01 & 66.00 & 82.34 & 47.73 & 53.32 & 60.00 & 18.41 & 64.92 & 64.94 & 46.59 & 58.64& 62.09 & 31.06 \\

ODIN \cite{odin18iclr} & 86.29 & 95.82 & 92.01 & 63.14 & 78.93 & 47.73 & 52.97 & 56.12 & 18.41 & 65.67 & 60.44 & 46.59 & 59.42 & 67.49 & 31.06 \\
 
MDS \cite{mahananobis18nips} 
& 89.71 & 97.66 & 92.01 
& 54.93 & 64.28 & 47.73
& 48.05 & 44.57 & 18.41 
& 58.44 & 69.67 & 46.59 
& 55.20 & 36.51 & 31.06 \\

ReAct \cite{sun2021react} & 88.89 & 96.80 & 92.01 & 62.90 & 79.98 & 47.73 & 52.60 & 59.74 & 18.41 & 64.21 & 63.39 & 46.59 & 59.15 & 63.39 & 31.06 \\

MLS \cite{species22icml} & 88.42 & 96.66 & 92.01 & 66.13 & 85.47 & 47.73 & 53.44 & 62.84 & 18.41 & 64.66 & 64.26 & 46.59 & 59.70 & 62.54 & 31.06 \\

KLM \cite{species22icml} & 84.45 & 94.28 & 92.01 & 63.50 & 73.04 & 47.73 & 52.36 & 52.38 & 18.41 & 62.82 & 58.74 & 46.59 & 57.52 & 53.44 & 31.06 \\

VIM \cite{wang2022vim}  & 90.87 & 98.48 & 92.01 & 58.53 & 74.07 & 47.73 & 48.90 & 48.87 & 18.41 & 59.89 & 67.93 & 46.59 & 56.32 & 40.82 & 31.06 \\

DICE \cite{sun2021dice} & 87.34 & 89.95 & 92.01 & 60.71 & 72.16 & 47.73 & 52.28 & 61.69 & 18.41 & 62.88 & 56.90 & 46.59 & 60.48 & 49.39 & 31.06 \\

EBO \cite{liu2020energy}& 88.50 & 96.90 & 92.01 & 66.01 & 86.21 & 47.73 & 53.61 & 64.15 & 18.41 & 64.37 & 63.49 & 46.59 & 60.08 & 62.37 & 31.06 \\

Relation \cite{kim2023neural} & 87.99 & 95.96 & 92.01 & 66.01 & 82.79 & 47.73 & 53.19 & 59.53 & 18.41 & 64.97 & 65.23 & 46.59 & 58.37 & 61.59 & 31.06 \\

FDBD \cite{liu2024fast} & 89.15 & 97.31 & 92.01 & 65.06 & 84.39 & 47.73 & 53.76 & 61.10 & 18.41 & 64.78 & 67.69 & 46.59 & 58.13 & 60.80 & 31.06 \\

GEN \cite{cvpr2023gen} & 88.50 & 96.87 & 92.01 & 66.43 & 86.51 & 47.73 & 53.94 & 63.20 & 18.41 & 64.71 & 64.54 & 46.59 & 60.15 & 62.19 & 31.06 \\

RMDS \cite{rmd21arxiv} & 90.26 & 96.74 & 92.01 & 58.50 & 65.27 & 47.73 & 51.73 & 50.24 & 18.41 & 64.23 & 64.06 & 46.59 & 59.43 & 56.62 & 31.06 \\

NNGuide \cite{park2023nearest} & 86.12 & 95.49 & 92.01 & 58.49 & 74.96 & 47.73 & 51.68 & 64.14 & 18.41 & 65.03 & 60.29 & 46.59 & 58.38 & 54.04 & 31.06 \\

SHE \cite{she23iclr}& 77.22 & 86.97 & 92.01 & 64.06 & 75.77 & 47.73 & 51.84 & 62.70 & 18.41 & 59.71 & 50.25 & 46.59 & 56.36 & 46.47 & 31.06 \\

\hline
\multicolumn{16}{l}{\textbf{VLM-based Methods}} \\
\hline

MaxLogits \cite{species22icml}&53.13&43.87&45.61&68.96&63.86&50.03&64.30&38.37&64.78&68.37&8.96&53.55&62.74&38.18&32.61\\
MCM  \cite{ming2022delving}&61.64&52.64&45.61&58.91&51.97&50.03&65.85&67.70&64.78&59.01&55.83&53.55&54.42&40.48&32.61\\
GL-MCM \cite{miyai2025zero}&61.85&52.48&45.61&76.48&51.80&50.03&64.92&67.26&64.78&57.36&56.78&53.55&54.75&42.29&32.61\\
CLIPN  \cite{wang2023clipn}& 52.89&56.29&28.69&49.51&48.90&38.38&59.21&55.86&62.16&45.13&66.30&28.54&44.77&78.79&21.13\\
NegLabel \cite{jiang2024negative} & 59.83 & 72.99 & 44.58 & 60.36 & 72.99 & 46.23 & 56.47 & 72.99 & 44.58 & 58.18 & 72.99 & 51.05 & 70.24 & 72.99 & 29.16 \\
CoOp \cite{zhou2022learning} & 86.30 & 93.26 & 89.97 & 66.19 & 72.80 & 69.21 & 62.75 & 74.41 & 36.73 & 65.12 & 87.35 & 72.43 & 59.74 & 38.30 & 39.68 \\
LoCoOp \cite{miyai2023locoop} & 86.72 & 91.41 & 89.79 & 68.95 & 74.01 & 71.33 & 64.00 & 75.79 & 42.86 & 68.84 & 84.64 & 74.43 & 60.86 & 40.03 & 41.06 \\
SCT \cite{yu2024selfcalibratedtuningvisionlanguagemodels} & 86.71 & 90.80 & 89.84 & 67.79 & 70.86 & 72.20 & 62.15 & 74.49 & 43.13 & 67.84 & 83.63 & 73.79 & 61.36 & 37.67 & 41.32 \\
DPM \cite{2024ECCV} & 91.02 & 98.88 & 90.84 & 73.11 & 92.10 & 68.73 & 61.11 & 74.55 & 41.17 & 72.57 & 91.58 & 73.83 & 64.71 & 76.22 & 40.16 \\

\bottomrule

\end{tabular}

\label{overview of ood1}
\end{table*}

\noindent\textbf{Results and analysis.} \textbf{(1)\textit{ Remote sensing pre-training is essential}.} Compared to aerial data, there is less satellite data available during the pre-training phase of the CLIP model. As can be seen in Fig.~\ref{fig:zs}, across all sub-datasets of satellite imagery in different modalities, GeoRSCLIP \cite{zhang2024rs5m} which is pre-trained on remote sensing images achieves significant zero-shot performance superiority. \textbf{(2)\textit{ Tuning visual encoder works}.} Satellite imagery suffers from a significant issue of insufficient pre-training. In 
 this case, finetuning more parameters of the image encoder can lead to better alignment. As shown in Fig.~\ref{fig:cs}, ResNet with all parameters fine-tuned achieves the best performance, while tuning a portion of the visual encoder to optimize visual features brings substantial enhancements to the CLIP series of models. 
\textbf{\textit{(3) Limited Cross-Domain Generalization in Foundation Models.}} Current foundation models demonstrate significantly degraded performance when processing cross-modal data, primarily due to insufficient multimodal training. This limitation reveals critical weaknesses in cross-modal alignment capabilities. Our findings highlight two key research priorities for advancing foundation models: (1) expanding the diversity and scale of multimodal training data, and (2) developing more effective cross-modal alignment methodologies.

\subsection{OOD Detection \& Generalization }

\textbf{Settings.} Following the framework of full-spectrum OOD detection \cite{yang2023full}, we unify OSR, OOD detection, and OOD generalization into a single evaluation task. We established five evaluation tasks: \textbf{\textit{Standard}}, \textit{\textbf{Resampling Bias}}, \textit{\textbf{Modal-shift (Aerial)}}\textbf{,} \textit{\textbf{Modal-shift (MS)}}\textbf{,} and \textit{\textbf{Modal-shift (IR)}}. For \textbf{\textit{Standard}} OOD detection, we use $\mathcal{D}_{R1}^{id}$ as the ID dataset. For OOD datasets, we utilize the OOD-Easy split $\mathcal{D}_{R1}^{oode}$, the OOD-Hard split $\mathcal{D}_{R1}^{oodh}$ as Near-OOD data, and SUN \cite{xiao2010sun}, which features a vast amount of natural scenes, as Far-OOD data. For \textit{\textbf{Resampling Bias}} OOD detection, we use the train set of $\mathcal{D}_{R1}^{id}$ as the ID trainset, and use the test set of $\mathcal{D}_{R2}^{id}$ as the ID test set, $\mathcal{D}_{R2}^{ood}$, $\mathcal{D}_{R1}^{oode}$, and $\mathcal{D}_{R1}^{oodh}$ as Near-OOD data, and SUN as Far-OOD data. For \textit{\textbf{Modal-shift (Aerial)}}\textbf{,} \textit{\textbf{Modal-shift (MS)}}\textbf{,} and \textit{\textbf{Modal-shift (IR)}} OOD detection, we use the train set of $\mathcal{D}_{R1}^{id}$ as the ID trainset, and use the test set of each model ($\mathcal{D}_{R3}^{id}$, $\mathcal{D}_{R4}^{id}$, and $\mathcal{D}_{R5}^{id}$) as the ID test set. We use the OOD split of each model ($\mathcal{D}_{R3}^{ood}$, $\mathcal{D}{R4}^{ood}$, and $\mathcal{D}_{R5}^{ood}$) as near-OOD data and SUN as far-OOD data. For each setting, we report the mean AUROC for both Near-OOD and Far-OOD. To evaluate the model's OSR capabilities, we also report the top-1 ID classification accuracy.

\begin{figure*}[t]
\begin{subfigure}{0.32\linewidth}
\centering
\includegraphics[scale=0.22]{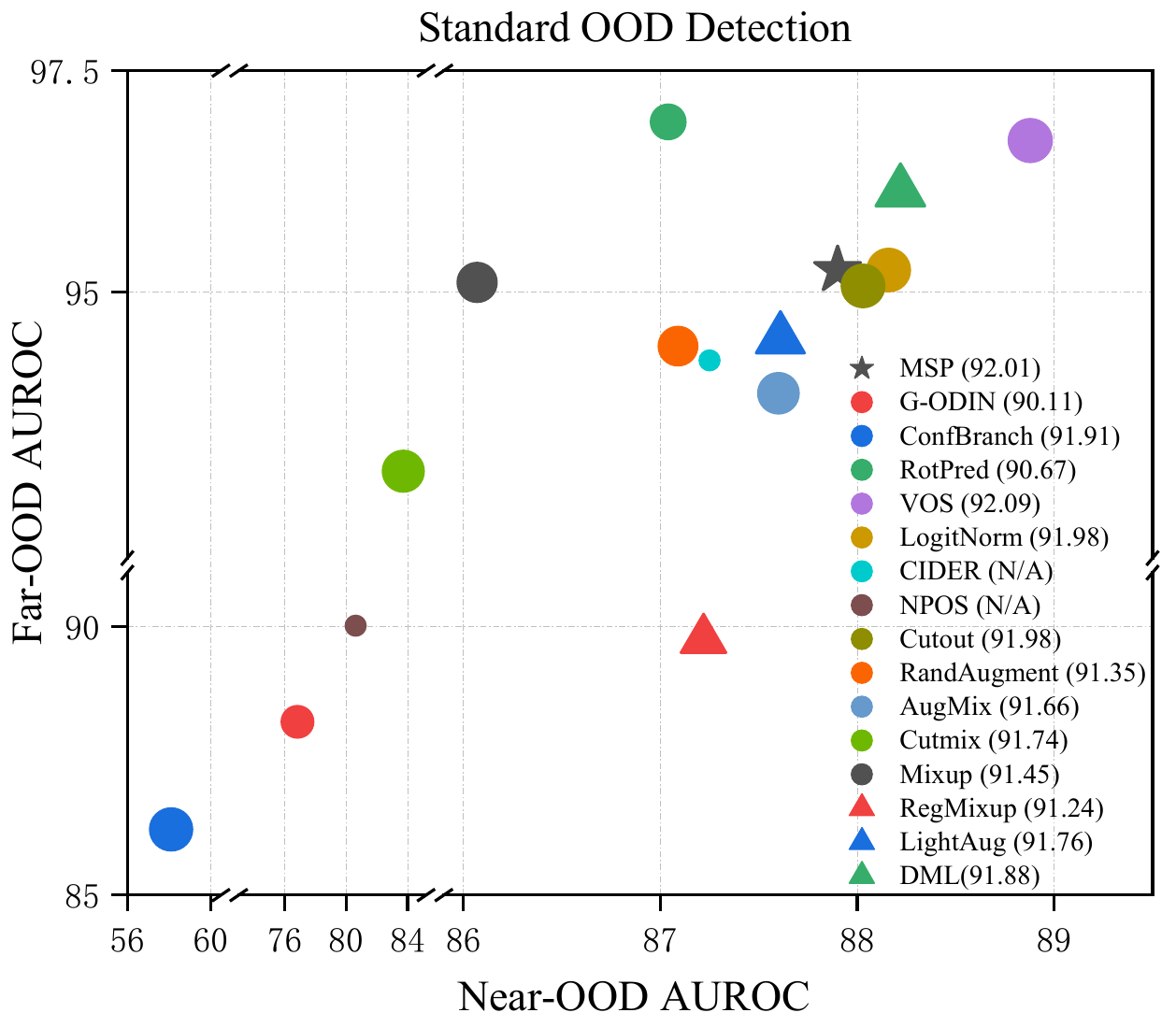}
\caption{}
\label{fig:a}
\end{subfigure}
\begin{subfigure}{0.32\linewidth}
\centering
\includegraphics[scale=0.22]{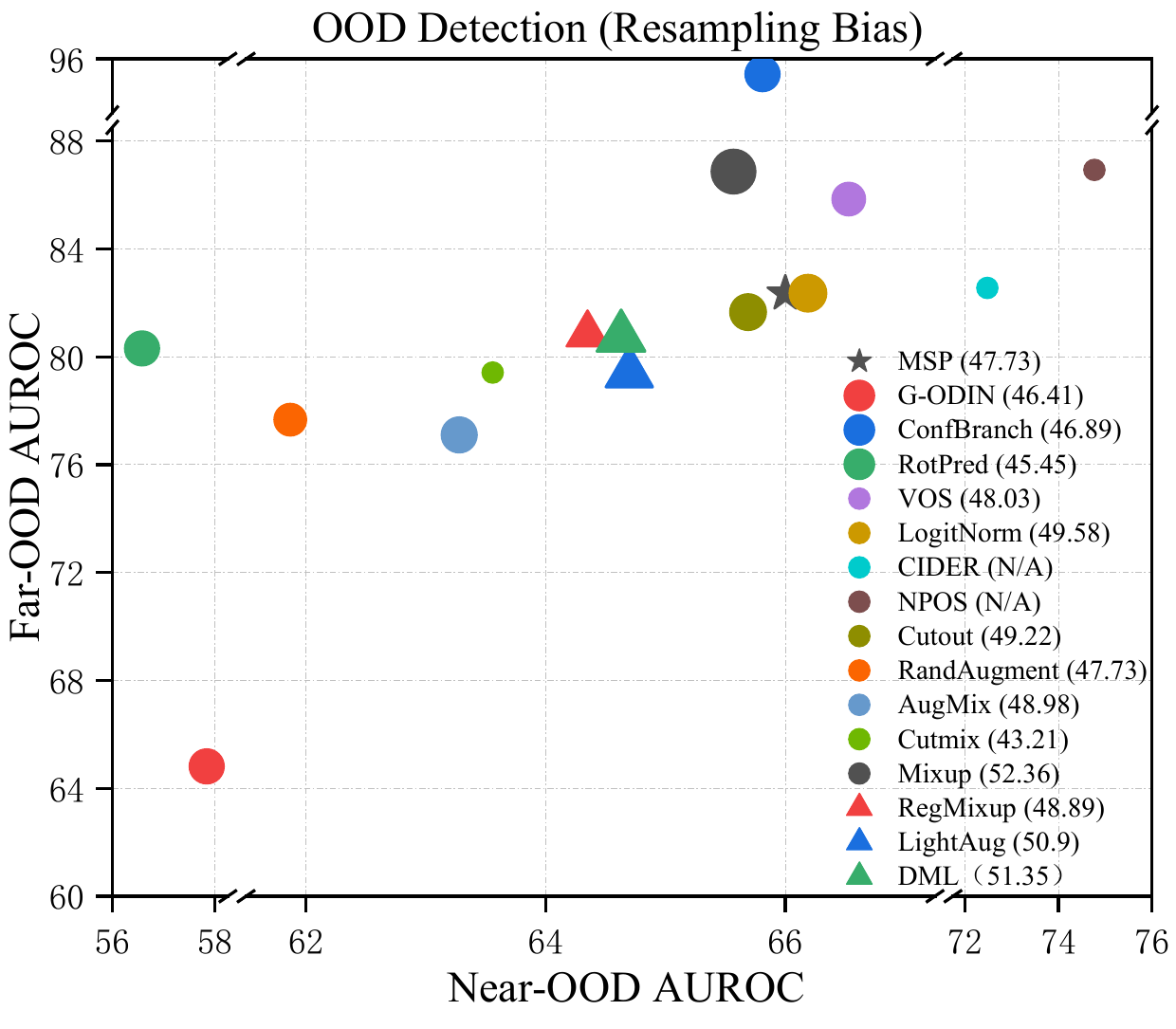}
\caption{}
\label{fig:a}
\end{subfigure}
\begin{subfigure}{0.32\linewidth}
\centering
\includegraphics[scale=0.22]{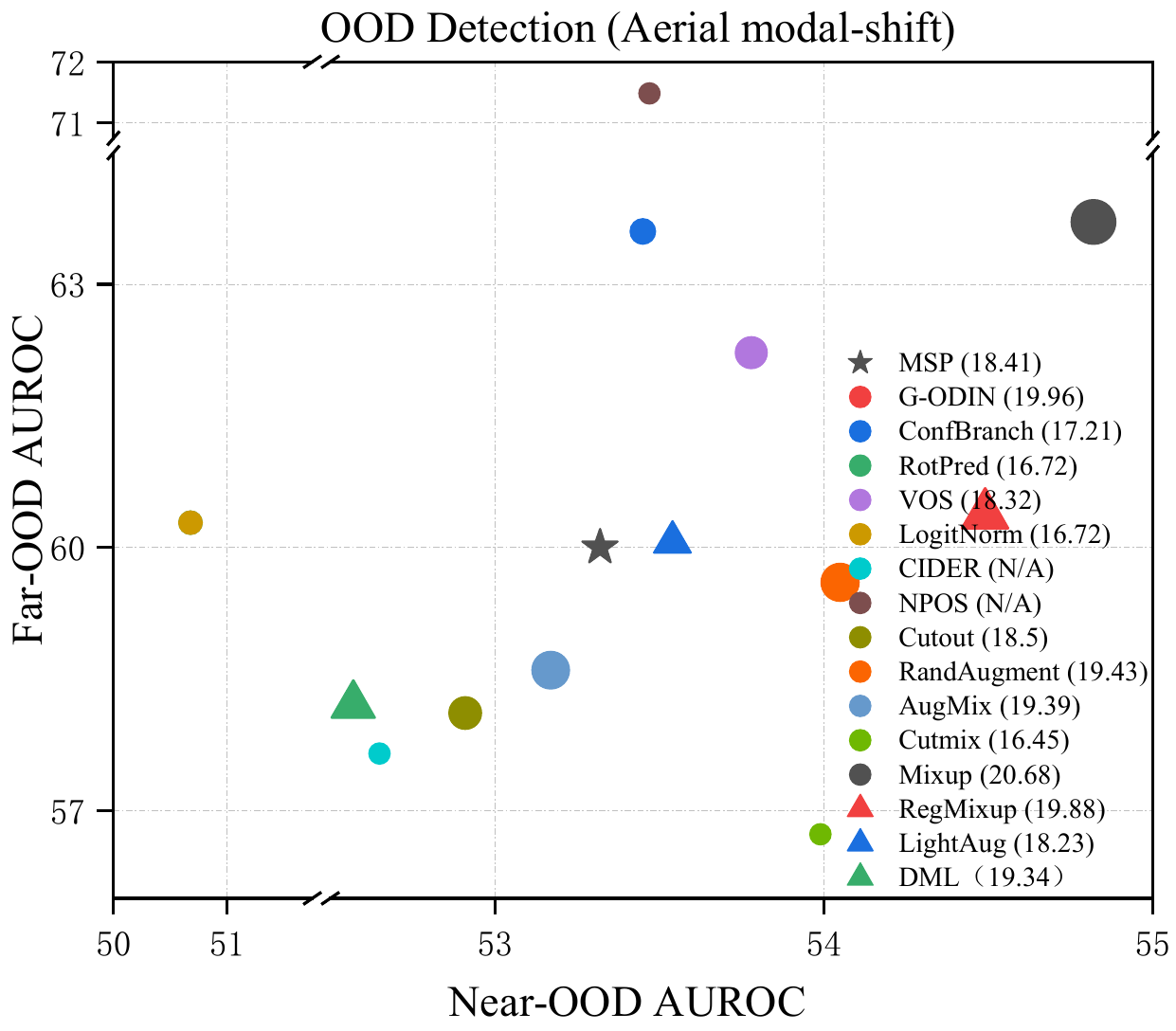}
\caption{}
\label{fig:a}
\end{subfigure}
\hspace{-0.8mm}

\caption{Full-spectrum OOD detection performances for unimodal training-required methods. Subfig(a)-(c) presents the performances on standard, resampling bias and aerial modal-shift OOD detection settings. }
\label{fig:ood}

\end{figure*}

\noindent \textbf{Baselines and evaluation methods.}We evaluate both the uni-modal OOD detection methods represented by the ResNet-50 architecture and the VLM-based OOD detection method represented by CLIP. For uni-modal OOD detection, we evaluate the post-hoc methods including OpenMax \cite{openmax16cvpr}, MSP \cite{hendrycks17baseline}, ODIN \cite{odin18iclr}, MDS \cite{mahananobis18nips}, GradNorm \cite{huang2021importance}, ReAct \cite{sun2021react}, MLS \cite{species22icml}, KLM \cite{species22icml}, VIM \cite{wang2022vim}, KNN \cite{sun2022knnood}, ASH \cite{djurisic2023extremely}, DICE \cite{sun2021dice}, EBO \cite{liu2020energy}, Relation \cite{kim2023neural}, FBDB \cite{liu2024fast}, GEN \cite{cvpr2023gen}, Rankfeat \cite{song2022rankfeat}, RMDS \cite{rmd21arxiv}, Gram \cite{gram20icml}, NNGuide \cite{park2023nearest}, Scale \cite{iclr2024scale}, SHE \cite{she23iclr} and MDSE \cite{mahananobis18nips}, training-required methods including G-ODIN \cite{hsu2020generalized}, ConfBranch \cite{confbranch2018arxiv}, RotPred \cite{rotpred},  VOS \cite{vos22iclr}, LogitNorm \cite{wei2022mitigating}, CIDER \cite{cider2023iclr}, NPOS \cite{npos2023iclr} and DML \cite{Zhang_2023_CVPR}. To further evaluate the impact of data augmentation on adapting to covariate shift, we also test the performance using CutOut \cite{devries2017improved}, RandAugment \cite{cvprw2020Randaugment}, AugMix \cite{hendrycks2020augmix}, Cutmix \cite{iccv2019cutmix}, Mixup \cite{iclr2018mixup}, RegMixup \cite{pinto2022RegMixup}, LightAug data augmentation with cross-entropy loss for training and MSP as OOD scores, where LightAug denotes augmentation applied to image brightness and grayscale. For VLM-based OOD detection, we evaluate MaxLogis \cite{species22icml}, MCM \cite{ming2022delving}, GL-MCM \cite{miyai2025zero}, CLIPN \cite{wang2023clipn}, NegLabel \cite{jiang2024negative}, CoOp \cite{zhou2022learning}, LoCoOp \cite{miyai2023locoop}, SCT \cite{yu2024selfcalibratedtuningvisionlanguagemodels},DPM \cite{2024ECCV} on CLIP with ViT-B/32, ViT-B/16, ResNet-50 and GeoRSCLIP with ViT-B/32.

\noindent\textbf{Evaluation details.} Considering that the models pre-trained on ImageNet \cite{deng2009imagenet} cannot be directly applied to OOD detection in remote sensing, we first train the model on the ID train set using Cross-Entropy loss with a learning rate of 0.01 for 100 epochs. The trained model is utilized for testing post-hoc OOD methods. For training-required and data augmentation approaches, we further fine-tune the model with a learning rate of 0.001 for 30 epochs. For VLM-based methods, we utilize the remote sensing pre-trained GeoRSCLIP \cite{zhang2024rs5m}, while the performance of other architectures is provided in the \textit{appendix.} In Tab.~\ref{overview of ood1}, we report the OOD detection results of post-hoc single-modal OOD detection methods and VLM-based methods on each OOD task. For training-required single-modal OOD detection methods and data augmentation methods,  we report the AUROC for Near-OOD and Far-OOD, as well as the ID accuracy for \textit{\textbf{Standard}}, \textit{\textbf{Resampling Bias}}, \textit{\textbf{Modal-shift (Aerial)}} tasks, as shown in Fig.~\ref{fig:ood}. Each point represents each method, with larger points indicating higher ID accuracy. The detailed hyperparameter configurations and corresponding performance metrics for each experimental setting are comprehensively documented in the \textit{appendix}.

\begin{figure*}[t]
\begin{subfigure}{0.32\linewidth}
\centering
\includegraphics[scale=0.22]{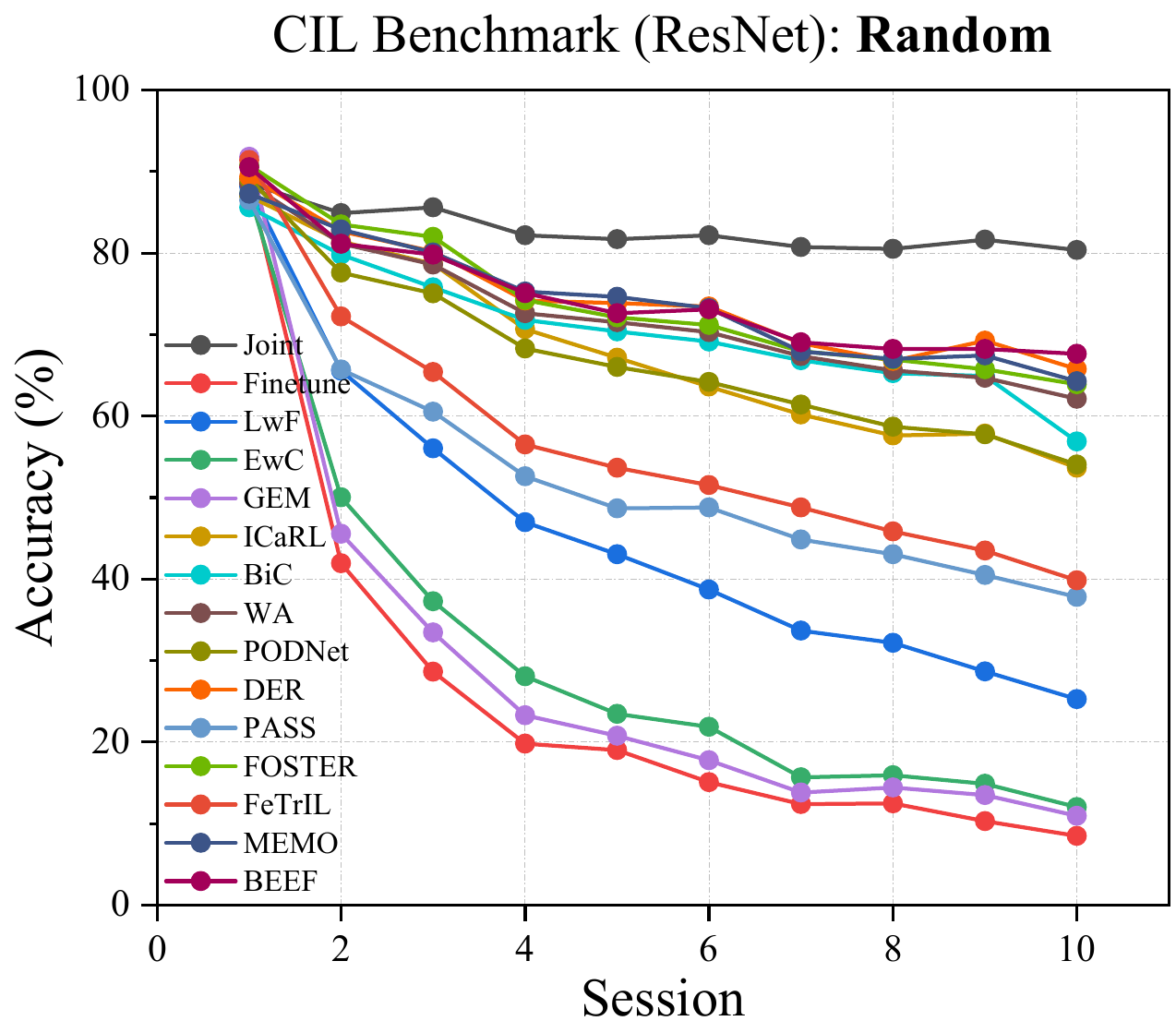}
\caption{}
\label{fig:a}
\end{subfigure}
\begin{subfigure}{0.32\linewidth}
\centering
\includegraphics[scale=0.22]{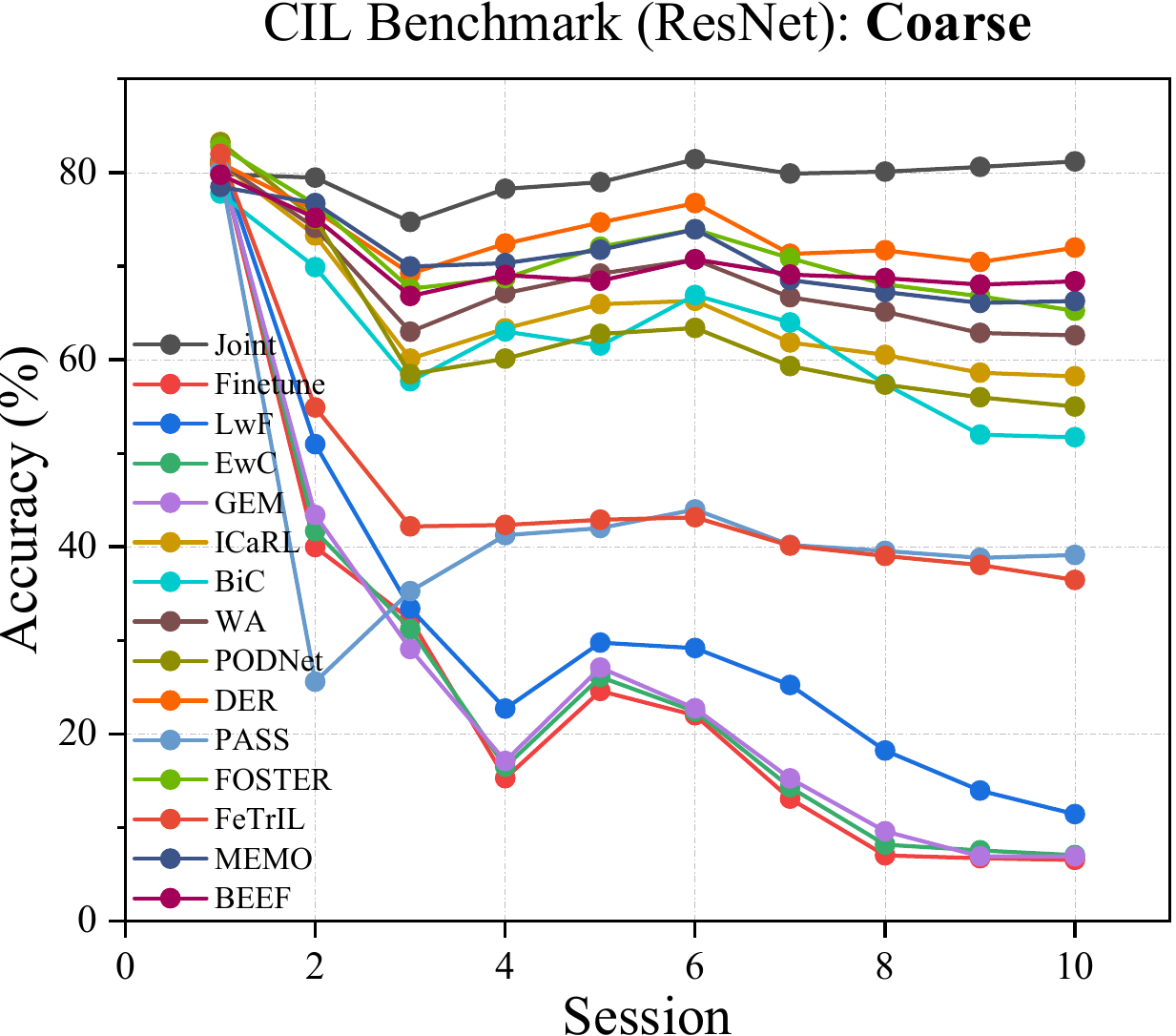}
\caption{}
\label{fig:b}
\end{subfigure}
\begin{subfigure}{0.32\linewidth}
\centering
\includegraphics[scale=0.22]{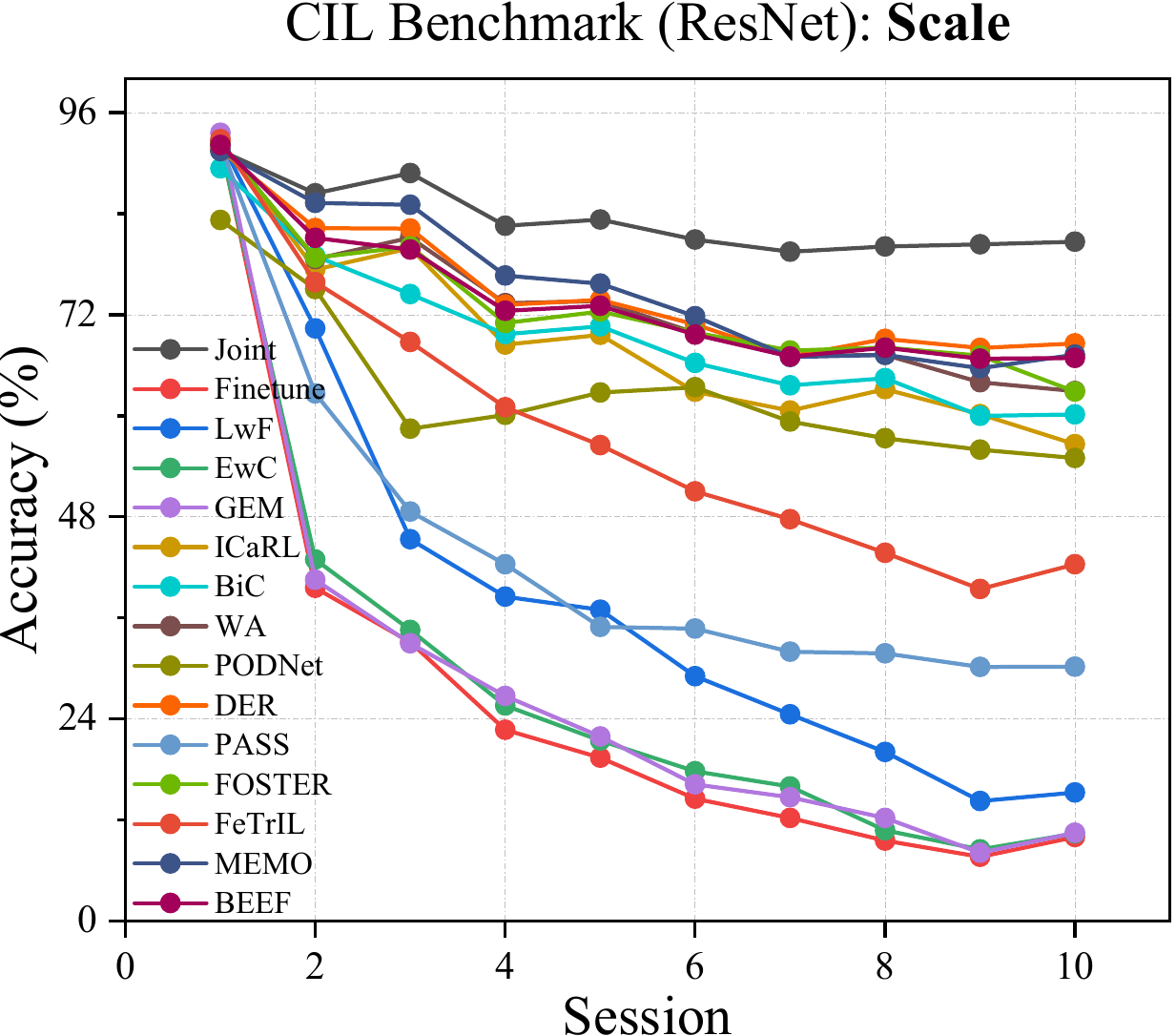}
\caption{}
\label{fig:c}
\end{subfigure} \\
\begin{subfigure}{0.32\linewidth}
\centering
\includegraphics[scale=0.22]{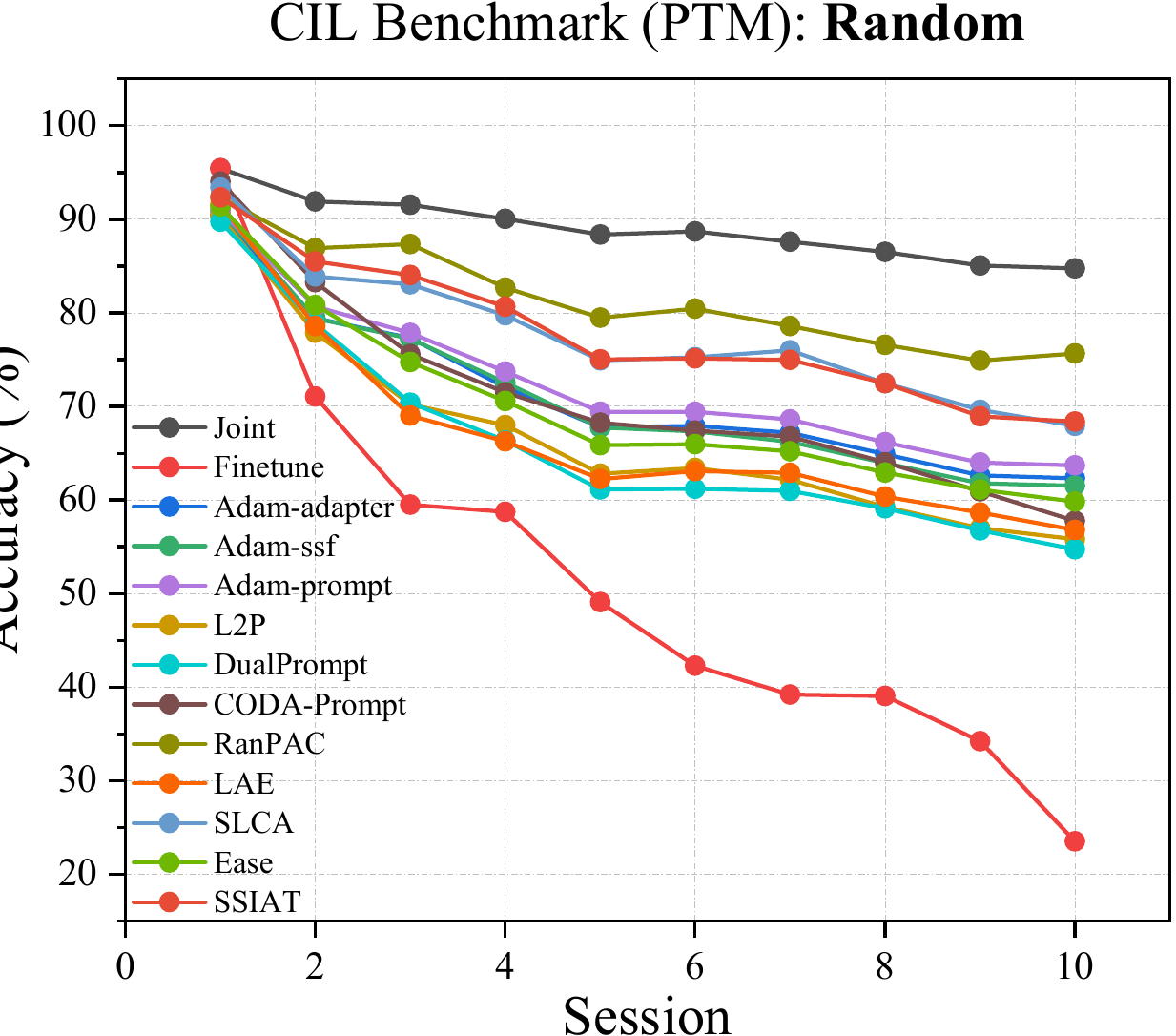}
\caption{}
\label{fig:d}
\end{subfigure}
\begin{subfigure}{0.32\linewidth}
\centering
\includegraphics[scale=0.22]{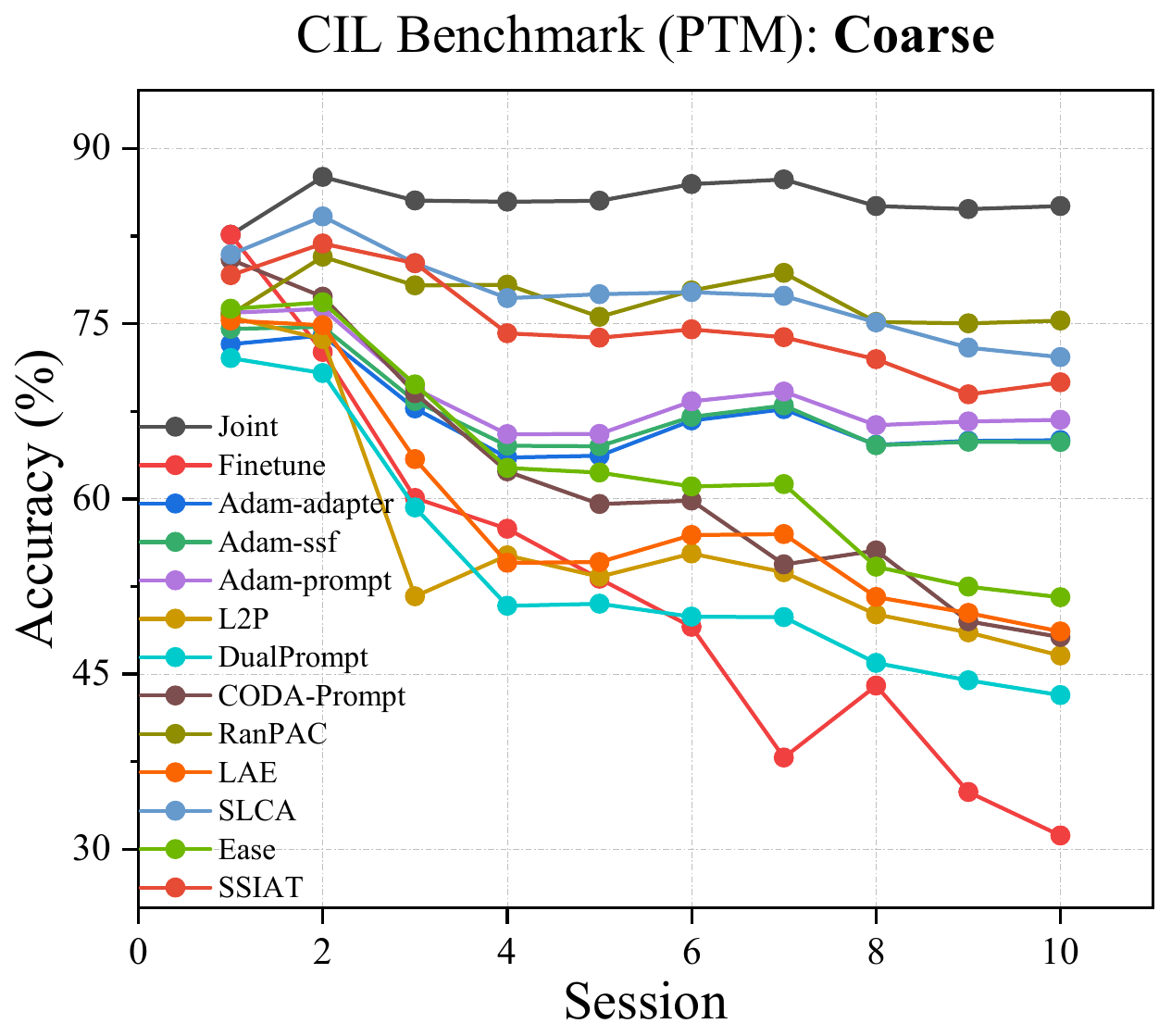}
\caption{}
\label{fig:e}
\end{subfigure}
\begin{subfigure}{0.32\linewidth}
\centering
\includegraphics[scale=0.22]{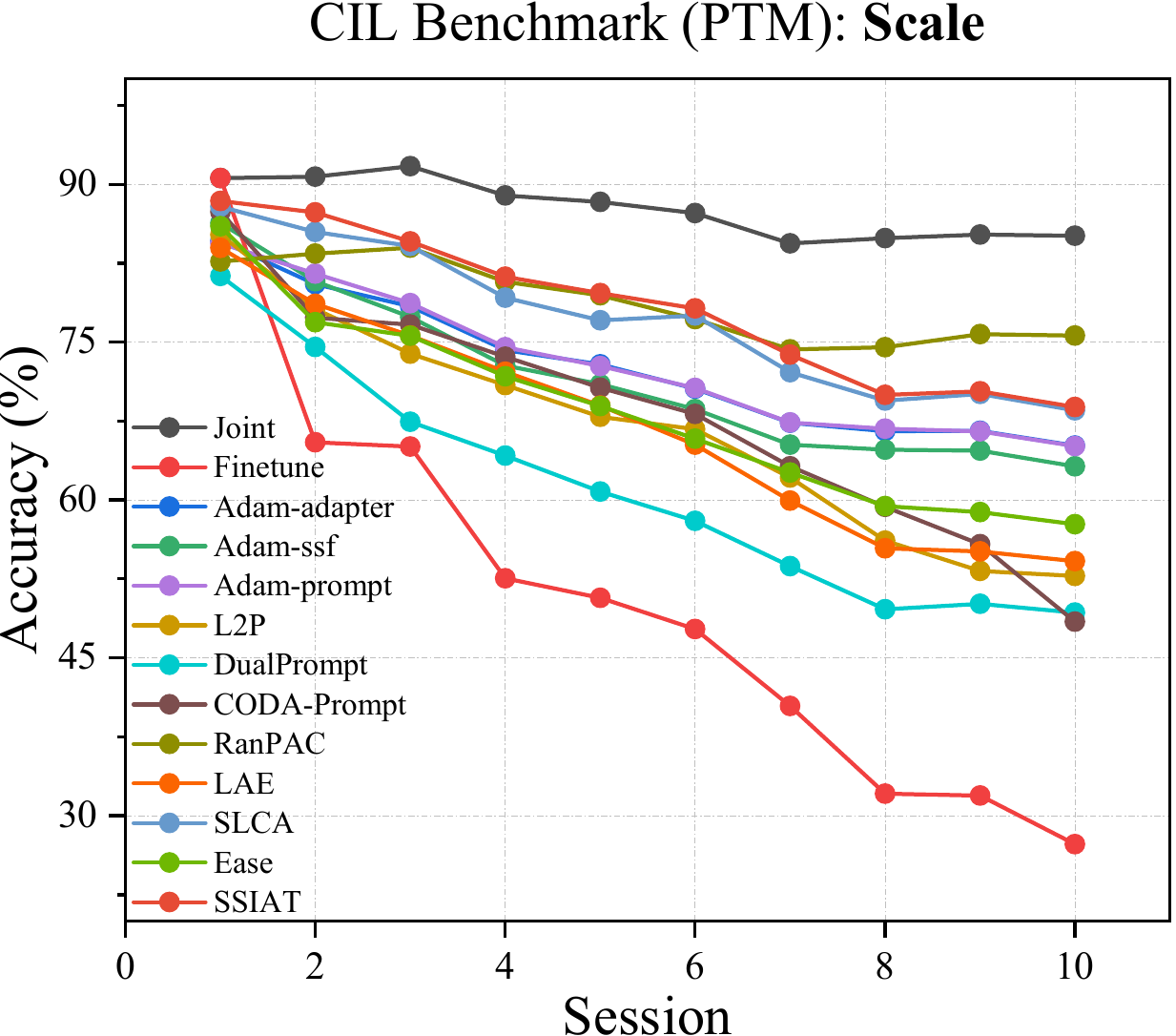}
\caption{}
\label{fig:f}
\end{subfigure}

\caption{CIL performance of all evaluated methods on benchmarks \textbf{Coarse} and \textbf{Scale}. Subfig (a) and (b) present the performance of traditional CIL methods, while subfig (c) and (d) present the performance of PTM-based methods.}
\label{CIL_re_f}
\end{figure*}

\noindent\textbf{Results and analysis.}
\textbf{(1) \textit{Simple baselines perform well}.} Among all unimodal post-hoc methods, the simple baselines MSP and MLS achieve relatively better performance across nearly all domains. Notably, MSP outperforms most training-required methods in all domains. \textbf{(2) \textit{Sufficient tuning is essential}.} VLM-based methods primarily rely on the capabilities provided by pre-training. However, training-free methods that depend solely on pre-trained VLMs, such as MCM, tend to underperform. While methods utilizing OOD labels, like NegLabel, show some improvement, they still lag behind tuning methods. Among tuning methods, DPM, which trains several visual encoder parameters, shows greater enhancement compared to methods like LoCoOp that only tune textual prompts, highlighting the importance of sufficient tuning in remote sensing data.
\textbf{(3) \textit{Full-spectrum OOD detection remains a significant challenge}.} Both unimodal and VLM-based methods exhibit performance drops when faced with covariate shift. In cases of resampling bias, average performance decreases by about 20-30\%, while for more challenging modal shifts, it drops by about 30-40\%. This suggests that covariate shifts are still challenging to existing OOD detectors. 
\textbf{(4) \textit{Data augmentation improves generalization abilities in certain cases}.} When facing specific covariate shifts, certain data augmentation methods work. For example, LightAug can enhance the performance on IR modality, while Mixup performs well with resampling bias data. However, there is still no method that performs well across all settings. Designing specific data augmentation for remote sensing is one of the improvement directions.

\subsection{Incremental Learning}
\noindent\textbf{Settings.} To test the CIL performance in more realistic scenarios, following the previous works, we evaluate the existing methods with $\mathcal{D}_{R1}^{all}$ on three benchmarks: \textbf{Random}, \textbf{Coarse} and \textbf{Scale}. In \textbf{Random}, 189 classes are divided equally among 10 sessions (18 classes in the last). In \textbf{Coarse}, classes in each session belong to the same coarse class, including \textit{Vegetation}, \textit{Agriculture}, \textit{Aviation}, \textit{Waterbody \& Facilities}, \textit{Resource Acquisition\& Utilization}, \textit{Land Transportation}, \textit{Nature \& climate}, \textit{Infrastructure}, \textit{Industrial Facilities} and \textit{Residential Building}. In \textbf{Scale}, we assign classes to 10 sessions in descending order of scale (as illustrated in Fig.~\ref{fig:scale}), with the same number of classes in each session. Besides, we evaluate the DIL performance with sub datasets containing the same semantic classes from RGB satellite ($\mathcal{D}_{R1}^{d}$), RGB aerial ($\mathcal{D}_{A}^{d}$), MSRGB ($\mathcal{D}_{M}^{d}$) and IR images ($\mathcal{D}_{I}^{d}$).

\noindent\textbf{Baselines and evaluation methods.} In CIL benchmarks, we evaluate both traditional CIL methods with ResNet-18 architecture and pre-trained model (PTM) based methods with ViT-B/16 (as illustrated in Fig.~\ref{CIL_re_f} and Tab.~\ref{CIL_re_t}), which is pre-trained on ImageNet21K and additionally fine-tuned on ImageNet1K. For traditional CIL methods, we evaluate LwF \cite{li2017learning}, EWC \cite{kirkpatrick2017overcoming}, GEM \cite{lopez2017gradient}, iCaRL \cite{rebuffi2017icarl}, BiC \cite{wu2019large}, WA \cite{zhao2020maintaining}, PODNet \cite{douillard2020podnet}, DER \cite{yan2021dynamically}, PASS \cite{zhu2021prototype}, FOSTER \cite{wang2022foster}, FeTrIL \cite{petit2023fetril}, MEMO \cite{zhou2022memo} and BEEF \cite{wang2022beef}. For PTM-based methods, we evaluate Adam \cite{zhou2024revisiting}, L2P \cite{wang2022learning}, DualPrompt \cite{wang2022dualprompt}, CODA-Prompt \cite{smith2023coda}, RanPAC \cite{mcdonnell2023ranpac}, LAE \cite{gao2023unified}, SLCA \cite{zhang2023slca}, Ease \cite{zhou2024expandable} and SSIAT \cite{tan2024semantically}. We also evaluate sequential finetuning as the lower bound performance and joint training as the upper bound performance. In DIL benchmarks, we also evaluate traditional methods and PTM-based methods. For traditional methods, we evaluate LwF, EWC, BEEF and DS-AL \cite{zhuang2024ds}. For PTM-based methods, we evaluate S-Prompt \cite{wang2022s}. In C2FSCIL benchmark, we evaluate traditional CIL methods and Knowe \cite{xiang2022coarse}. Traditional CIL methods include LwF, WA and ScaIL \cite{belouadah2020scail}. Knowe is an effective method designed for C2FSCIL.

\begin{table}[h]
    \caption{Evaluation on CNN-based and ViT-based methods with different settings: \textbf{Random}, \textbf{Coarse} and \textbf{Scale}. $\mathcal{A}_{Last}$ and $\mathcal{A}_{Avg}$ denote the last session and average accuracy respectively.}
    \footnotesize
    \setlength{\tabcolsep}{0.2mm}
    \begin{tabular}{ccccccc}
    \hline 
    \multicolumn{1}{c}{\multirow{2}{*}{Method}} 
    & \multicolumn{2}{c}{\multirow{1}{*}{\textbf{Random}}}
    & \multicolumn{2}{c}{\multirow{1}{*}{\textbf{Coarse}}}
    & \multicolumn{2}{c}{\multirow{1}{*}{\textbf{Scale}}}\\
    & \multicolumn{1}{c}{\multirow{1}{*}{$\mathcal{A}_{Last}$ $\uparrow$}}
    & \multicolumn{1}{c}{\multirow{1}{*}{$\mathcal{A}_{Avg}$ $\uparrow$}}
    & \multicolumn{1}{c}{\multirow{1}{*}{$\mathcal{A}_{Last}$ $\uparrow$}}
    & \multicolumn{1}{c}{\multirow{1}{*}{$\mathcal{A}_{Avg}$ $\uparrow$}}
    & \multicolumn{1}{c}{\multirow{1}{*}{$\mathcal{A}_{Last}$ $\uparrow$}}
    & \multicolumn{1}{c}{\multirow{1}{*}{$\mathcal{A}_{Avg}$ $\uparrow$}}\\
    \hline
    \multicolumn{7}{c}{\textbf{CNN-based Methods}}\\
    \hline
    Joint & 80.37 & 82.82 & 81.19 & 79.45 & 80.67 & 83.42\\
    Finetune & 8.49 & 25.68 & 6.52 & 24.83 & 9.95 & 26.10\\
    \hline
    LwF \cite{li2017learning}  & 25.26 & 45.82 & 11.40 & 31.60 & 15.25 & 38.70\\
    EWC \cite{kirkpatrick2017overcoming} & 12.03 & 30.65 & 7.03 & 25.60 & 10.39 & 27.99 \\
    GEM \cite{lopez2017gradient} & 10.95 & 28.53 & 6.90 & 25.78 & 10.47 & 27.74 \\
    iCaRL \cite{rebuffi2017icarl} & 53.64 & 67.77 & 58.21 & 64.91 & 56.62 & 69.12 \\
    BiC \cite{wu2019large} & 56.89 & 70.64 & 51.71 & 62.18 & 60.15 & 69.76 \\
    WA \cite{zhao2020maintaining} & 62.12 & 72.23 & 62.60 & 68.23 & 62.94 & 73.03 \\
    PODNet \cite{douillard2020podnet} & 54.05 & 67.21 & 54.98 & 63.05 & 54.98 & 63.05 \\
    DER \cite{yan2021dynamically} & 65.75 & 74.43 & 71.96 & 73.54 & 68.60 & 74.67 \\
    PASS \cite{zhu2021prototype} & 37.80 & 52.89 & 39.11 & 42.59 & 30.18 & 43.95 \\
    FOSTER \cite{wang2022foster} & 63.88 & 73.82 & 65.21 & 71.25 & 62.87 & 73.04 \\
    FeTrIL \cite{petit2023fetril} & 39.85 & 56.87 & 36.44 & 46.10 & 42.35 & 57.92 \\
    MEMO \cite{zhou2022memo} & 64.26 & 74.08 & 66.27 & 70.92 & 67.26 & 75.31 \\
    BEEF \cite{wang2022beef} & 67.62 & 74.47 & 68.37 & 70.41 & 66.87 & 73.69 \\
    \hline
    \multicolumn{7}{c}{\textbf{ViT-based Methods}}\\
    \hline
    Joint &84.75 & 88.99 & 85.07 & 85.60 & 85.11 & 87.71\\
    Finetune & 23.51 & 51.20 & 31.18 & 52.29 & 27.30 & 50.38\\
    \hline
    Adam-adapter \cite{zhou2024revisiting} & 62.31 & 71.19 & 65.04 & 67.13 & 65.15 & 72.68 \\
    Adam-ssf \cite{zhou2024revisiting} & 61.52 & 70.86 & 64.87 & 67.61 & 63.20 & 71.48 \\
    Adam-prompt \cite{zhou2024revisiting} & 63.69 & 72.45 & 66.78 & 69.01 & 65.12 & 72.82 \\
    L2P \cite{wang2022learning} & 55.79 & 66.68 & 46.60 & 56.38 & 52.78 & 66.71 \\
    DualPrompt \cite{wang2022dualprompt} & 54.72 & 65.92 & 43.21 & 53.74 & 49.29 & 60.90 \\
    CODA-Prompt \cite{smith2023coda} & 57.77 & 70.85 & 48.17 & 61.63 & 48.44 & 68.06 \\
    RanPAC \cite{mcdonnell2023ranpac} & 75.65 & 81.49 & 75.27 & 77.14 & 75.63 & 78.76 \\
    LAE \cite{gao2023unified} & 56.79 & 66.94 & 48.65 & 58.71 &54.19 & 66.93 \\
    SLCA \cite{zhang2023slca} & 67.92 & 77.62 & 72.14 & 77.53 & 68.52 & 77.14 \\
    Ease \cite{zhou2024expandable} & 59.81 & 69.84 & 51.58 & 62.84 & 57.69 & 68.35 \\
    SSIAT \cite{tan2024semantically} &68.37 & 77.74 & 69.98 & 74.85 & 68.83 & 78.23 \\
    \hline 
    \end{tabular}
    \label{CIL_re_t}
\end{table}

\noindent\textbf{Results and analysis.} \textbf{(1)\textit{ Catastrophic forgetting remains serious}.} In the analysis of remote sensing  data in OES, the evaluated methods prove successful in alleviating catastrophic forgetting. Nevertheless, in relation to the upper bound performance, the majority of the methods exhibit varying degrees of forgetting, showcasing a performance decline by $10-20\%$. Notably, the most effective method, RanPAC, shows a comparatively smaller performance decline by $5-10\%$. \textbf{(2)\textit{ Benchmarks closer to real-world environments show poorer performance}.} In the settings of \textbf{Coarse} and \textbf{Scale}, it is observed that CIL performance typically falls short when compared to performance in \textbf{Random}. This observation underscores the heightened complexity of CIL within practical environments, wherein models leverage diverse coarse class data acquired from distinct specialized satellites, along with scale data gathered by satellites operating on varied orbits and possessing different resolution capabilities. \textbf{(3)\textit{ Limited performance gains from pre-trained models}.} In contrast to the significant performance enhancement in CIL achieved by the PTM when applied to natural images, the performance improvements are constrained when the PTM is utilized on remotely sensed images. This limitation could be attributed to the inadequate generalization capacity of the PTM, which is hindered by the domain gap between natural images and remotely sensed images. \textbf{(4) \textit{PTM exhibits both adaptability and limitations}.} Leveraging pre-trained knowledge, the model can adapt well to a specific data domain. However, as the data domain continues to evolve, continuous finetuning leads  to a significant degradation in the performance of the PTM. \textbf{(5)\textit{ C2FSCIL remains a significant challenge}.} In C2FSCIL setting, most existing methods struggle to  balance the performance between coarse and fine classes. Methods such as LwF, ScaIL, and WA experience significant degradation in coarse-grained class performance due to continual finetuning, despite incorporating various strategies to mitigate forgetting in the finetuning process.

\begin{table}[t]
\centering
\caption{The experimental results of DIL.}
\label{tab:DIL}
\footnotesize
\setlength{\tabcolsep}{2pt}
\renewcommand{\arraystretch}{1.5}
\begin{tabular}{c|cccc|cccc} 
\hline
Methods & $\mathcal{D}_{R1}^{d}$ & $\mathcal{D}_{A}^{d}$ & $\mathcal{D}_{M}^{d}$ & $\mathcal{D}_{I}^{d}$ & $\mathcal{D}_{I}^{d}$ & $\mathcal{D}_{M}^{d}$ & $\mathcal{D}_{A}^{d}$ &
$\mathcal{D}_{R1}^{d}$ \\
\hline
Joint & \multicolumn{4}{c|}{47.68} & \multicolumn{4}{c}{68.39} \\
Finetune & 3.85 & 32.00 & 20.60 & 8.90 & 45.38 & 43.41 & 26.33 & 20.66\\
\hline
LwF \cite{li2017learning}  & 3.75 & 30.69 & 18.84 & 13.04  & 45.71 & 36.16 & 28.28 & 20.96\\
EWC \cite{kirkpatrick2017overcoming}  & 3.75 & 32.02 & 21.30 & 8.23 & 71.99 & 40.60 & 33.04 & 31.88\\
BEEF \cite{wang2022beef}  & 3.80 & 38.56 & 29.41 & 32.72 & 46.23 & 49.38 & 44.84 & 44.91 \\
DS-AL \cite{zhuang2024ds} & 3.16 & 27.60 & 34.40 & 35.13 & 44.60 & 23.18 & 29.14 & 29.89 \\
S-Prompt \cite{wang2022s} & 95.47 & 65.14 & 45.15 & 28.72 & 95.02 & 64.19 & 44.22 & 28.16\\
\hline
\end{tabular}
\end{table}

\begin{table}
\centering
\caption{The experimental results of C2FSCIL.}
\label{C2S}
\setlength{\tabcolsep}{5pt} 
\renewcommand{\arraystretch}{1.5} 
\begin{tabular}{c|cccc} 
\hline
Methods & $\mathcal{A}_{Total}$ & $\mathcal{A}_{Now}$ & $\mathcal{A}_{Coarse}$ & $\mathcal{A}_{Fine}$  \\
\hline
Joint & 57.78 & 63.28 & 59.14  & 67.77 \\
Finetune & 38.11 & 73.44 & 18.84 & 51.68\\
\hline
LwF \cite{li2017learning}  & 12.49 & 78.56 & 7.33 & 19.67 \\
WA \cite{zhao2020maintaining}  & 33.60 & 62.90 & 11.01 & 51.79 \\
ScaIL \cite{belouadah2020scail} & 12.49 & 78.56 & 7.33 & 19.67 \\
Knowe \cite{xiang2022coarse}& 54.13 & 62.18 & 85.33 & 55.13 \\
\hline
\end{tabular}

\end{table}

\section{Conclusion}

In this paper, we present OpenEarthSensing (OES), a novel large-scale benchmark designed to evaluate semantic and domain shifts in open-world scenarios. Unlike existing datasets with limited scope, OES integrates five diverse sub-datasets spanning five domains and three modalities, providing a comprehensive testbed to assess model robustness under both semantic shifts (e.g., novel categories) and covariate shifts (e.g., modal distribution changes). Through extensive experiments, we benchmark state-of-the-art open-world models on OES for critical tasks including out-of-distribution detection and incremental learning. Our results reveal significant challenges, particularly in recognizing unseen semantic-shift categories and adapting to abrupt distributional changes, highlighting the limitations of current approaches in dynamic environments. OES demonstrates substantially higher difficulty compared to conventional benchmarks, underscoring the urgent need for advanced methods to handle open-world dynamics. These results establish OES as a rigorous evaluation platform for real-world adaptability and pave the way for future research in robust, adaptive learning systems.

\clearpage

\begin{appendices}

\section{Details of OES}

\subsection{Detailed description of notations.}
We provide a detailed explanation of all notations used in Table 1 of the main paper, as shown in Tab.~\ref{table:notations}.

\subsection{Detailed statistics of OES}

Tab.~\ref{tab:subdatasets}  presents  the compositions of Sub-datasets 1, 2, 4, and 5. The data from Sub-dataset 3 are sourced from CC3M \cite{sharma2018conceptual} and RS5M \cite{zhang2024rs5m}. Overall, OES contains data from 23 public available datasets, comprising 189 categories and a total of 157,674 images. In Tab.~\ref{table:coarse}, we provide the correspondence between all coarse-grained and fine-grained categories, as well as the OOD split included in OES.

\subsection{Licencing Details}
All images in the OES dataset are collected from publicly available sources. It is important to note that OES does not provide a unified usage license. Instead, the permissible usage of OES is strictly governed by the individual license terms and restrictions of each constituent dataset. For specific licensing information, please refer to the component dataset licenses presented in Tab.~\ref{lic}.

\subsection{Details about scale in OES}

In Tab.~\ref{table:scene} and Tab.~\ref{table:object}, we provide the scale scores of all categories  and the corresponding task divisions. In our methodology, we first conduct manual screening to identify relevant scene and object categories. These categories are then organized in descending order of their scale for incremental learning sessions. To quantitatively assess relative scale measurements, we employ Qwen-VL-chat \cite{bai2023qwen} to assign standardized scores and establish rankings for both scene and object categories independently. This scale-based ranking system ultimately determines the configuration of our incremental learning sessions.

\subsection{Details about multi-modal images}

Beyond scale variations, OES's four sub-datasets enable multiple modal shift scenarios, including RGB band to all-band or IR-band. While standalone IR images are uncommon in satellite imagery, they frequently occur in drone data. For the main component of the OES dataset, we employ three-channel visible light imagery instead of full-spectrum data, based on the following key considerations: (1) compatibility with pretrained 3-channel models (e.g., VLMs), (2) consistency with established open-world methods, which use 3-channel RGB data (including OOD detection and incremental learning), and (3) limitations of publicly available datasets—where existing 13-band collections either exhibit insufficient resolution (e.g., BigEarthNet) or inconsistent spectral coverage (e.g.,  4-8 bands in fMoW \cite{christie2018functional}). For MSRGB data in sub-dataset 4, the majority of images are sourced from 50 categories of the fMoW dataset, while the remainder originate from USTC SmokeRS \cite{ba2019smokenet} and MRSSC2.0 \cite{liu2022remote}, maintaining visual consistency with the fMoW style. Sub-dataset 5 consists of infrared images from two sources: (1) infrared bands extracted from multispectral images in fMoW, BigEarthNet \cite{sumbul2019bigearthnet}, MRSSC2.0, and NaSC-TG2 \cite{zhou2021nasc}, and (2) aerial drone imagery from VisDrone \cite{zhu2021detection}.

\subsection{Geospatial analysis}

\begin{figure*}
\hsize=\textwidth 
\centering
\includegraphics[width=0.92\textwidth]{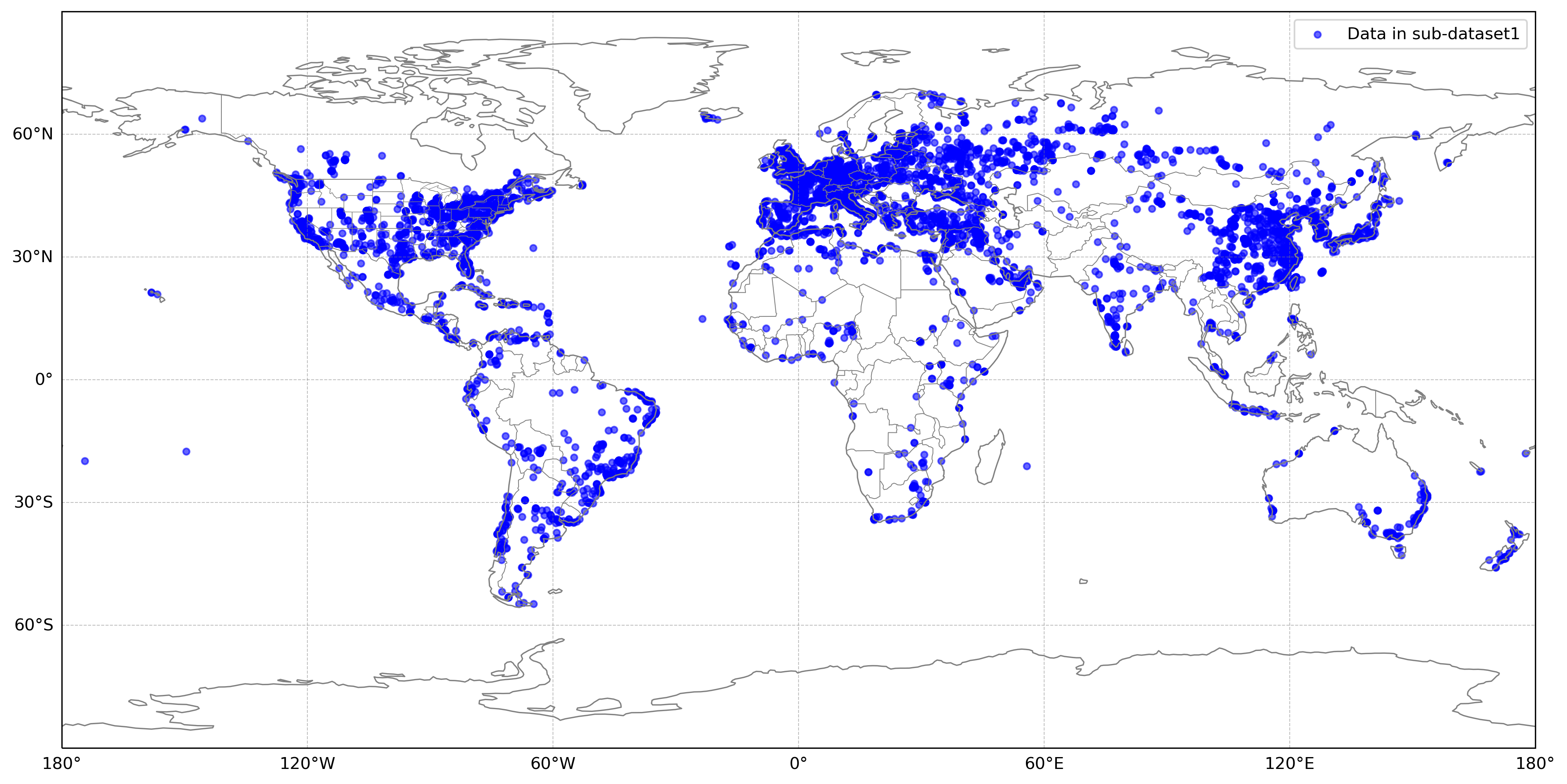}

\caption{Data distribution on OES, we randomly selected 10,000 images with geospatial metadata from Sub-dataset 1 for display.
}

\label{data_distribution_map}
\end{figure*}

Geospatial metadata is essential for analyzing covariate shifts in remote sensing, yet most OES images from public datasets lack these annotations. Only a few datasets available for OES (e.g., fMoW \cite{christie2018functional}, RSD46-WHU \cite{xiao2017high}, FAIR1M \cite{sun2022fair1m}, BigEarthNet \cite{sumbul2019bigearthnet} and NaSC-TG2 \cite{zhou2021nasc}) provide complete spatio metadata. To address this gap, we simulate distribution shifts using images of the same category across different datasets, capturing temporal, geographic, and sensor variations. Although we cannot provide spatio metadata for all classes, we conduct a visualization analysis of spatial characteristics on sub-dataset 1 using available data, as shown in Fig.~\ref{data_distribution_map}.

\subsection{Example Images on OES}
\begin{figure*}[h]

\begin{subfigure}{0.24\linewidth}
\centering
\includegraphics[scale=0.11]{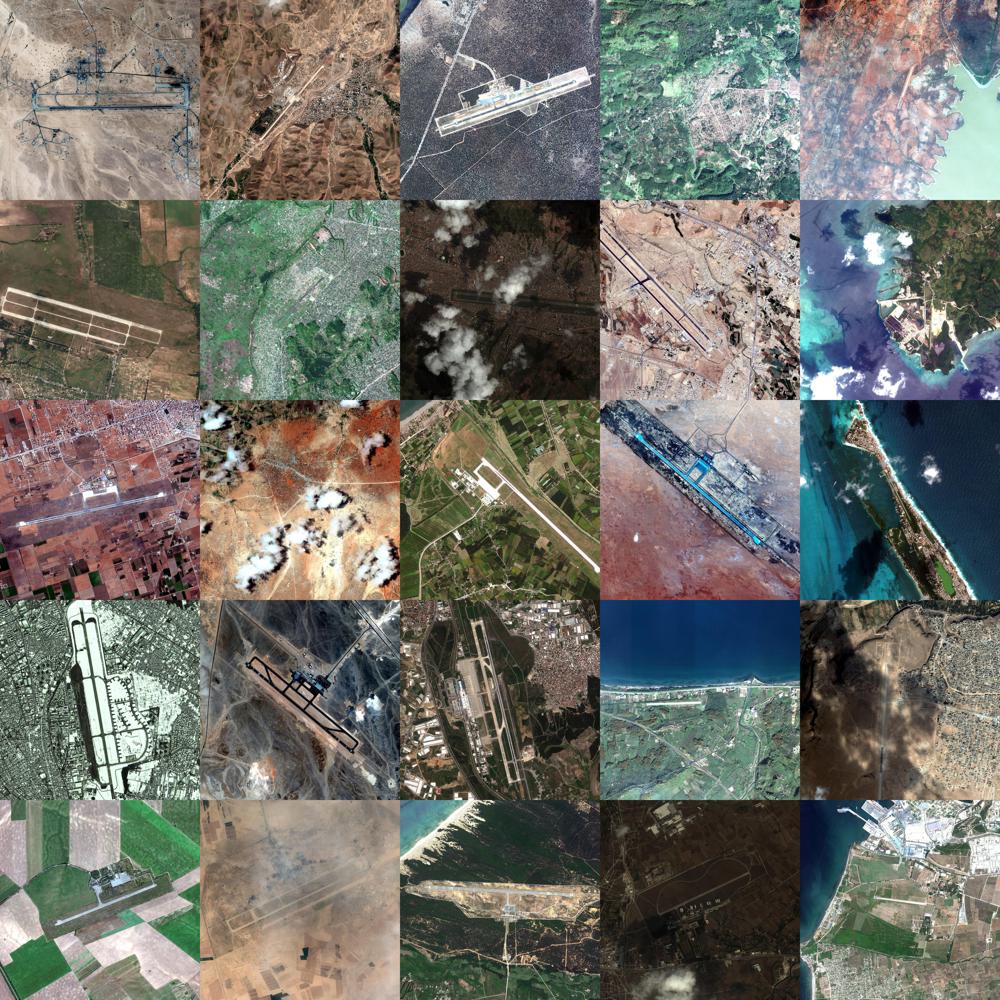}
\caption{Airport}
\label{fig:a}
\end{subfigure}
\begin{subfigure}{0.24\linewidth}
\centering
\includegraphics[scale=0.11]{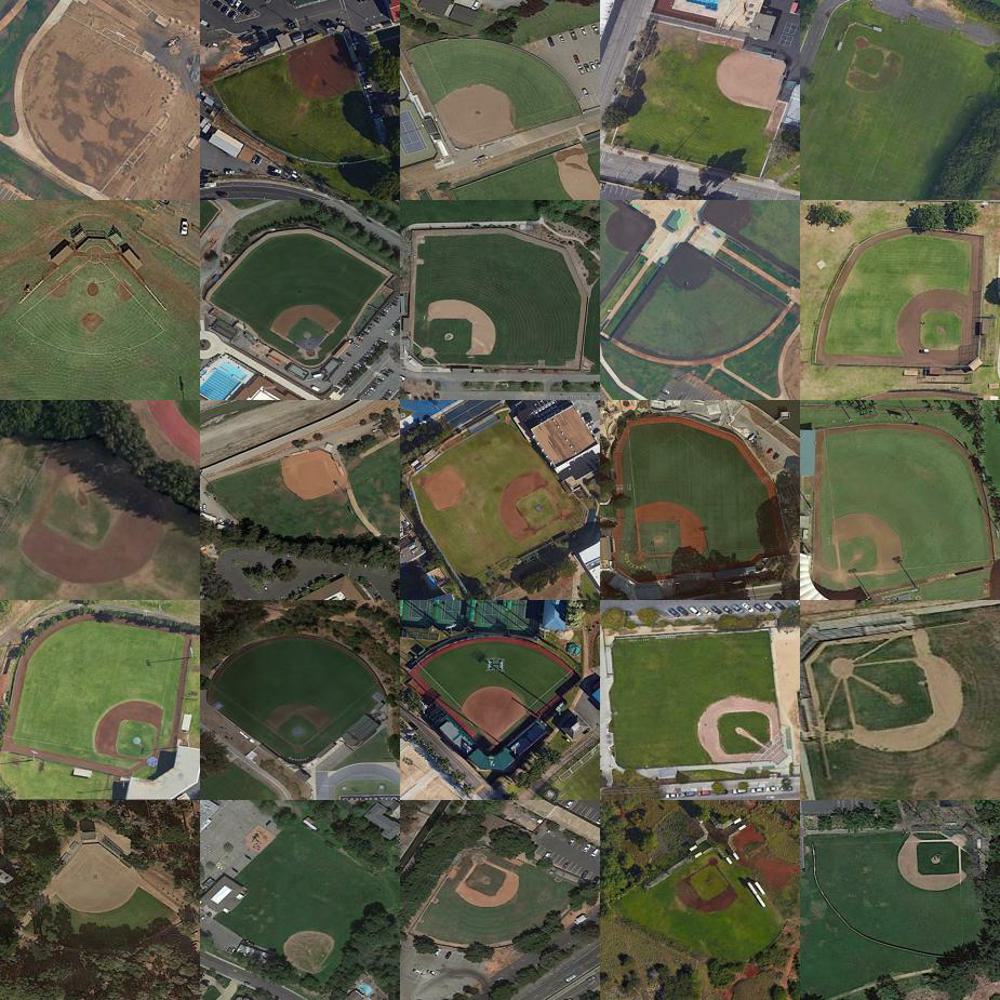}
\caption{Baseball field}
\label{fig:a}
\end{subfigure}
\begin{subfigure}{0.24\linewidth}
\centering
\includegraphics[scale=0.11]{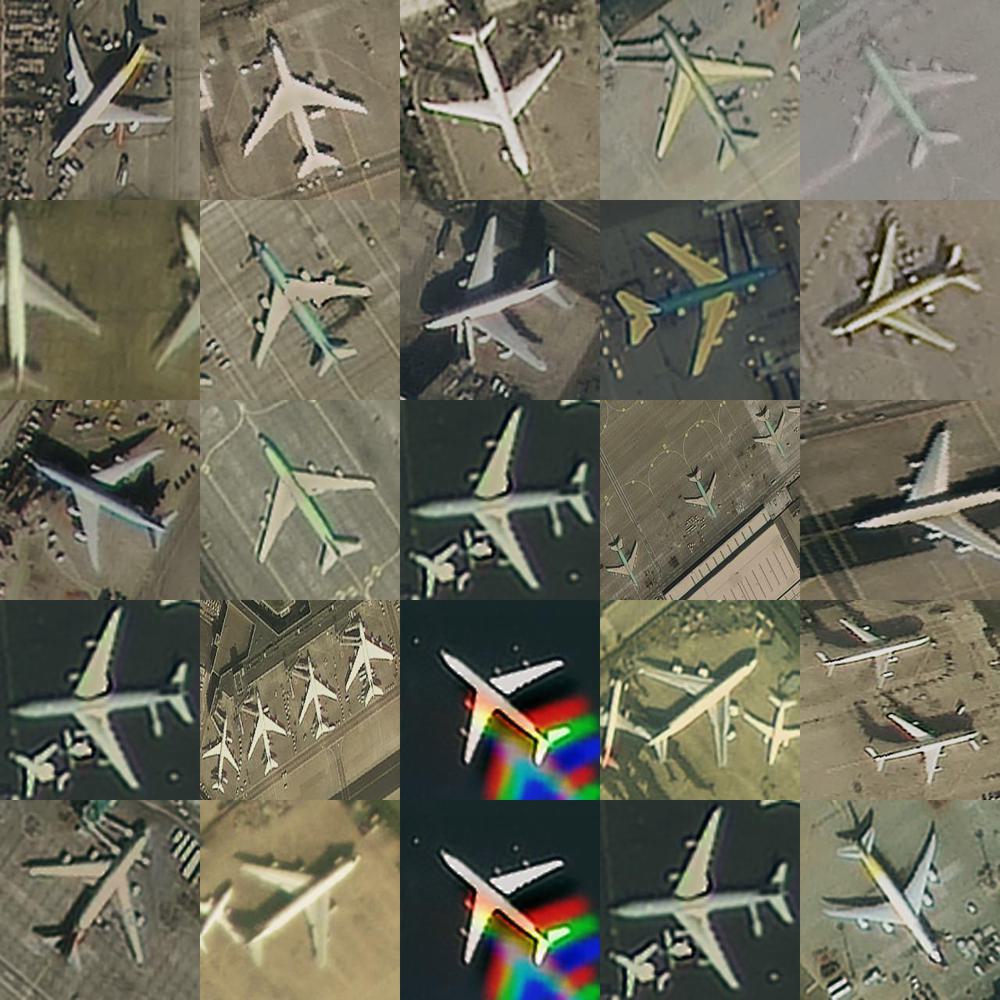}
\caption{Boeing747}
\label{fig:a}
\end{subfigure}
\begin{subfigure}{0.24\linewidth}
\centering
\includegraphics[scale=0.11]{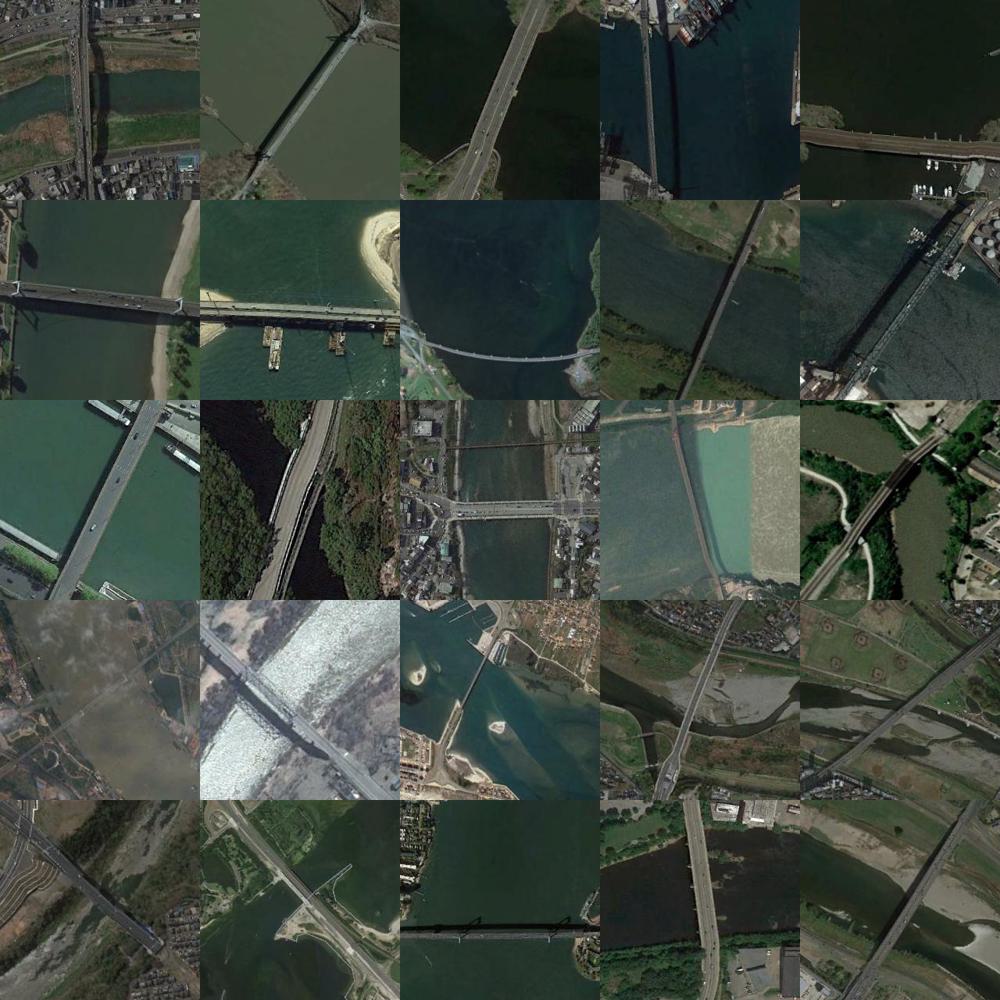}
\caption{Bridge}
\label{fig:a}
\end{subfigure}

\begin{subfigure}{0.24\linewidth}
\centering
\includegraphics[scale=0.11]{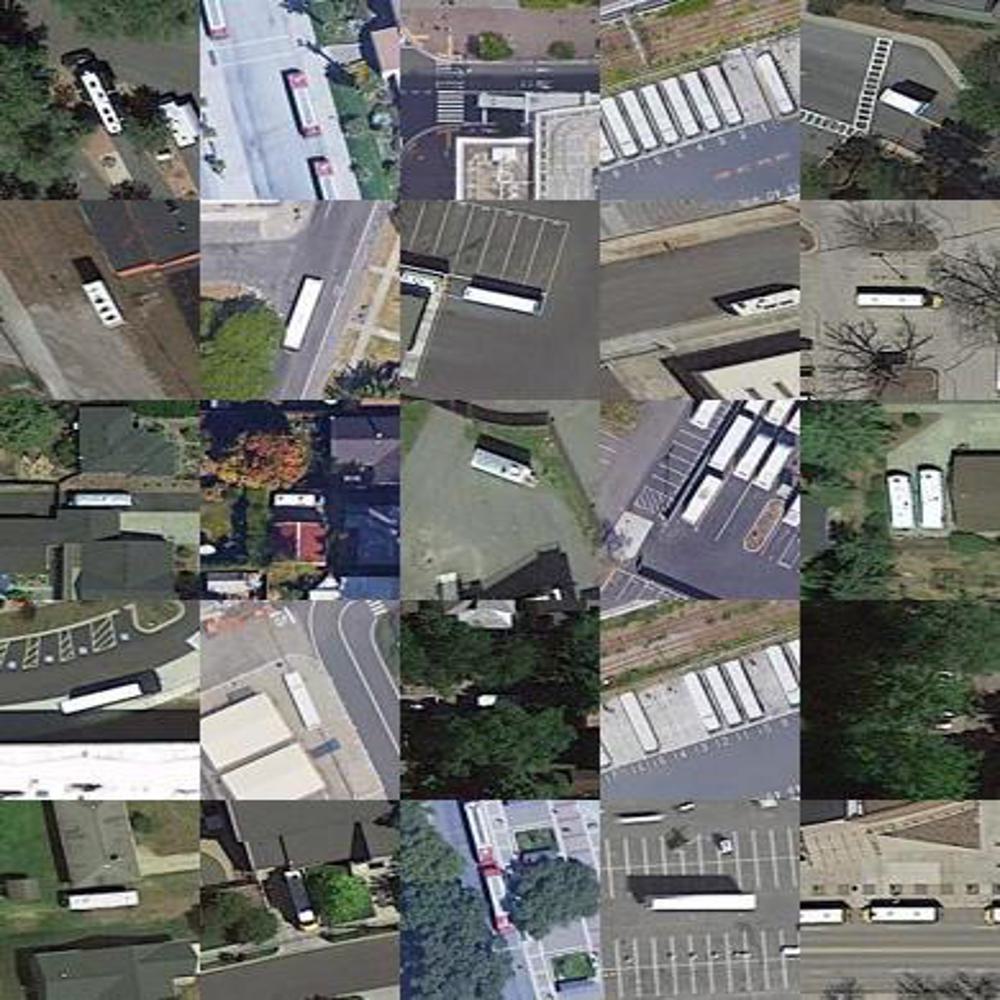}
\caption{Bus}
\label{fig:a}
\end{subfigure}
\begin{subfigure}{0.24\linewidth}
\centering
\includegraphics[scale=0.11]{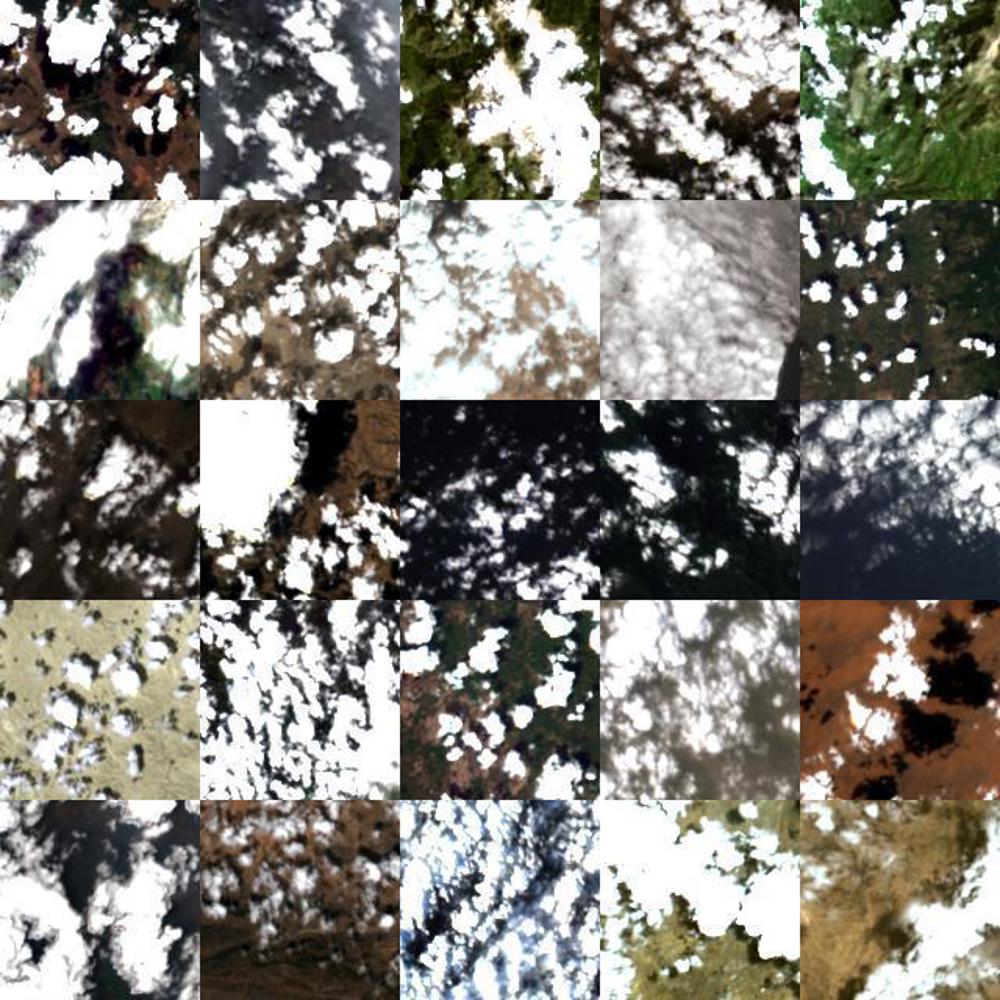}
\caption{Cloud}
\label{fig:a}
\end{subfigure}
\begin{subfigure}{0.24\linewidth}
\centering
\includegraphics[scale=0.11]{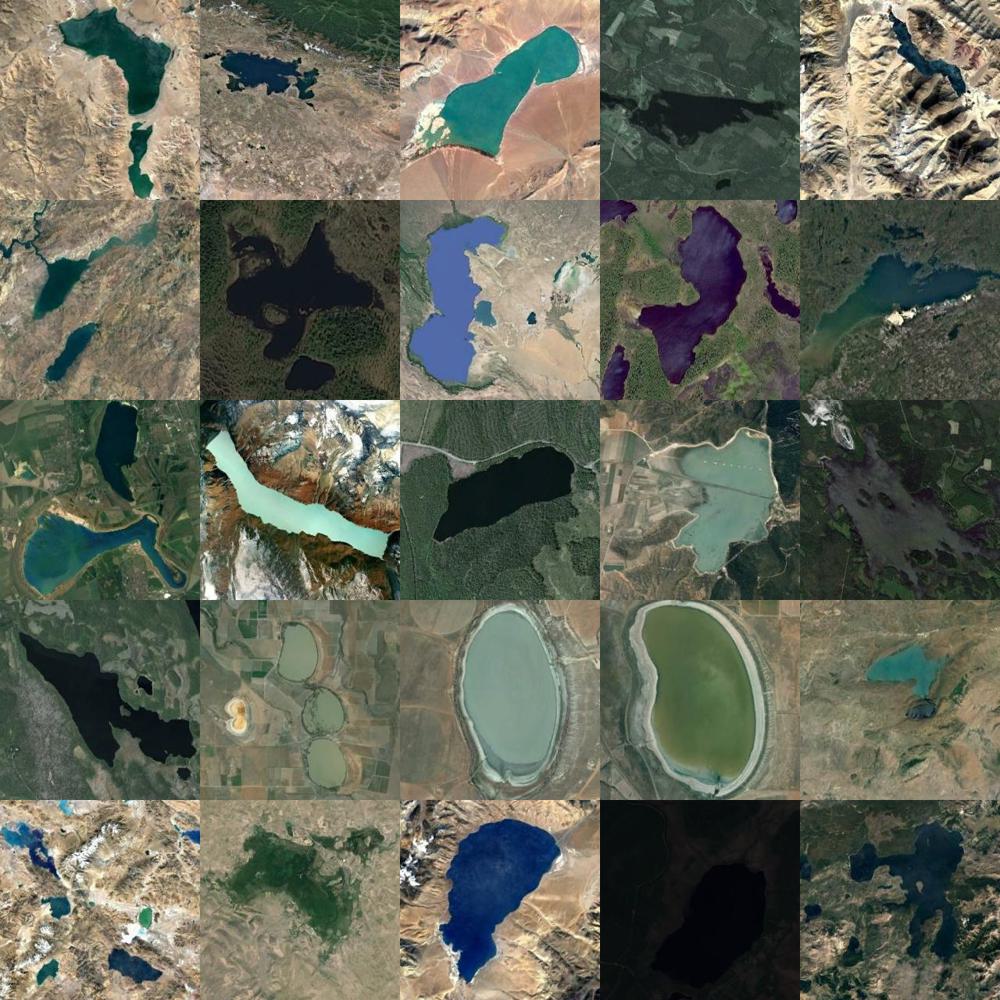}
\caption{Lake}
\label{fig:a}
\end{subfigure}
\begin{subfigure}{0.24\linewidth}
\centering
\includegraphics[scale=0.11]{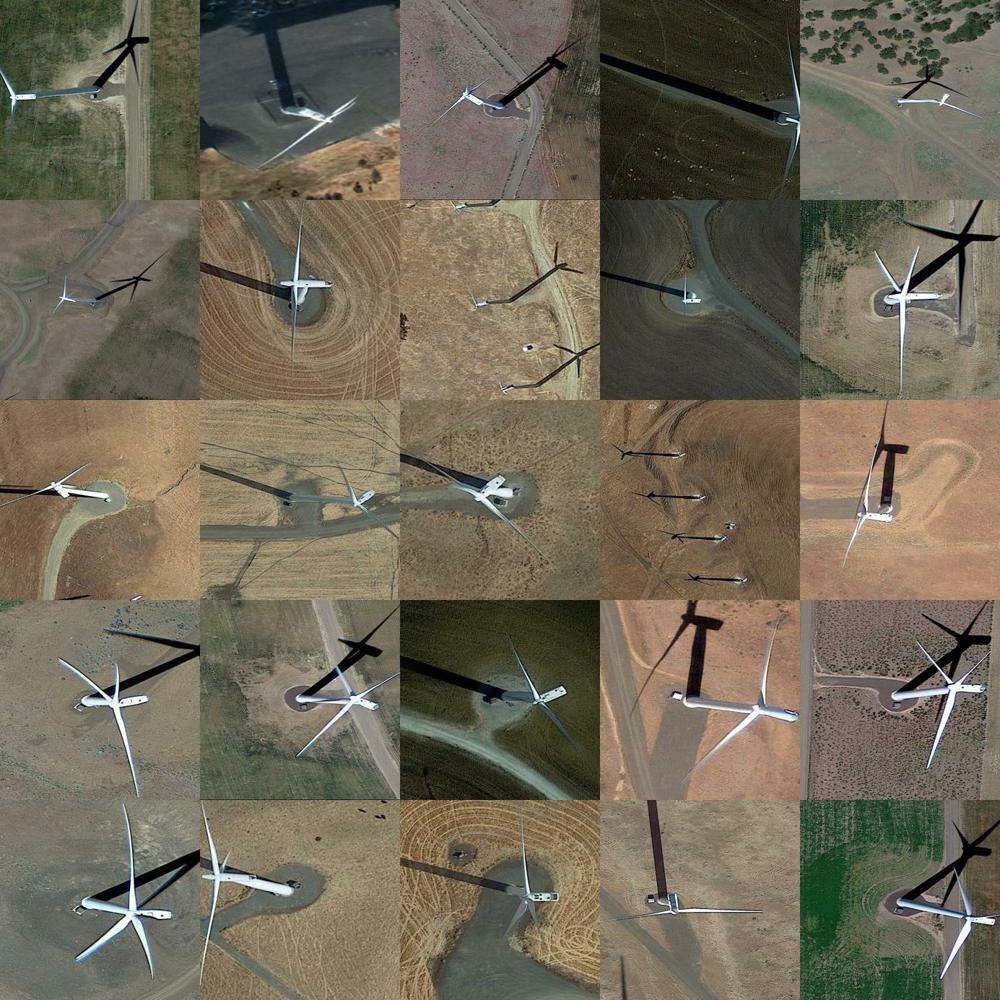}
\caption{Wind turbine}
\label{fig:a}
\end{subfigure}

\caption{Example images of sub-dataset 1 RGB.}
\label{fig:sub1}
\end{figure*}

\begin{figure*}[h]

\begin{subfigure}{0.24\linewidth}
\centering
\includegraphics[scale=0.11]{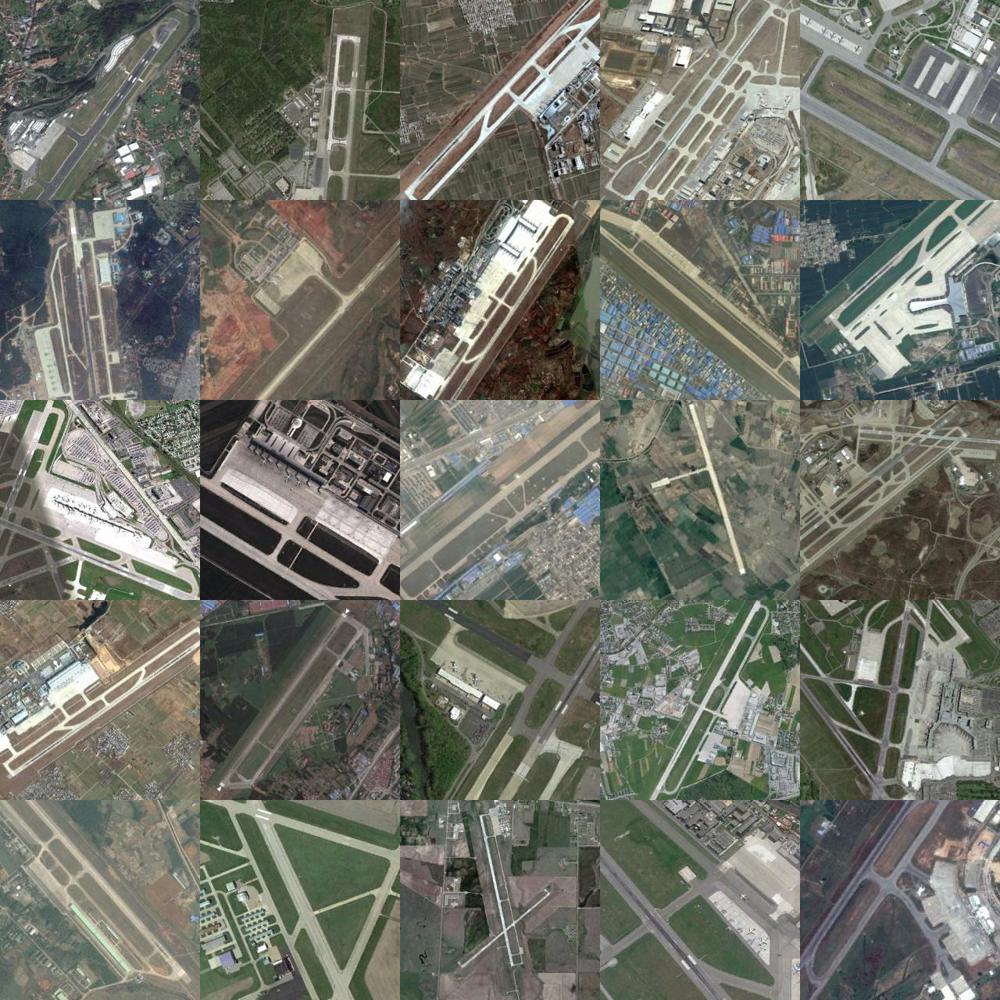}
\caption{Airport}
\label{fig:a}
\end{subfigure}
\begin{subfigure}{0.24\linewidth}
\centering
\includegraphics[scale=0.11]{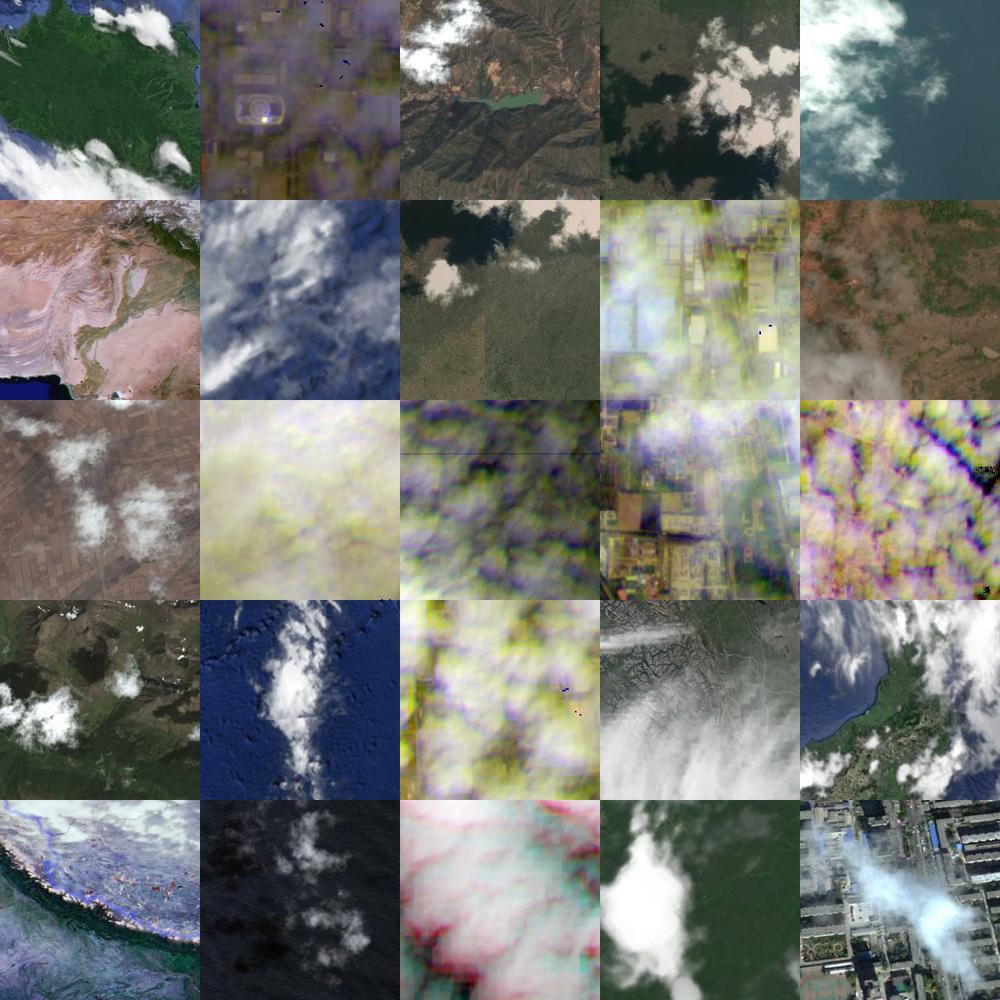}
\caption{Cloud}
\label{fig:a}
\end{subfigure}
\begin{subfigure}{0.24\linewidth}
\centering
\includegraphics[scale=0.11]{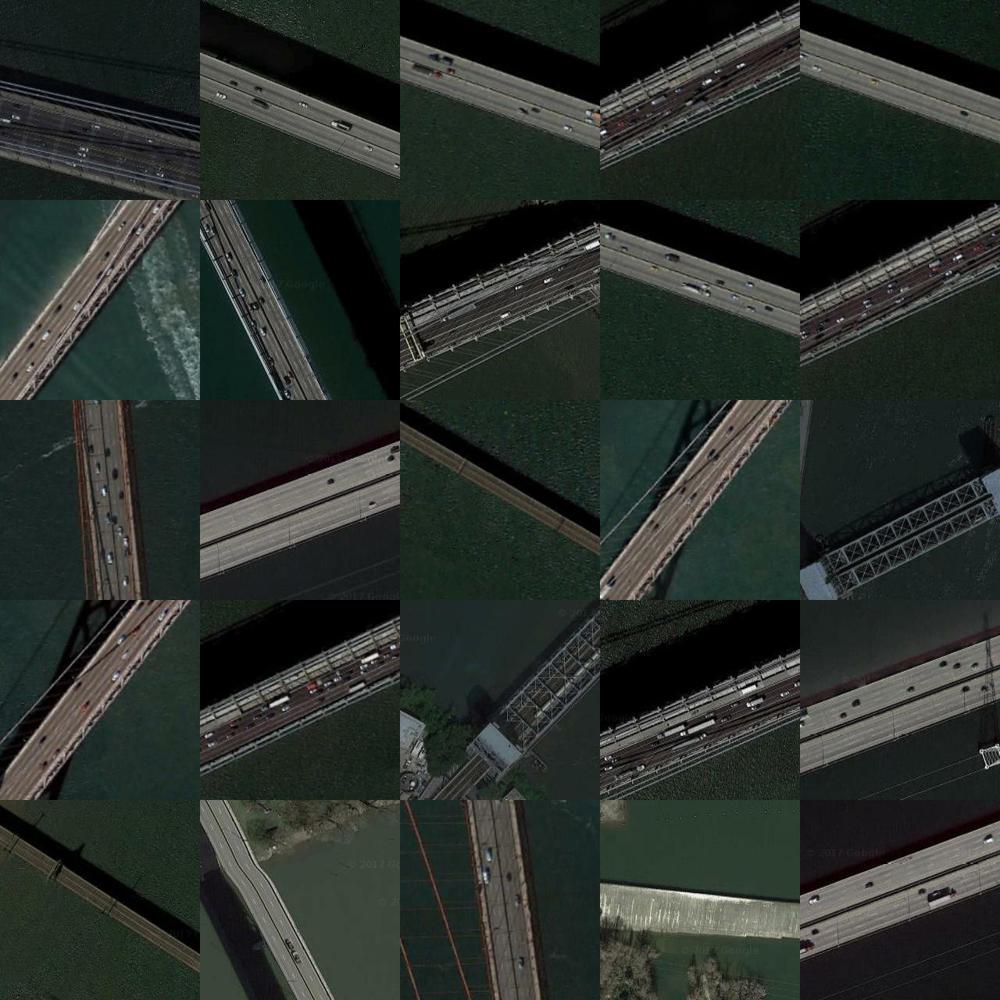}
\caption{Bridge}
\label{fig:a}
\end{subfigure}
\begin{subfigure}{0.24\linewidth}
\centering
\includegraphics[scale=0.11]{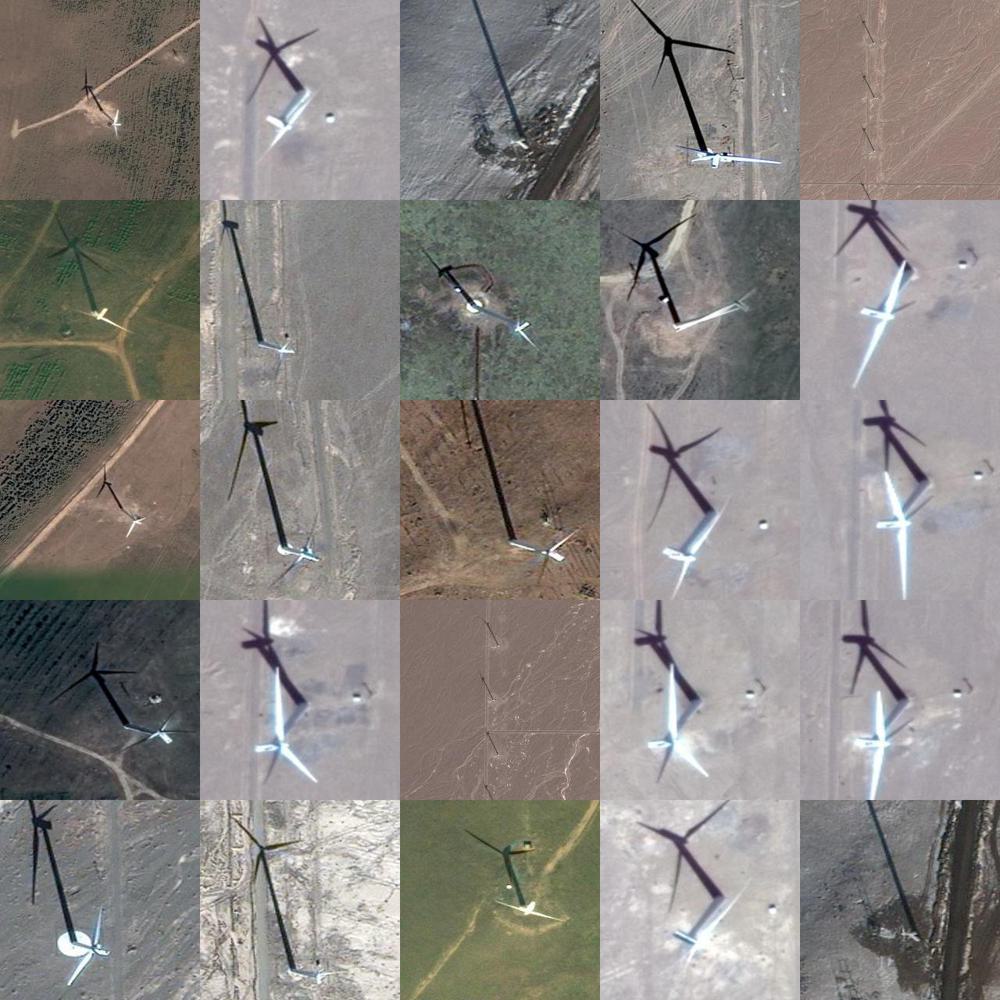}
\caption{Wind turbine}
\label{fig:a}
\end{subfigure}

\caption{Example images of sub-dataset 2 RGB.}
\label{fig:sub2}
\end{figure*}

\begin{figure*}[h]

\begin{subfigure}{0.24\linewidth}
\centering
\includegraphics[scale=0.11]{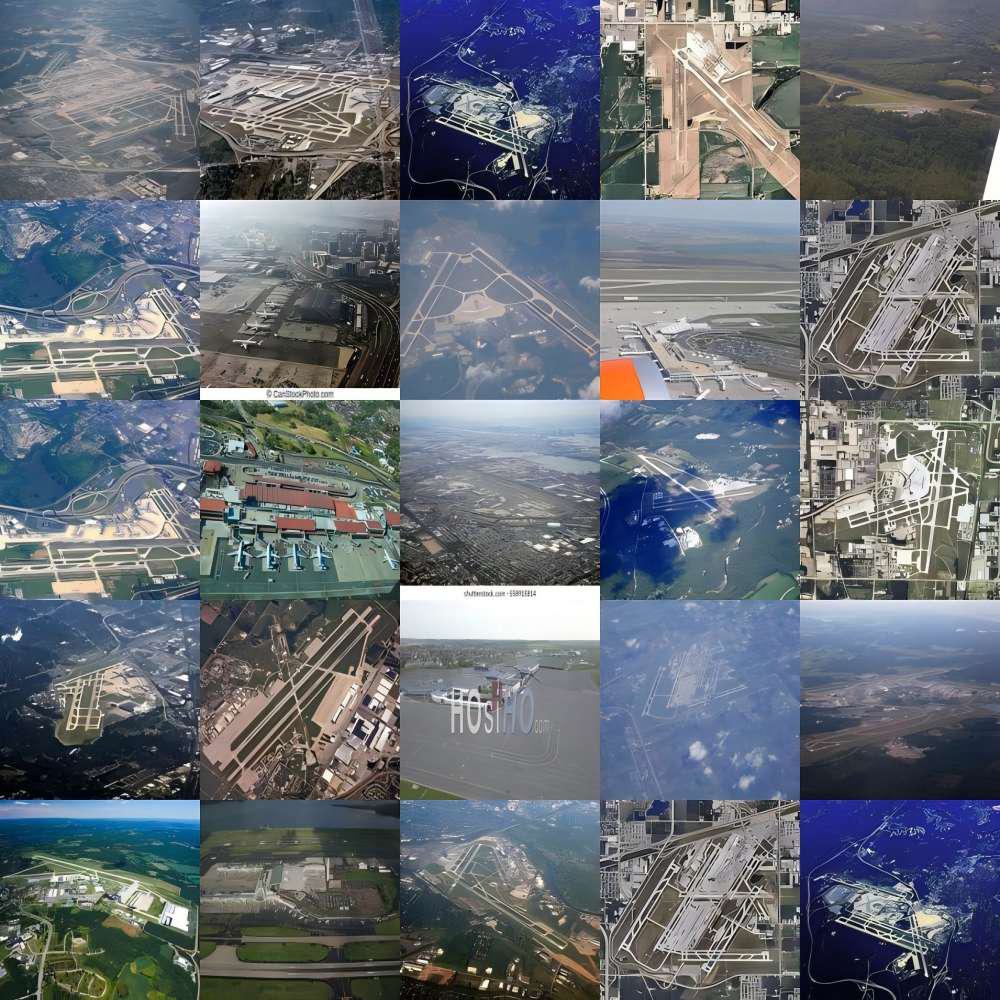}
\caption{Airport}
\label{fig:a}
\end{subfigure}
\begin{subfigure}{0.24\linewidth}
\centering
\includegraphics[scale=0.11]{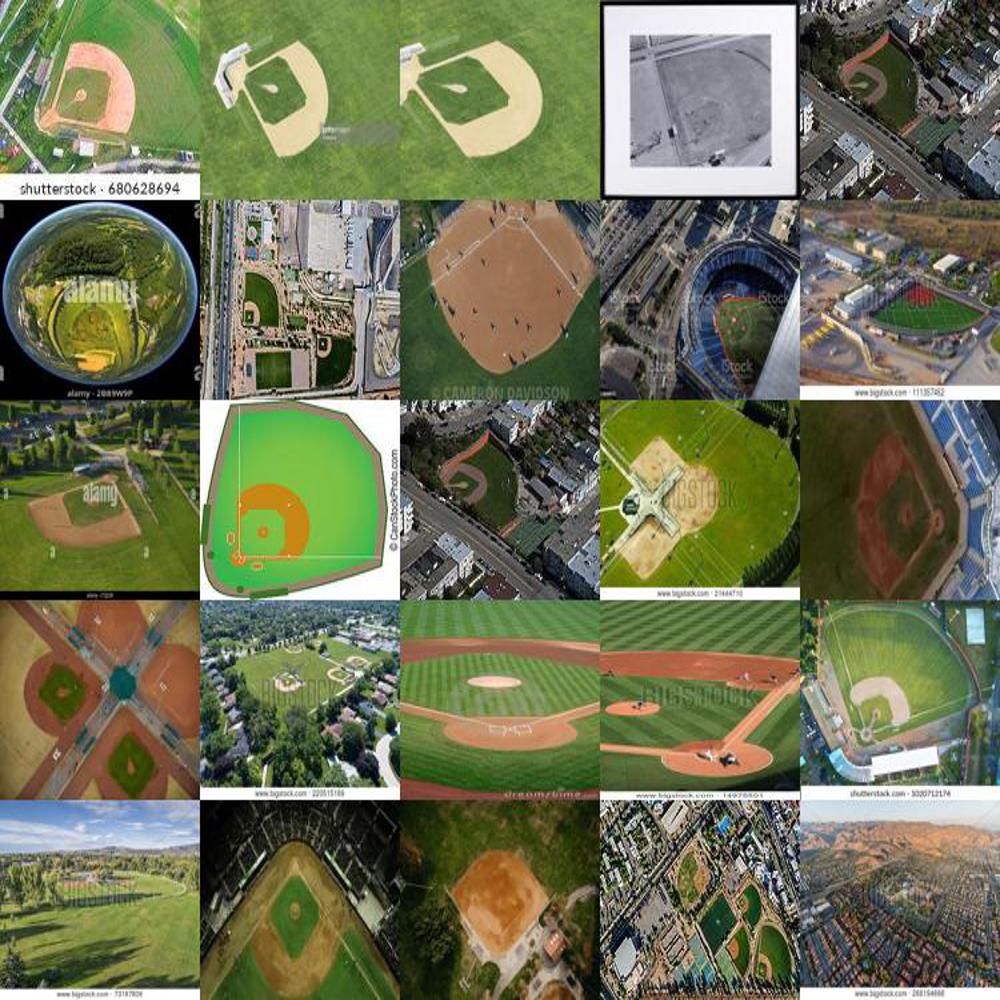}
\caption{Baseball field}
\label{fig:a}
\end{subfigure}
\begin{subfigure}{0.24\linewidth}
\centering
\includegraphics[scale=0.11]{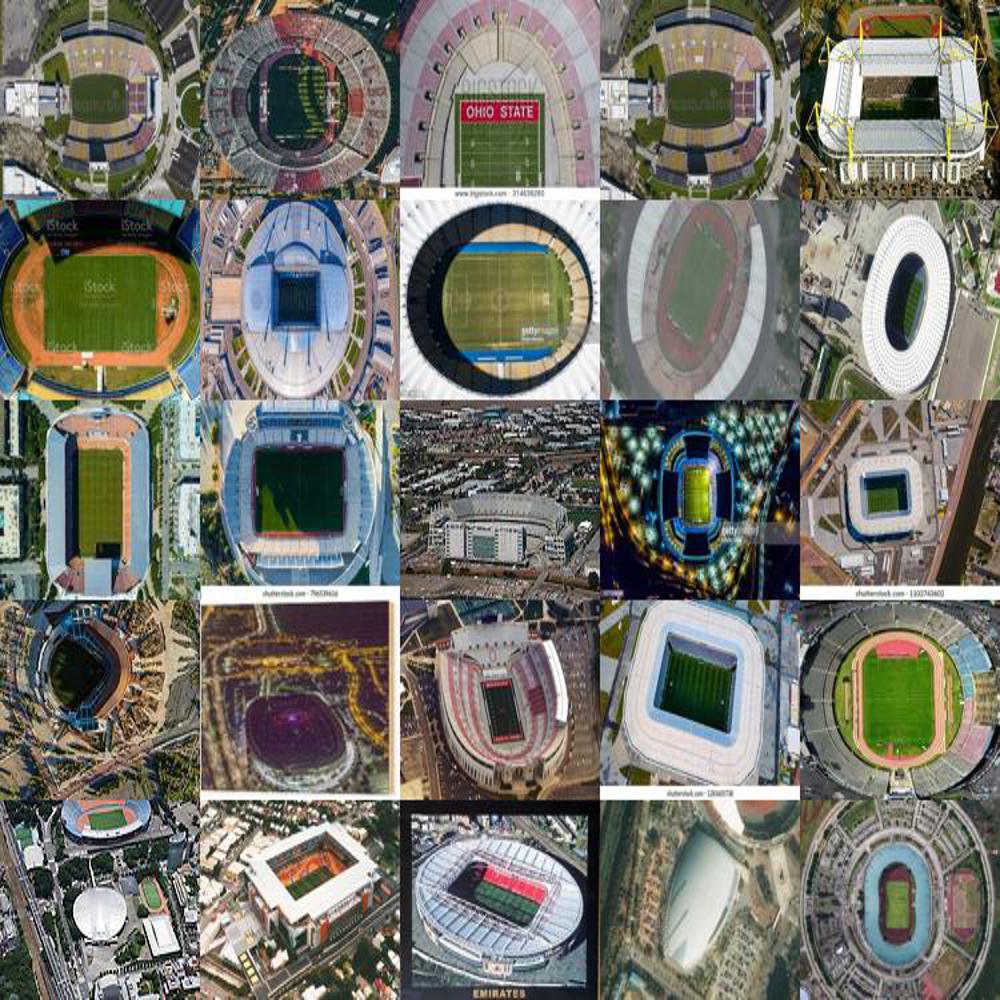}
\caption{Stadium}
\label{fig:a}
\end{subfigure}
\begin{subfigure}{0.24\linewidth}
\centering
\includegraphics[scale=0.11]{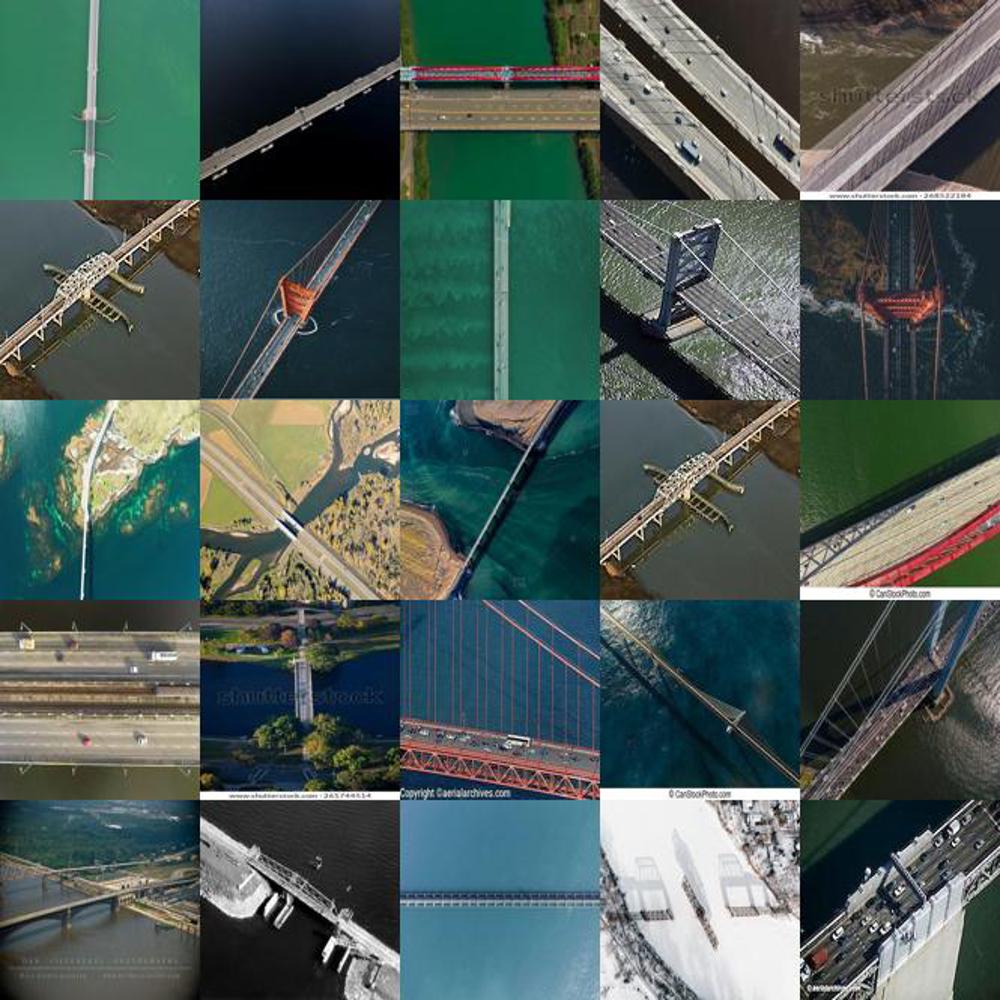}
\caption{Bridge}
\label{fig:a}
\end{subfigure}

\begin{subfigure}{0.24\linewidth}
\centering
\includegraphics[scale=0.11]{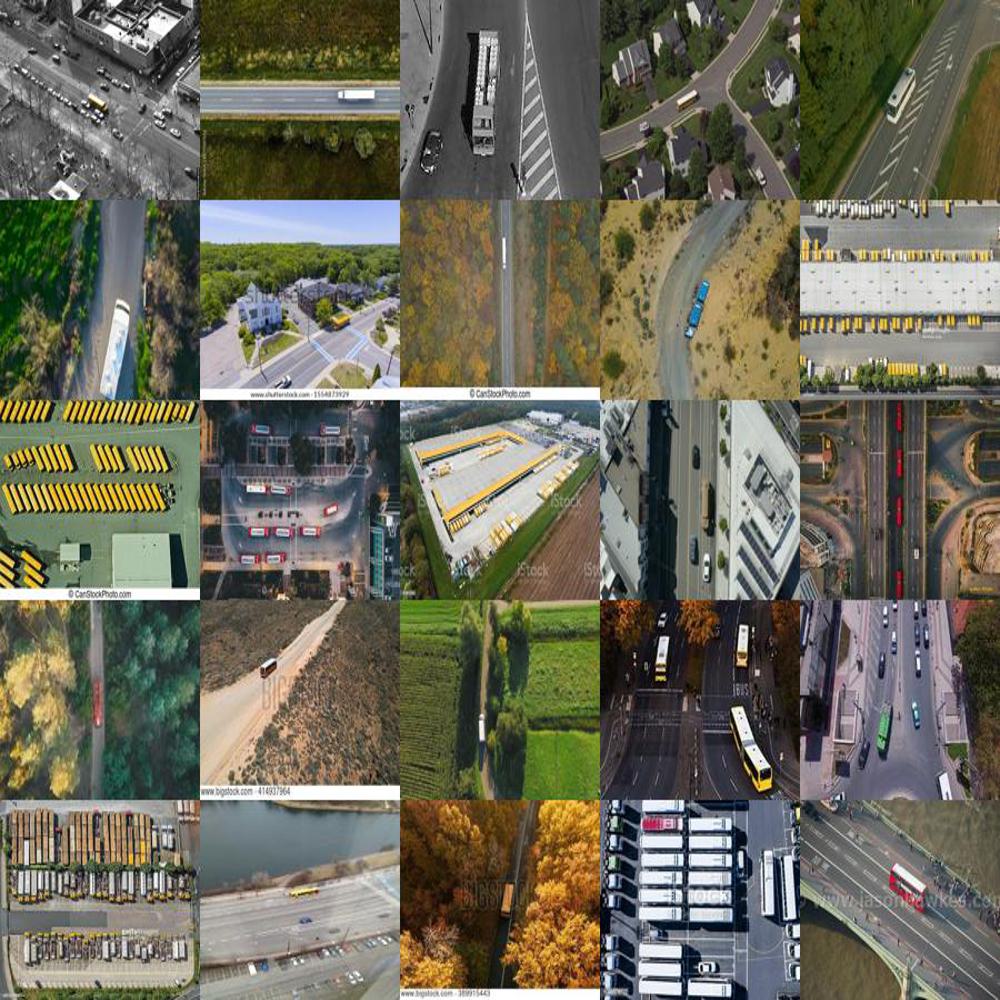}
\caption{Bus}
\label{fig:a}
\end{subfigure}
\begin{subfigure}{0.24\linewidth}
\centering
\includegraphics[scale=0.11]{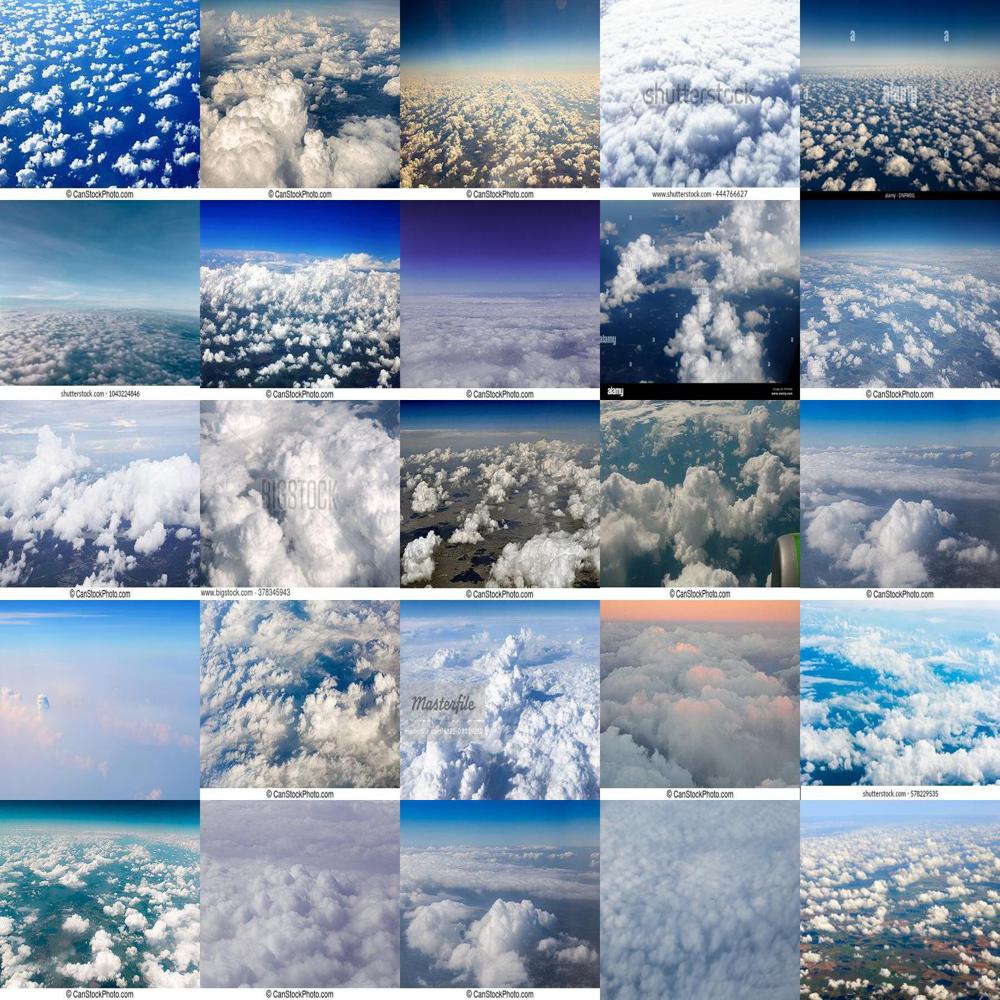}
\caption{Cloud}
\label{fig:a}
\end{subfigure}
\begin{subfigure}{0.24\linewidth}
\centering
\includegraphics[scale=0.11]{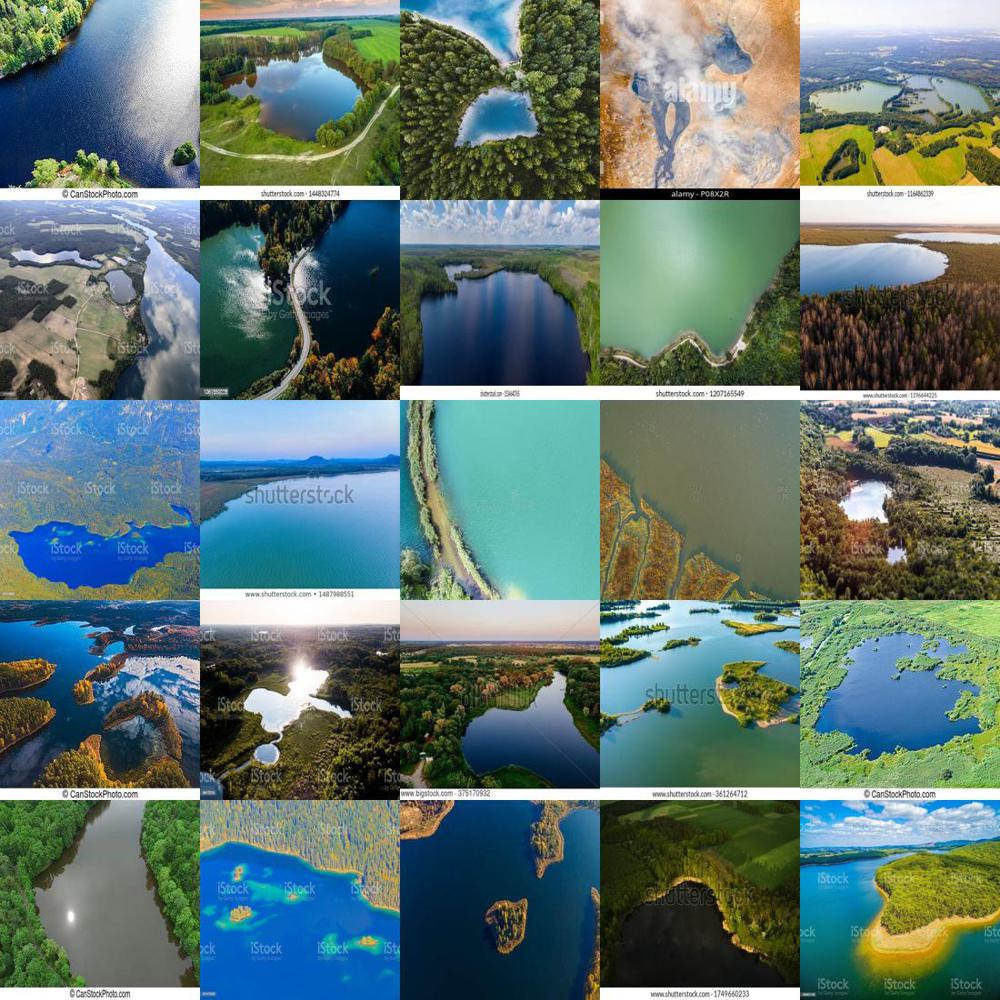}
\caption{Lake}
\label{fig:a}
\end{subfigure}
\begin{subfigure}{0.24\linewidth}
\centering
\includegraphics[scale=0.11]{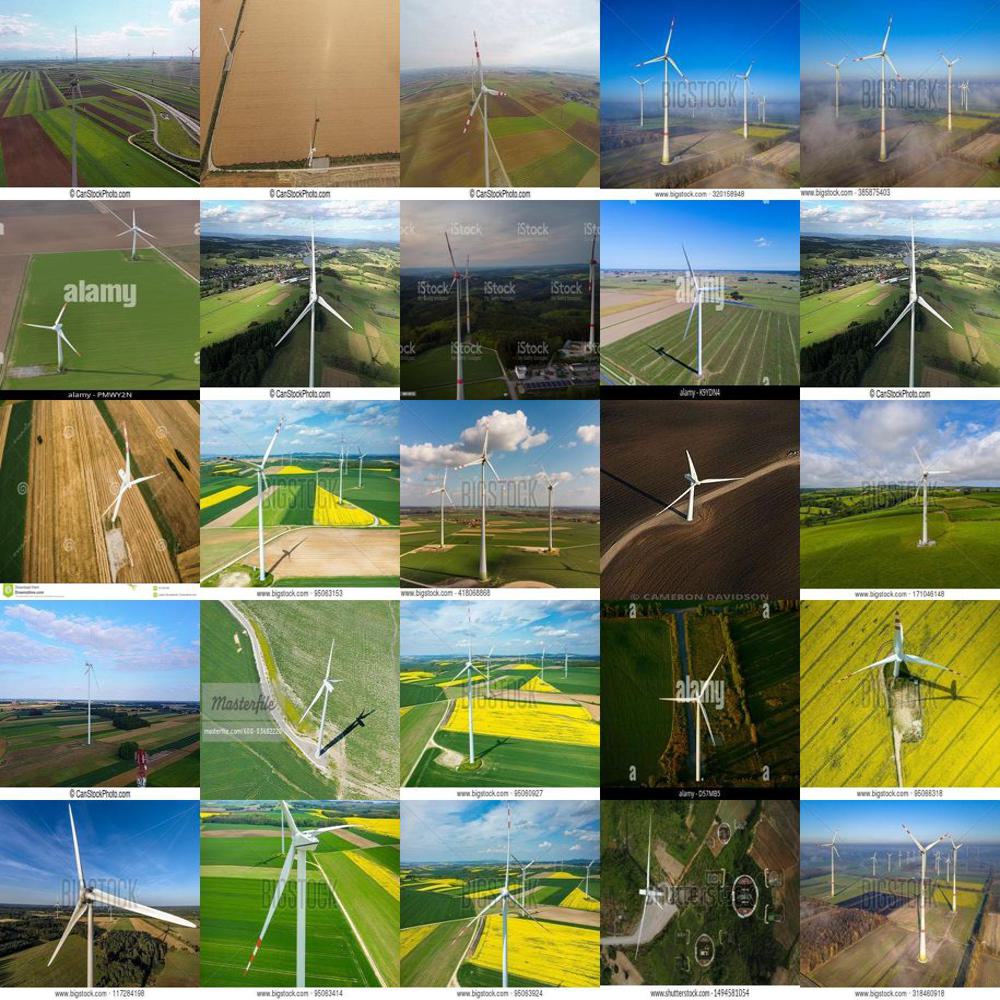}
\caption{Wind turbine}
\label{fig:a}
\end{subfigure}

\caption{Example images of sub-dataset 3 Aerial.}
\label{fig:sub3}
\end{figure*}

\begin{figure*}[h]

\begin{subfigure}{0.24\linewidth}
\centering
\includegraphics[scale=0.11]{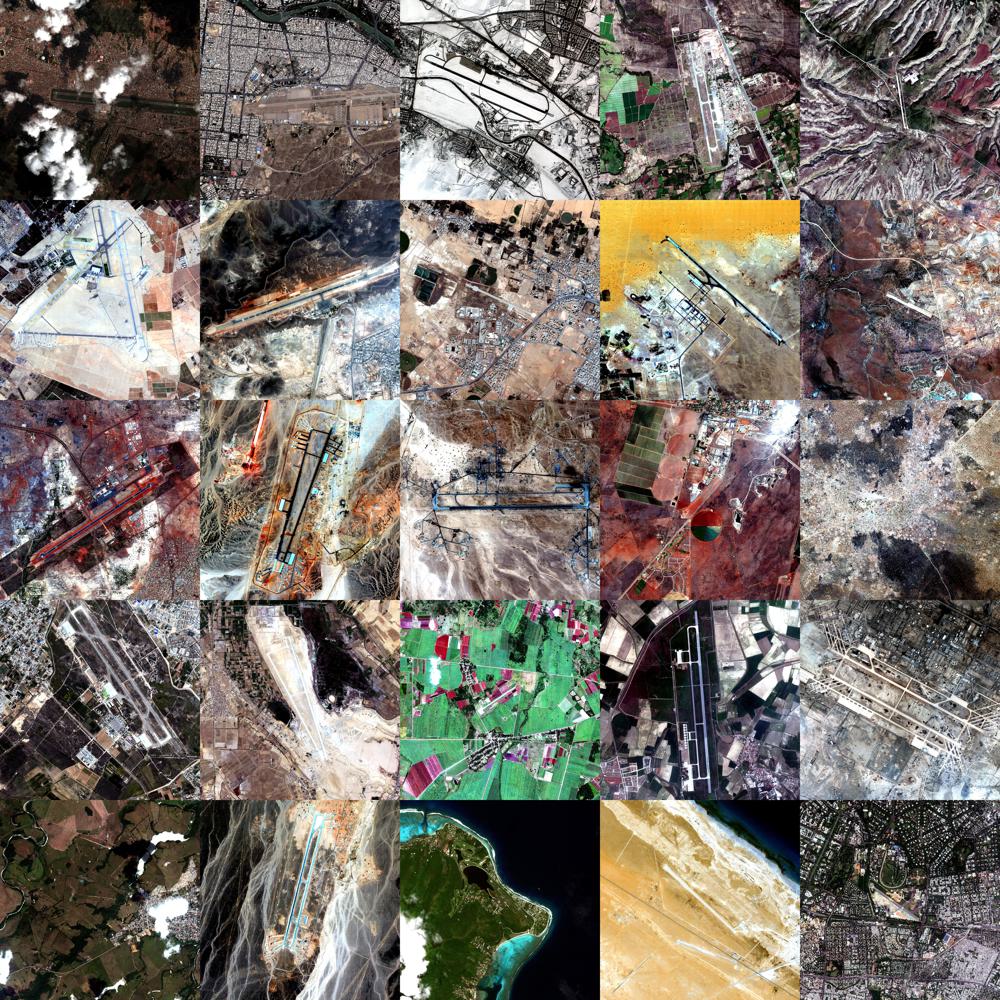}
\caption{Airport}
\label{fig:a}
\end{subfigure}
\begin{subfigure}{0.24\linewidth}
\centering
\includegraphics[scale=0.11]{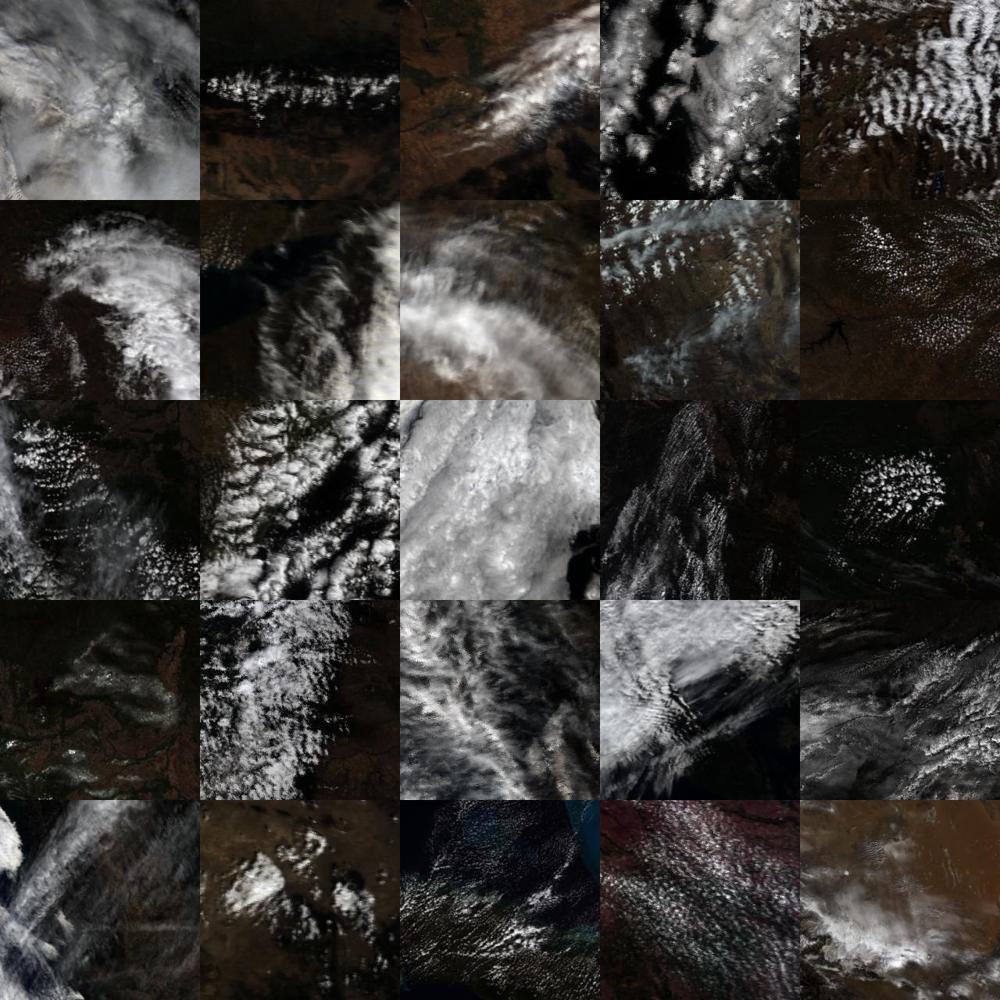}
\caption{Cloud}
\label{fig:a}
\end{subfigure}
\begin{subfigure}{0.24\linewidth}
\centering
\includegraphics[scale=0.11]{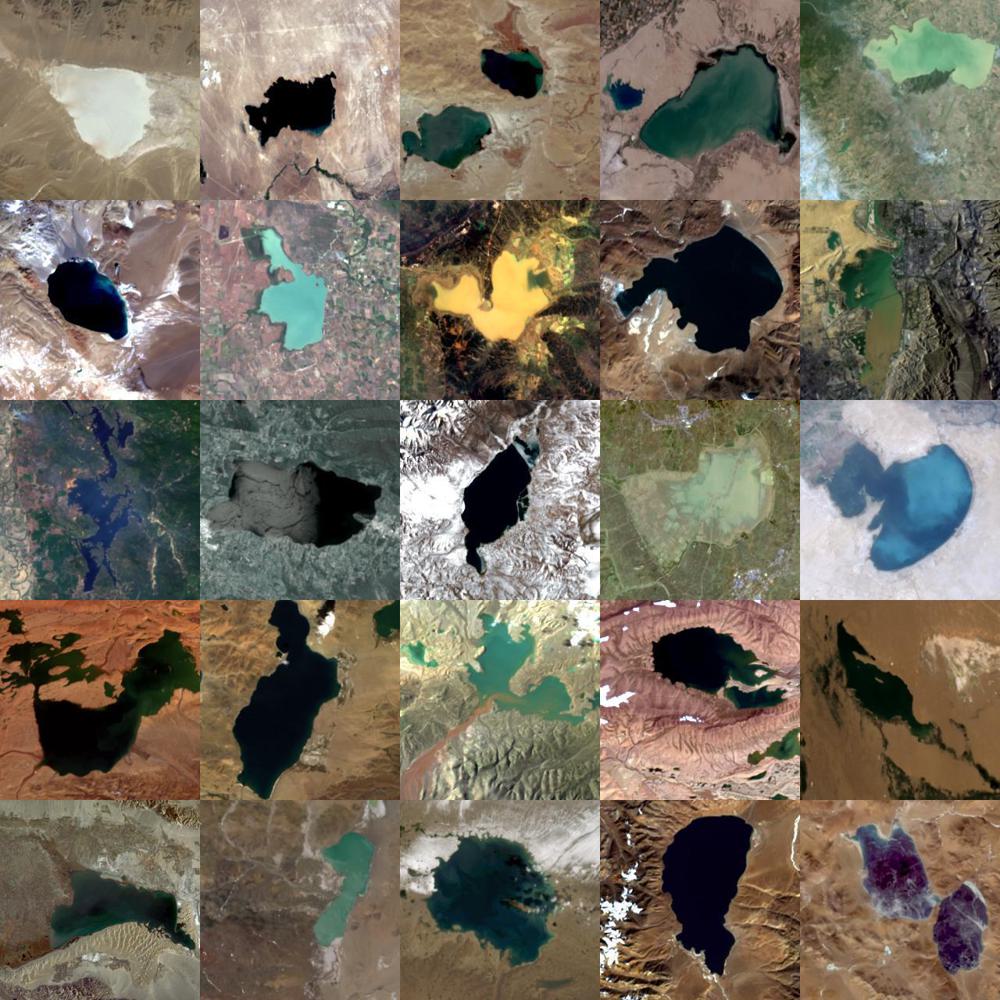}
\caption{Lake}
\label{fig:a}
\end{subfigure}
\begin{subfigure}{0.24\linewidth}
\centering
\includegraphics[scale=0.11]{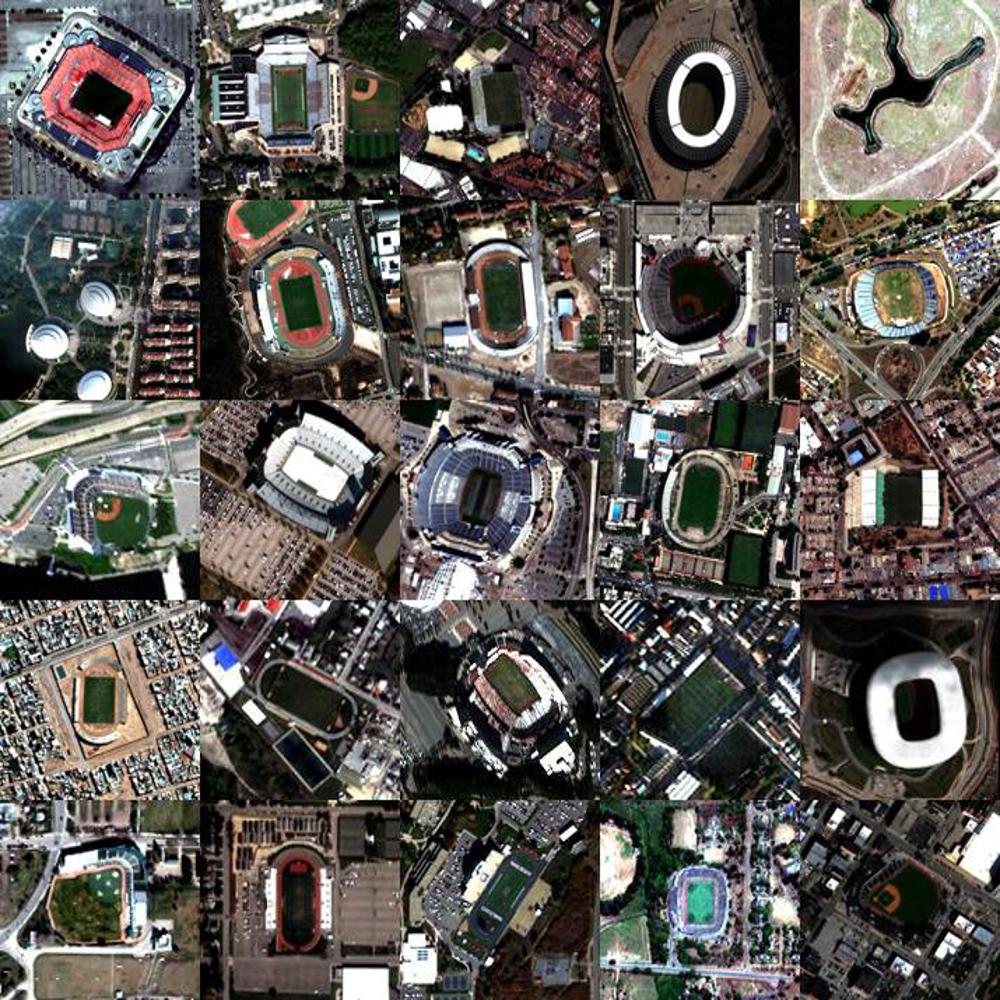}
\caption{Stadium}
\label{fig:a}

\end{subfigure}

\caption{Example images of sub-dataset 4 MSRGB.}
\label{fig:sub4}
\end{figure*}

\begin{figure*}[h]

\begin{subfigure}{0.24\linewidth}
\centering
\includegraphics[scale=0.11]{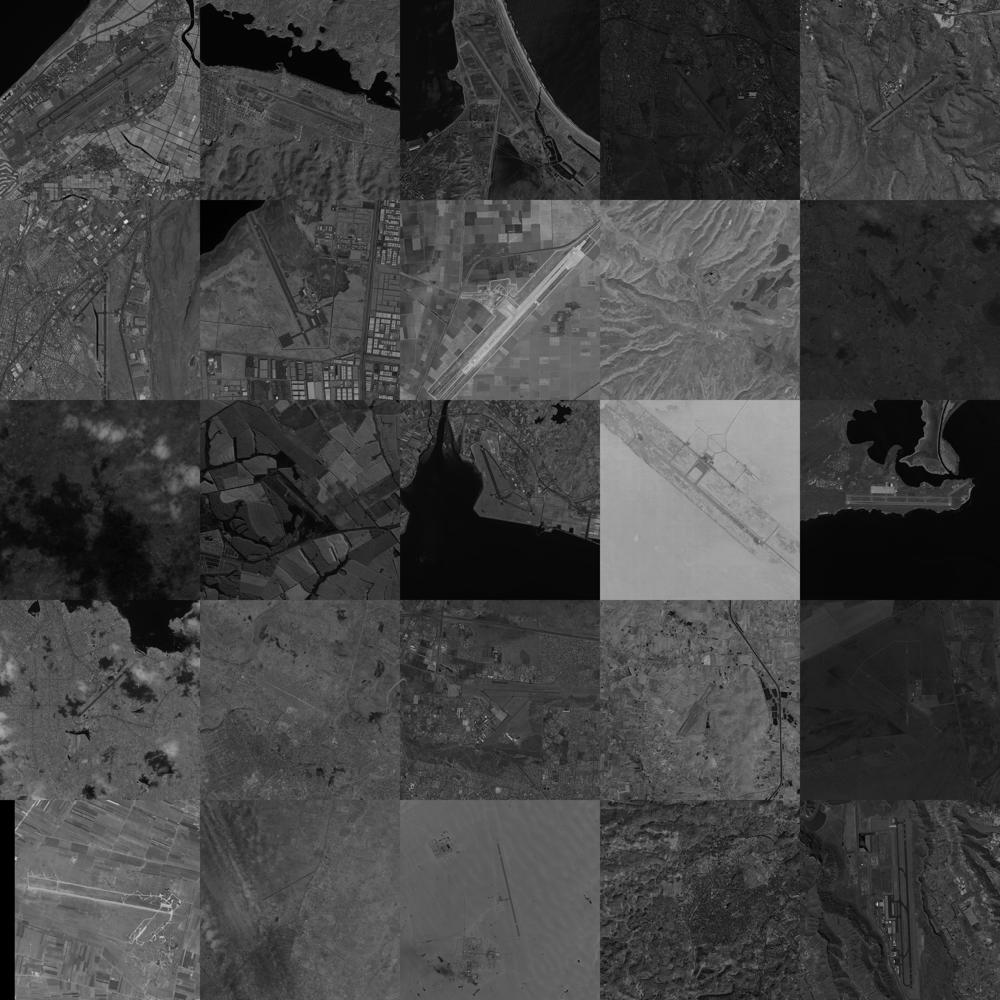}
\caption{Airport}
\label{fig:a}
\end{subfigure}
\begin{subfigure}{0.24\linewidth}
\centering
\includegraphics[scale=0.11]{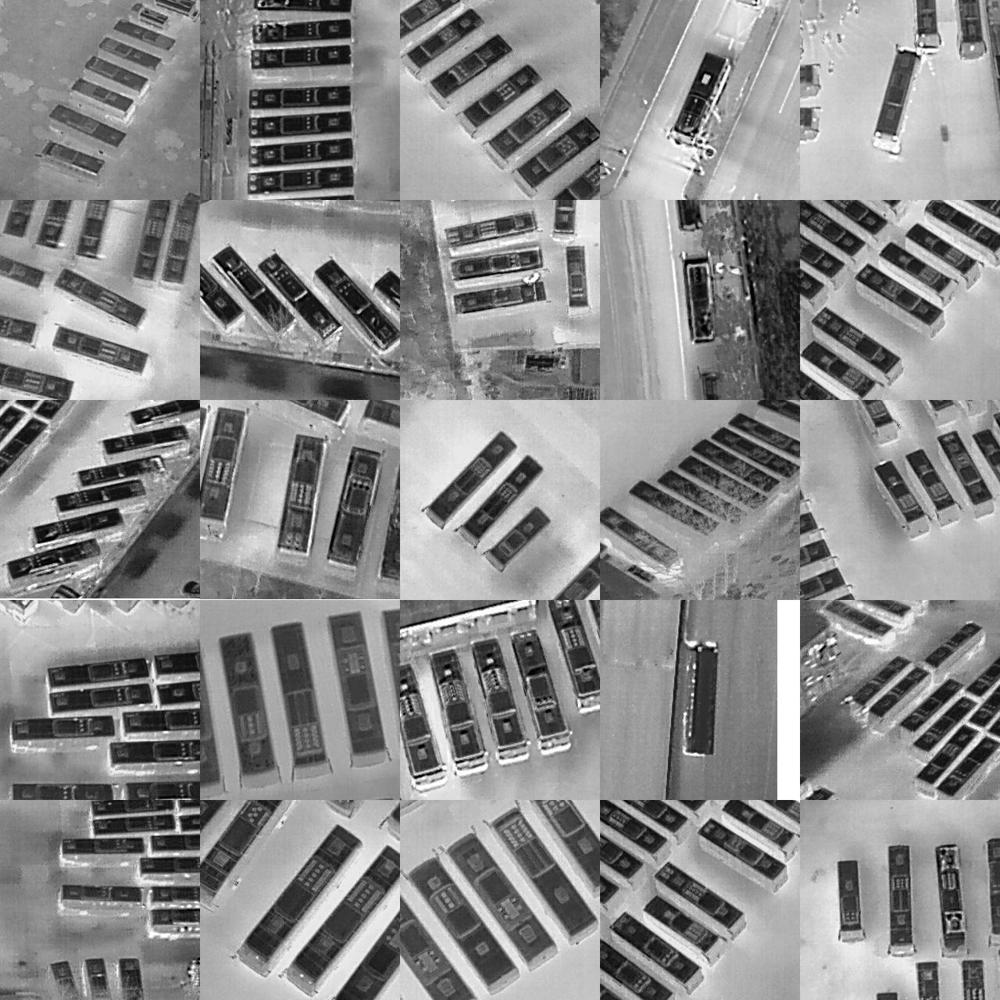}
\caption{Bus}
\label{fig:a}
\end{subfigure}
\begin{subfigure}{0.24\linewidth}
\centering
\includegraphics[scale=0.11]{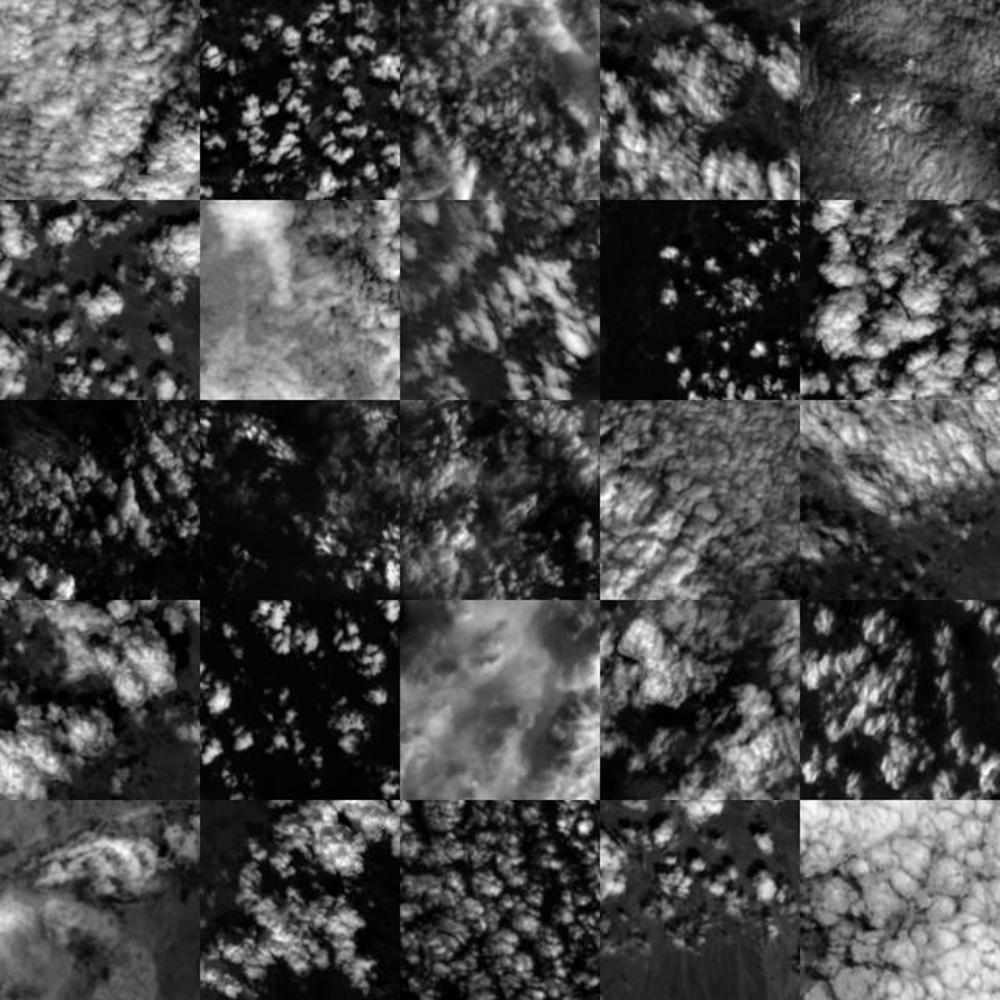}
\caption{Cloud}
\label{fig:a}
\end{subfigure}
\begin{subfigure}{0.24\linewidth}
\centering
\includegraphics[scale=0.11]{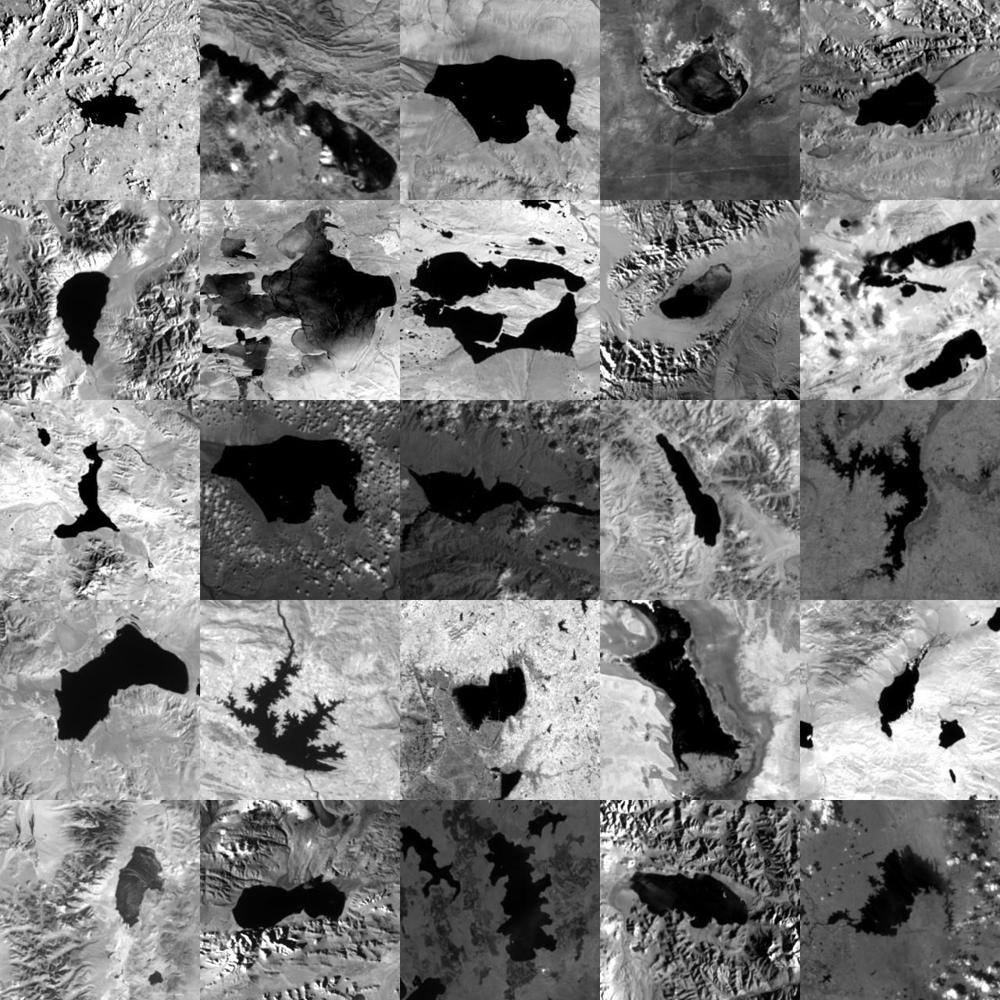}
\caption{Lake}
\label{fig:a}
\end{subfigure}

\caption{Example images of sub-dataset 5 IR.}
\label{fig:sub5}
\end{figure*}

Fig.~\ref{fig:sub1},~\ref{fig:sub2}, ~\ref{fig:sub3},~\ref{fig:sub4}, ~\ref{fig:sub5} showcase images from Sub-datasets 1, 2, 3, 4, and 5, respectively, highlighting the differences across the various domains in OES.

\section{Evaluation details}
\subsection{Close-set Classification } \label{sec:Close-set Classification}
We evaluate the closed-set classification performance of models under different architectures and finetuning methods in $\mathcal{D}_{R1}^{all}$, $\mathcal{D}_{R3}^{all}$, $\mathcal{D}_{R4}^{all}$ and $\mathcal{D}_{R5}^{all}$. Tab.~\ref{table:close} presents detailed closed-set classification results for various models on the OES dataset. For single-modal visual models, all models are initialized with ImageNet-1k pre-trained weights and further trained for 100 epochs using SGD optimizer (momentum: 0.9, weight decay: 0.0005), with a batch size of 128. We test the model's performance under different learning rates and ultimately set it to 0.01 to achieve optimal average performance. For data augmentation, we perform the following pipeline: (1) \textbf{Resize}: size = (256, 256), (2) \textbf{CenterCrop}: size = (224, 224), (3) \textbf{RandomHorizontalFlip}: p = 0.5, (4) \textbf{RandomCrop}: size = (224, 224), padding=4, (5) \textbf{ToTensor}, (6) \textbf{Normalize}.
The training for single-modal visual models is conducted on a single NVIDIA RTX 4090 GPU. 

For vision-language models (VLMs), we select CLIP as the representative model and evaluate different finetuning approaches, including text prompt tuning (TPT) and visual adaptation (VT). For TPT, we follow CoOp \cite{zhou2022learning} and insert 16 learnable class-specific prompts with 'end' class token position.
For VT, we follow DPM \cite{2024ECCV} and insert two projection modules after the vision encoder: one for refining region-level visual features, consisting of a 1×1 convolutional layer, Group Normalization, ReLU activation, another 1×1 convolutional layer, and Group Normalization; and another for refining global visual features, comprising a linear layer (512, 512), Layer Normalization, ReLU activation, a linear layer (512, 512), and Layer Normalization. For both VT and TPT, we conduct finetuning based on OpenAI's pretrained CLIP model \cite{radford2021learning} and GeoRSCLIP \cite{zhang2024rs5m} model with the following uniform configurations: a batch size of 512, SGD optimizer with an initial learning rate of 0.01 (decayed via cosine schedule over 20 epochs), and a warmup strategy (fixed lr=1e-4 for the first epoch). The data preprocessing pipeline is intentionally minimal, comprising only: (1) \textbf{Resize}: size = (224, 224), (2) \textbf{ToTensor}, and (3)  \textbf{Normalize}. The training for TPT and VT is conducted on one and two NVIDIA RTX 4090 GPUs, respectively.

\subsection{Zero-shot Classification}

We use CLIP as the representative model for VLMs and employ the standardized prompt template 'a photo of a \{cls\}' for text input processing.

\subsection{OOD Detection \& Generalization}

\subsubsection{Post-hoc OOD Detection}

For post-hoc OOD detection, we evaluate all methods using the same model trained on $\mathcal{D}_{R1}^{id}$, denoted as $\mathcal{M}_{R1}^{id}$. Specifically, all models are trained on the $\mathcal{D}_{R1}^{id}$ training set using the hyperparameters and training strategies specified in Section B.1. Method-specific hyperparameter settings will be detailed subsequently:

\textbf{OpenMax} \cite{openmax16cvpr}: We perform score recalibration on the top $\alpha=3$ classes, while the remaining classes are left unchanged. The distance metric is defined as a weighted sum of the Euclidean distance ($w_{e}=0.5$) and the cosine distance ($w_{c}=1$). For the Weibull distribution fitting, we use the largest $\eta=20$ distances as the tail size, and set the OOD threshold to $\epsilon=0.9$.

\textbf{MSP} \cite{hendrycks17baseline}: We set the temperature $\tau=1$ in the softmax function to calculate the confidence score.

\textbf{ODIN} \cite{odin18iclr}: We perform a grid search over the temperature parameter $\tau$ in $[1,10,100,1000]$ and the perturbation magnitude $\epsilon$ in $[0.0014,0.0028]$ to optimize the OOD detection performance. Then we use msp score as the OOD score.

\textbf{MDS} \cite{mahananobis18nips}: This method computes the Mahalanobis distance between input features and class-conditional Gaussian distributions as the OOD score, and does not require hyperparameter tuning.

\textbf{GradNorm} \cite{huang2021importance}: This method uses the norm of loss gradients as the OOD score, and does not require hyperparameter tuning.

\textbf{ReAct} \cite{sun2021react}: We search over the percentile $p$ in $[85,90,95,99]$ to determine the truncation threshold $c$, such that activation values above $c$ are clipped during inference. Then we use msp score as the OOD score.

\textbf{MLS} \cite{species22icml}: This method uses max logits as the OOD score and does not require hyperparameter tuning.

\textbf{KLM} \cite{species22icml}: This method uses Kullback-Leibler divergence as the OOD score and does not require hyperparameter tuning.

\textbf{ViM} \cite{wang2022vim}: We search over the feature subspace dimension $N$ in $[256, 1000]$ for principal component projection in the ViM score calculation. Then we use the ViM score as the OOD score.

\textbf{KNN} \cite{sun2022knnood}: We search over the nearest neighbors $K$ in $[50,100,200,500,1000]$ to compute the distance-based OOD score within the feature space.

\textbf{EBO} \cite{liu2020energy}: We set the temperature parameter $\tau=1$ in the calculation of the Helmholtz free energy, which serves as the OOD score and is commonly referred to as the energy score.

\textbf{ASH} \cite{djurisic2023extremely}: We search over the percentile $p$ in $[65,70,75,80,85,90,95]$ to determine the activation threshold. Activations below or equal to this threshold are pruned, while those above the threshold are scaled accordingly. Then we use the energy score as the OOD score.

\textbf{DICE} \cite{sun2021dice}: We set the sparsity parameter $p=90$, which determines the threshold for masking the weights. A higher value of $p$ results in a greater fraction of weights being pruned. When $p=0$, the output is equivalent to the original dense transformation. Then we use the energy score as the OOD score.

\textbf{Relation} \cite{kim2023neural}: We search over the power $p$ in $[1,2,4,6,8]$ to control the sharpness of the kernel value distribution in the relation graph. We use a chunk size of 50 for batch-wise kernel computation and set the threshold for relation values to 0.03, below which relations are ignored. Then we use the relation score as the OOD score.

\textbf{FDBD} \cite{liu2024fast}: This method uses feature distances to decision boundaries and does not require hyperparameter tuning.

\textbf{GEN} \cite{cvpr2023gen}: We perform a grid search over the power of generalized entropy $\gamma$ in $[0.01, 0.1, 0.5, 1, 2, 5, 10]$, which adjusts the sensitivity of the entropy measure, and the number of top classes $M$ in $[10, 50, 100]$, which reduces noise from negligible tail probabilities. Then we use the generalized entropy score as the OOD score.

\textbf{Rankfeat} \cite{song2022rankfeat}: We set the temperature factor $\tau=1$ to calculate the energy score as the OOD score. The logits used to compute the energy score are obtained by averaging the logits from SVD-processed features of different layers.

\textbf{RMDS} \cite{rmd21arxiv}: This method uses relative Mahalanobis distance and does not require hyperparameter tuning.

\textbf{Gram} \cite{gram20icml}: We set the power list to $[1, 2, 3, 4, 5]$ to compute the $p$-th order Gram matrices for each feature layer. The OOD score is then calculated by aggregating the normalized deviations of these Gram features from their corresponding minimum and maximum values estimated on the training set. 

\textbf{NNGuide}  \cite{park2023nearest},: We set the sampling ratio $\alpha=0.01$ and the number of nearest neighbors $K=100$. The OOD score is computed by multiplying the mean similarity to the $K$ nearest neighbors in the feature bank  with the energy score.

\textbf{Scale} \cite{iclr2024scale}: We set the percentile $p=85$ to determine the scale factor $r$. Then we use the energy score as the OOD score.

\textbf{SHE} \cite{she23iclr}: We choose inner product to calculate the simplified Hopfield energy score as the OOD score.

\textbf{MDSE} \cite{mahananobis18nips}: We set the noise magnitude $\epsilon=0.0014$ and weights of logistic regression detector $\alpha=1$ for combining Mahalanobis scores from different layers. The OOD score is obtained by a weighted sum of the Mahalanobis distances across multiple feature layers.

\subsubsection{Training-required OOD Detection}

For training-required OOD detection, method-specific hyperparameter settings will be detailed subsequently:

\textbf{G-ODIN} \cite{hsu2020generalized}: We choose cosine classifier $h^{C}_{i}(x)$ after the penultimate layer and set the noise scaling factor \textit{noise magnitude} to 0.0025. We use $\mathcal{M}_{R1}^{id}$ to initialize the weights and train for 30 epochs with a batch size of 128 and a learning rate of 0.001.

\textbf{ConfBranch} \cite{confbranch2018arxiv}: We set the budget value of $\beta=0.3$, $\lambda$ for confidence loss to 0.1 and  noise perturbation $\epsilon$ to 1.0e-12. We use $\mathcal{M}_{R1}^{id}$ to initialize the weights and train for 30 epochs with a batch size of 128 and a learning rate of 0.001.

\textbf{RotPred} \cite{rotpred}: Considering the increased number of samples per batch due to RotPred's rotation-based augmentation, we reduced the batch size. We use $\mathcal{M}_{R1}^{id}$ to initialize the weights and train for 30 epochs with a batch size of 64 and a learning rate of 0.001.

\textbf{VOS} \cite{vos22iclr}: We sample 1000 virtual outliers and set the weight $\beta$ for $\mathcal{L}_{uncertainty}$ to 0.1. We use $\mathcal{M}_{R1}^{id}$ to initialize the weights and train for 30 epochs with a batch size of 128 and a learning rate of 0.001.

\textbf{LogitNorm}  \cite{wei2022mitigating}: We use $\mathcal{M}_{R1}^{id}$ to initialize the weights and train for 30 epochs with 128 batchsize, 0.001 learning rate and 0.04 temperature parameter $\tau$.

\textbf{CIDER} \cite{cider2023iclr}: We set the weight $\lambda_{c}$ for $\mathcal{L}_{comp}$ to 2, temperature in $\mathcal{L}_{comp}$ to 0.1 and prototype update factor $\alpha$ to 0.95. We use $\mathcal{M}_{R1}^{id}$ to initialize the weights and train for 30 epochs with a batch size of 128 and a learning rate of 0.001 .
    
\textbf{NPOS} \cite{npos2023iclr}:  For outlier synthesis, we sample 500 candidate boundary samples from the training set with Gaussian kernel covariance $\sigma^2 = 0.1$. Starting from epoch 1, we apply k-NN boundary selection (k=400) to obtain 300 final boundary samples. When computing the  OOD score,we set the temperature $\tau=0.1$. We use $\mathcal{M}_{R1}^{id}$ to initialize the weights and train for 30 epochs with a batch size of 128 and a learning rate of 0.001.

\textbf{DML} \cite{Zhang_2023_CVPR}: We use a cosine annealing learning rate schedule decaying from 1e-1 to 1e-6 and train two ResNet networks for 100 epochs  with a batch size of 128, where one model is trained with Center Loss and the other with Focal Loss.
\subsubsection{Data Aug for OOD Detection}

For all data-augmented training methods, we initialize the models with $\mathcal{M}_{R1}^{id}$ weights and train for 30 epochs using a batch size of 128.  Method-specific hyperparameter settings will be detailed subsequently:

\textbf{Mixup} \cite{iclr2018mixup}: We use $\alpha$=0.2 for Beta distribution. 

\textbf{RegMixup} \cite{pinto2022RegMixup}: We use $\alpha$=10 for Beta distribution. 

\textbf{RandAugment} \cite{cvprw2020Randaugment}: We use $n$=2, $m$=9, which indicates that two consecutive augmentation operations are applied per image, each performed at high intensity. 

\textbf{AugMix}  \cite{hendrycks2020augmix}: We employ Jensen-Shannon Divergence as a regularization term with a Beta distribution parameter of 12, while utilizing the following configuration: severity level 1 for mild augmentations, activation of all augmentation operations, 3 parallel augmentation branches, uniform mixing weights through a Dirichlet distribution, and automatic operation depth selection. 

\textbf{Cutmix} \cite{iccv2019cutmix}: We apply this augmentation with probability 0.5, while the cropping region's shape and size are determined by a $\beta$(1.0, 1.0) distribution. Models are trained with 64 batch size for 30 epochs.

\textbf{CutOut} \cite{devries2017improved}: We apply random 16×16 pixel square masking to one region per image.

\textbf{LightAug}: We convert images to 3-channel grayscale with 25\% probability and applies brightness/contrast enhancement with 50\% probability

\subsubsection{VLM-based OOD Detection}
For VLM-based OOD detection, method-specific hyperparameter settings will be detailed subsequently:

\textbf{MaxLogits} \cite{species22icml}: This method uses max logits and does not require hyperparameter tuning.

\textbf{MCM} \cite{ming2022delving},: We set the temperature $\tau=1$ in the softmax function to calculate the MCM score.

\textbf{GL-MCM} \cite{miyai2025zero}: We set the temperature $\tau=1$ in the softmax function and the weight of local MCM score $\lambda=1$ to calculate the GL-MCM score.

\textbf{NegLabel} \cite{jiang2024negative}: We select $M=1000$ negative labels with cosine similarities below the threshold $\eta=0.05$. We employ the NegLabel score in the sum-softmax form with temperature $\tau=1$, and apply a grouping strategy with $n_g=100$ groups.

\textbf{CLIPN}  \cite{wang2023clipn}: We use the official checkpoints for evaluation and set the temperature $\tau=1$.

\textbf{CoOp} \cite{zhou2022learning}: During training, the hyperparameters are set to be consistent with those used in Section~\ref{sec:Close-set Classification}. During testing, we set the temperature $\tau=100$ to calculate the MCM score.

\textbf{LoCoOp} \cite{miyai2023locoop}: During training, the hyperparameters are set to be consistent with those used in CoOp. For LoCoOp-specific settings, we set the weight of OOD regularization loss  $\lambda_{ood}=0.25$ and the number of extracted OOD regions $K=20$. During testing, we set the temperature $\tau=100$ and the weight of local MCM score $\lambda=1$ to calculate the GL-MCM score. 

\textbf{SCT} \cite{yu2024selfcalibratedtuningvisionlanguagemodels}:  The hyperparameters for SCT are set to be consistent with those used in LoCoOp.
 
\textbf{DPM}  \cite{2024ECCV}: During training, the hyperparameters are set to be consistent with those used in Section~\ref{sec:Close-set Classification}. During testing, we set the temperature $\tau=100$ and determine the visual modality affinity factor $\beta$ by searching over a range of candidates for each specific task and model. Note that when $\beta=0$, the DPM score reduces to the MCM score.

\subsection{Incremental Learning}
For incremental learning, we evaluat two categories of methods: conventional CNN-based methods and pre-trained ViT finetuning strategies. The experiments are implemented using PyTorch and PyCIL. 

For the CNN-based methods, we adopt ResNet18 as the backbone architecture. We use SGD with an initial learning rate of 0.1 and momentum of 0.9. The training epoch is set to 170 for all datasets with a batch size of 128. The learning rate undergoes a decay of 0.1 at 80 and 120 epochs. It must be noted that Finetune, \textbf{EWC} and \textbf{LwF} are exemplar-free methods, and we do not use any exemplar set for them. For other methods, we follow the benchmark setting to set the number of exemplars to 3780, with 20 samples for each class. We follow the original paper to set the algorithm-specific parameters, e.g., splitting 10\% exemplars from the exemplar set as validation for \textbf{BiC}, setting the temperature to 5 and using a 10 epochs warm-up for \textbf{DER}, using $\ell_2$ norm to normalize the fully-connected layers in \textbf{WA}. For \textbf{EWC}, the $\lambda$ parameter is determined via a grid search among $\{1, 10^1 , 10^2 , 10^3 , 10^4\}$, and we find $10^3$ leads to its best performance.

For ViT-based methods, we adopt ViT-B/16 as the backbone, which is pre-trained on ImageNet-21K. The initial learning rate is set to 0.01 and we only train the first session for 20 epochs in the first session adaptation methods, like \textbf{RanPAC}, and 20 epochs for later sessions in other methods. For \textbf{SSIAT} and \textbf{SLCA}, we only train 10 epochs in the incremental sessions and 5 epochs for the classifier alignment step. We employ the Adam optimizer with cosine annealing learning rate scheduling. For prompt-based method, like \textbf{Adam-prompt}, \textbf{L2P}, \textbf{DualPrompt}, and \textbf{CODA-Prompt}, we use the deep prompt version, which sets learnable prompt for each block. 

\section{Detailed results}

\subsection{Closed-set Classification}
Tab.~\ref{table:close} presents detailed closed-set classification results for various models on OES. Surprisingly, ResNet \cite{he2016deep} outperforms both ViT \cite{dosovitskiy2020vit} and CLIP \cite{radford2021learning}. This suggests that the downstream performance of ViT and CLIP depends heavily on pre-training. In remote sensing, the significant domain gap in pre-training data reduces their effectiveness, resulting in suboptimal performance. In contrast, the GeoRSCLIP \cite{zhang2024rs5m}, specifically pre-trained on remote sensing data, shows substantial performance improvements, highlighting the importance of effective pre-training.

\subsection{Zero-shot Classification}
Tab.~\ref{table:zero} presents the zero-shot classification performance of various CLIP architectures on  OES. RemoteCLIP \cite{remoteclip}, despite using remote sensing data in pre-training, has limited data diversity, resulting in poor performance. In contrast, GeoRSCLIP benefits from more diverse pre-training data, significantly enhancing its performance. Using the same ViT-B/32 architecture, GeoRSCLIP improves acc@1 by 24.86\% and acc@5 by 29.34\% compared to the original CLIP.

\subsection{OOD Detection \& Generalization}
Tab.~\ref{overview of ood} offers a detailed overview of the results on the OES dataset. It presents the OOD detection performance for each sub-dataset, reporting the AUROC for both Near-OOD and Far-OOD, in addition to the Top-1 ID accuracy. 
Tab.~\ref{clipood1},~\ref{clipood2},~\ref{clipood3} present the OOD detection performance of various methods under various CLIP architectures across different sub-datasets. Each method demonstrates substantial improvements across various settings when applied to the GeoRSCLIP underscoring the importance of pre-training. Tab.~\ref{table:singleood1},~\ref{table:singleood2},~\ref{singleood3} provide a detailed comparison of the OOD detection performance of various unimodal methods across different sub-datasets as referenced in the main text.

\subsection{Incremental Learning}
Tab.~\ref{table:cil-tradition} and Tab.~\ref{table:cil-ptm} respectively present a detailed comparison of class incremental learning performance of traditional and pre-trained model-based methods along with Joint and Finetune on benchmarks \textbf{Random}, \textbf{Coarse} and \textbf{Scale}.

\begin{table*}[h]
    \centering
\caption{Detailed description of dataset notations in the main paper.}
    \begin{tabular}{ccccc} 
        \hline
        Dataset &Introduction &\\
        \hline
        $\mathcal{D}_{R1}^{all}$
        &All the satellite RGB images and classes in sub-dataset 1 with 189 classes.\\
        
        $\mathcal{D}_{R1}^{id}$  
        &All images from the 94 in-distribution(ID) classes within the 189 classes in $\mathcal{D}_{R1}^{all}$.\\
        $\mathcal{D}_{R1}^{oode}$&
        Easy OOD split with 48 out-of-distribution(OOD) classes within the 189 classes in $\mathcal{D}_{R1}^{all}$.\\
        $\mathcal{D}_{R1}^{oodh}$&Hard OOD split with 47 OOD classes within the 189 classes in $\mathcal{D}_{R1}^{all}$.
        \\
        $\mathcal{D}_{R1}^{d}$&All the images in 50 classes within the 189 classes in $\mathcal{D}_{R1}^{all}$ used for domain incremental learning.\\

        $\mathcal{D}_{R2}^{all}$& All 65 classes satellite RGB images in Sub-dataset 2. All categories in $\mathcal{D}_{R2}^{all}$ already present in $\mathcal{D}_{R1}^{all}$.
        \\
        $\mathcal{D}_{R2}^{id}$ &  Contains all 43 ID class images from $\mathcal{D}_{R2}^{all}$'s 65 classes, with all categories existing in $\mathcal{D}_{R1}^{ID}$.
       \\
        $\mathcal{D}_{R2}^{ood}$ &  Contains all 22 OOD class images from $\mathcal{D}_{R2}^{all}$'s 65 classes, with all categories existing in $\mathcal{D}_{R1}^{OOD}$.
       \\

        $\mathcal{D}_{R3}^{all}$& All 137 classes aerial RGB images in Sub-dataset 3. All categories in $\mathcal{D}_{R3}^{all}$ already present in $\mathcal{D}_{R1}^{all}$.
        \\
        $\mathcal{D}_{R3}^{id}$ &  Contains all 71 ID class images from $\mathcal{D}_{R3}^{all}$'s 137 classes, with all categories existing in $\mathcal{D}_{R1}^{ID}$.
       \\
        $\mathcal{D}_{R3}^{ood}$ &  Contains all 66 OOD class images from $\mathcal{D}_{R3}^{all}$'s 65 classes, with all categories existing in $\mathcal{D}_{R1}^{OOD}$.
       \\
    
         $\mathcal{D}_{R3}^{d}$&All the images in 50 classes within the 137 classes in $\mathcal{D}_{R3}^{all}$ used for domain incremental learning.\\

        $\mathcal{D}_{R4}^{all}$& All 56 classes aerial RGB images in Sub-dataset 4. All categories in $\mathcal{D}_{R4}^{all}$ already present in $\mathcal{D}_{R1}^{all}$.
        \\
        $\mathcal{D}_{R4}^{id}$ &  Contains all 34 ID class images from $\mathcal{D}_{R4}^{all}$'s 56 classes, with all categories existing in $\mathcal{D}_{R1}^{ID}$.
       \\
        $\mathcal{D}_{R4}^{ood}$ &  Contains all 22 OOD class images from $\mathcal{D}_{R4}^{all}$'s 56 classes, with all categories existing in $\mathcal{D}_{R1}^{OOD}$.
       \\
    
         $\mathcal{D}_{R4}^{d}$&All the images in 50 classes within the 56 classes in $\mathcal{D}_{R4}^{all}$ used for domain incremental learning.\\

        $\mathcal{D}_{R5}^{all}$& All 62 classes aerial RGB images in Sub-dataset 5. All categories in $\mathcal{D}_{R5}^{all}$ already present in $\mathcal{D}_{R1}^{all}$.
        \\
        $\mathcal{D}_{R5}^{id}$ &  Contains all 36 ID class images from $\mathcal{D}_{R5}^{all}$'s 62 classes, with all categories existing in $\mathcal{D}_{R1}^{ID}$.
       \\
        $\mathcal{D}_{R5}^{ood}$ &  Contains all 26 OOD class images from $\mathcal{D}_{R5}^{all}$'s 62 classes, with all categories existing in $\mathcal{D}_{R1}^{OOD}$.
       \\
    
         $\mathcal{D}_{R4}^{d}$&All the images in 50 classes within the 62 classes in $\mathcal{D}_{R4}^{all}$ used for domain incremental learning.\\
        \hline
    \end{tabular}

        \label{table:notations}
\end{table*}

\begin{table*}[h]
    \centering
 \caption{Detailed statistics of sub-datasets}
    \begin{subtable}[h]{0.48\textwidth}
        \centering
        
  \setlength{\tabcolsep}{2pt}{
\begin{tabular}{ccccc} 
        \hline
        Datasets  & Images & Resolution
  & Size& Classes \\  
        \hline
         WHU-RS19 \cite{xia2010structural} & 45 
& 0.5m & 600 & 1\\  
         RESISC45 \cite{cheng2017remote} & 14841&0.2-30m& 600 & 31\\
        RSD46-WHU \cite{xiao2017high} & 11094 
& 0.5-2m & 256 & 23\\
         AID \cite{xia2017aid} & 2130 
& 3m & 600 & 7\\
         MillionAID \cite{long2021creating}& 3302& 0.5–153 m & 600 & 16\\
         MLRSNet \cite{qi2020mlrsnet} & 1548& 0.1-10m & 256 &4 \\

        RSI-CB256 \cite{li2020rsi} & 6873& 0.3-3m & 256 &17 \\

        BigEarthNet \cite{sumbul2019bigearthnet} & 355 
& 10m & 120 & 2\\
        fMoW \cite{christie2018functional} & 19017 
& 0.3m & - & 50\\
        TreeSatAI \cite{ahlswede2022treesatai} & 6297
& 10m & 304 & 14\\  
        FAIR1M \cite{sun2022fair1m} & 6925 
& 0.3-0.8m & 256 &20\\  

        FGSC-23 \cite{2021FGSC23AL} & 753 
& - &  & 6\\
        FGSCR-42 \cite{di2019public} & 2027 
& - & - & 6\\
        NaSC-TG2 \cite{zhou2021nasc} & 500 & 100m & 256 & 1\\
        \hline
    \end{tabular}}

            \caption{Sub-dataset 1}
         
    \end{subtable}
    \vspace{-10mm}
\renewcommand\arraystretch{1.2}
    \begin{subtable}[h]{0.48\textwidth}
        \centering
            \vspace{-18mm}
  \setlength{\tabcolsep}{0.50mm}{
    \begin{tabular}{ccccc} 
        \hline
        Datasets  & Images & Resolution
  & Size& Classes \\  
        \hline
         FAIR1M \cite{sun2022fair1m} & 1774& 0.3-0.8m & 256 & 4\\  
         MillionAID \cite{long2021creating} & 1088& 0.5-153m & 600 & 5\\
         MLRSNet \cite{qi2020mlrsnet} & 1500& 0.1-10m & 256 & 3\\
          Optimal-31 \cite{wang2018scene} & 480& - & 256 & 8\\
        PatternNet
\cite{zhou2018patternnet}& 11998& 0.06-4.7m & 256 & 24\\
        UCMLandUse \cite{yang2010bag} & 100& 0.3m & 256& 1\\
        RSI-CB128 \cite{li2020rsi} & 7337& 0.3-3m & 128 & 16\\
                 RSD46-WHU \cite{xiao2017high} & 2000& 0.5-2m & 256 & 4\\
                         \hline
    \end{tabular}}
            \caption{Sub-dataset 2}
                 
    \end{subtable}
\renewcommand\arraystretch{1.0}

\begin{subtable}[h]{0.48\textwidth}
        \centering
        \vspace{+10mm}
  \setlength{\tabcolsep}{0.80mm}{
    \begin{tabular}{ccccc} 
        \hline
        Datasets  & Images & Resolution
  & Size& Classes \\  
        \hline
         fMoW \cite{christie2018functional} & 19153& 0.3m& - & 50\\  
        SmokeRS \cite{ba2019smokenet}& 500& 1000m & 256 & 1\\
                 MRSSC2.0 \cite{liu2022remote}& 2000& 100m& 256& 5\\
                         \hline
    \end{tabular}}
                \caption{Sub-dataset 4}
         
    \end{subtable}
\renewcommand\arraystretch{1.2}
        \begin{subtable}[h]{0.48\textwidth}
        \centering
        \vspace{-4mm}
\setlength{\tabcolsep}{0.80mm}{
    \begin{tabular}{ccccc} 
        \hline
        Datasets  & Images & Resolution
  & Size& Classes \\  
        \hline
         fMoW \cite{christie2018functional} & 19142& 0.3m & - & 50\\  
         BigEarthNet \cite{sumbul2019bigearthnet} & 355 
& 10m & 120 & 2\\
                 VisDrone\cite{zhu2021detection}& 461& - & 256& 3\\
                 MRSSC2.0\cite{liu2022remote}& 2416&100m & 128& 5\\
                 NaSC-TG2\cite{zhou2021nasc}& 1000& 100m&256 & 2\\
                         \hline
    \end{tabular}}
            \caption{Sub-dataset 5}
          
    \end{subtable}

    \label{tab:subdatasets}
\end{table*}

\begin{table*}[h]
    \centering
    \caption{License/Usage for the component datasets of OES(Edu/Res/Com = Education/Research/Commercial).}
    \setlength{\tabcolsep}{1pt} 
    \begin{tabular}{@{}cc|cc|cc@{}} 
        \hline
        Dataset & Licence/Usage & Dataset & Licence/Usage & Dataset & Licence/Usage \\
        \hline
        
        WHU-RS19 \cite{xia2010structural} & CC BY 4.0 & RSD46-WHU \cite{xiao2017high} & Edu/Res/Com & MillionAID \cite{long2021creating} & CC BY-NC-ND 4.0 \\
        
        RESISC45 \cite{cheng2017remote} & CC BY-NC 4.0 & AID \cite{xia2017aid} & Research only & MLRSNet \cite{qi2020mlrsnet} & CC BY 4.0 \\
        
        RSI-CB256 \cite{li2020rsi} & Academic & BigEarthNet \cite{sumbul2019bigearthnet} & CDLA-Permissive-1.0 & fMoW \cite{christie2018functional} & Academic \\

        TreeSatAI \cite{ahlswede2022treesatai} & CC BY-SA 4.0 & FAIR1M \cite{sun2022fair1m} & CC BY-NC-SA 3.0 & FGSC-23 \cite{2021FGSC23AL} & Not found \\
        
        FGSCR-42 \cite{di2019public} & Not found & NaSC-TG2 \cite{zhou2021nasc} & CC BY 4.0 & Optimal-31 \cite{wang2018scene} & Not found \\
        
        RSI-CB128 \cite{li2020rsi} & Academic & PatternNet \cite{zhou2018patternnet} & Academic & UCMLandUse \cite{yang2010bag} & Public domain \\
        
        SmokeRS \cite{ba2019smokenet} & Res/Edu & MRSSC2.0 \cite{liu2022remote} & Education & VisDrone \cite{zhu2021detection} & CC-BY-NC-SA 3.0 \\
        \hline
    \end{tabular}
     \label{lic}
\end{table*}

\begin{table*}[h]
    \centering
    \setlength{\tabcolsep}{0.5mm}
    \caption{Detailed statistics of coarse and fine classes. For fine classes, the class in bold represents OOD-Easy category, class in italic represents the OOD-Hard category, and the remaining represents the ID category.}
    \begin{tabular}{cc} 
        \hline
        Coarse Class & Fine Class \\
        \hline
        \multirow{6}{*}{Vegetation}
        & Shrubwood, Sparse shrub land, Mangrove, Artificial grassland, \color{red}\textbf{Sapling}\color{black}, \\
        & \color{red}\textbf{River protection forest}\color{black}, \color{red}\textbf{Orchard}\color{black}, Chaparral, Meadow, Artificial dense forest land, \\
        & Artificial sparse forest land, Natural sparse forest land, Natural dense forest land, \\
        & \color{blue}\textit{Acer pseudoplatanus}\color{black}, Abies alba, \color{blue}\textit{Fagus sylvatica}\color{black}, Larix decidua, Picea abies, \\
        & Pinus nigra, Pinus strobus, Pinus sylvestris, \color{blue}\textit{Populus spec}\color{black}, Pseudotsuga menziesii, \\
        & \color{blue}\textit{Quercus robur}\color{black}, \color{blue}\textit{Quercus rubra}\color{black}, \color{blue}\textit{Tilia spec}\color{black}, \color{blue}\textit{Quercus petraea}\color{black} \\
        \hline
        \multirow{3}{*}{Agriculture}  
        & Crop field, Aquaculture, Barn, Rectangular farmland, Circular farmland \\
\hline
    \end{tabular}
    \label{table:coarse}
\end{table*}

\begin{table*}[h]
    \centering
 \setlength{\tabcolsep}{3mm}
        \caption{Detailed statistics of scene classes. The scales of scenes and object categories are respectively scored by Qwen-VL, and an incremental process is arranged in descending order of scale, evenly distributed across 10 tasks.}
    \begin{tabular}{ccccc} 
        \hline
        Task &Class name\\
        \hline
        \multirow{5}{*}{Task 0}
        &\multirow{1}{*}{Golf course (100), Railway (96.20), Bridge (94.90), Coast (94.62), Road (94.58),}\\ 
        &\multirow{1}{*}{Steelsmelter (94.54), Center (94.46), Freeway (94.39), Playground (93.77),}\\ 
        &\multirow{1}{*}{Tennis court (93.61), Railway station (93.56), Nuclear powerplant (93.47), River (93.07),}\\
        &\multirow{1}{*}{Palace (92.99), Natural dense forest land (92.96), Terrace (92.86), Stadium (92.76),}\\
        &\multirow{1}{*}{Thermal power station (92.71), School (92.65)}\\
        \hline
        \multirow{4}{*}{Task 1}
        &\multirow{1}{*}{Artificial dense forest land (92.54), Avenue (92.45), Quarry (92.43), Lake (92.31),}\\ 
        &\multirow{1}{*}{Shopping mall (92.29), Port (92.26), Construction site (92.06), Artificial grassland,}\\
\hline
    \end{tabular}

        \label{table:scene}
\end{table*}

\begin{table*}[h]
    \centering
 \setlength{\tabcolsep}{3mm}
        \caption{Detailed statistics of scene and objecct classes. The scales of scenes and object categories are respectively scored by Qwen-VL, and an incremental process is arranged in descending order of scale, evenly distributed across 10 tasks.}
    \begin{tabular}{ccccc} 
        \hline
        Task &Class name\\
        \hline
        \multirow{3}{*}{Task 8}
        &\multirow{1}{*}{Container (100), Megayacht (97.34), Pipeline (96.10), Boeing747 (94.44),}\\ 
        &\multirow{1}{*}{Cargo truck (93.53), Van (93.13), Boeing777 (93.06), Wind turbine (92.92), Bus (92.52),}\\ 
        &\multirow{1}{*}{Excavator (91.88), Storage tank (91.86), Boeing787 (91.71), Warship (91.08),}\\
        &\multirow{1}{*}{Trailer (90.58), Truck tractor (90.47), Bulk carrier (90.35), Tank ship (90.27),}\\
        &\multirow{1}{*}{Container ship (94.05), Dump truck (90.65)}\\
        \hline
        \multirow{3}{*}{Task 9}
        &\multirow{1}{*}{Car carrier (89.88), A350 (89.49), Sand carrier (89.48), Boeing737 (88.09), Tractor (85.19),}\\ 
        &\multirow{1}{*}{Civil yacht (84.40), Transmission tower (84.89), A330 (84.17), Towing vessel (84.07),}\\ 
        &\multirow{1}{*}{Liquefied gas ship (84.07), Passenger ship (82.63), Barge (82.38), A321 (81.30),}\\
        &\multirow{1}{*}{Fishing boat (80.03), Small car (75.92), C919 (68.99), ARJ21 (66.63), A220 (55.38)}\\ 
        \hline
    \end{tabular}

        \label{table:object}
\end{table*}

\renewcommand\arraystretch{1.1}
\begin{table*}[h]
    \centering
    \caption{Detailed closed-set performance. FT: Finetune, TPT: Textual Prompt Tuning, VT: Visual Tuning}
    \setlength{\tabcolsep}{0.5mm} 
    \begin{tabular}{cc *{10}{c}}
        \toprule
        \multirow{2}{*}{Model} & \multirow{2}{*}{Training} & \multicolumn{2}{c}{Satellite RGB} & \multicolumn{2}{c}{Aerial RGB} & \multicolumn{2}{c}{MSRGB} & \multicolumn{2}{c}{IR} & \multicolumn{2}{c}{Average} \\
        \cmidrule(lr){3-4} \cmidrule(lr){5-6} \cmidrule(lr){7-8} \cmidrule(lr){9-10} \cmidrule(lr){11-12}
        & & acc@1 & acc@5 & acc@1 & acc@5 & acc@1 & acc@5 & acc@1 & acc@5 & acc@1 & acc@5 \\
        \midrule
        ResNet-18 & FT & 84.52 & 96.74 & 78.90 & 92.37 & 70.03 & 90.38 & 68.84 & 84.97 & 75.57 & 92.33 \\
        ResNet-34 & FT & 86.06 & 97.23 & 79.03 & 93.00 & 72.62 & 90.87 & 70.29 & 90.64 & 77.00 & 92.94 \\
        ResNet-50 & FT & 87.58 & 97.70 & 80.52 & 93.45 & 75.94 & 92.22 & 74.19 & 91.90 & 79.56 & 93.82 \\
        ResNet-101 & FT & 87.87 & 97.79 & 81.06 & 93.95 & 77.20 & 93.12 & 75.17 & 92.56 & 80.33 & 94.36 \\
        ResNet-152 & FT & 87.91 & 97.84 & 81.38 & 93.80 & 76.80 & 93.28 & 75.68 & 92.48 & 80.44 & 94.35 \\
        ViT-B/32 & FT & 84.44 & 96.93 & 70.15 & 90.70 & 70.93 & 91.05 & 67.52 & 89.54 & 73.26 & 92.06 \\
\bottomrule
    \end{tabular}
    \label{table:close}
\end{table*}

\begin{table*}[h]
    \centering
    \caption{Detailed zero-shot classification performance on OES.}
    
    \setlength{\tabcolsep}{0.5mm} 
    \begin{tabular}{cc *{10}{c}}
        \toprule
        \multirow{2}{*}{Model} & \multirow{2}{*}{Image Encoder} & \multicolumn{2}{c}{Satellite RGB} & \multicolumn{2}{c}{Aerial RGB} & \multicolumn{2}{c}{MSRGB} & \multicolumn{2}{c}{IR} & \multicolumn{2}{c}{Average} \\
        \cmidrule(lr){3-4} \cmidrule(lr){5-6} \cmidrule(lr){7-8} \cmidrule(lr){9-10} \cmidrule(lr){11-12}
        & & acc@1 & acc@5 & acc@1 & acc@5 & acc@1 & acc@5 & acc@1 & acc@5 & acc@1 & acc@5 \\
        \midrule
        CLIP & ResNet-50 & 13.21 & 29.69 & 39.98 & 72.30 & 15.65 & 39.88 & 7.65 & 27.86 & 19.12 & 42.43 \\
        CLIP & ViT-B/16 & 17.83 & 39.74 & 51.67 & 84.14 & 23.36 & 48.97 & 14.09 & 37.08 & 26.74 & 52.48 \\
        CLIP & ViT-B/32 & 19.79 & 45.16 & 59.47 & 89.02 & 24.06 & 51.06 & 16.73 & 39.88 & 30.01 & 56.28 \\
        RemoteCLIP & ResNet-50 & 12.11 & 30.85 & 10.96 & 33.69 & 10.90 & 33.27 & 9.83 & 29.73 & 10.95 & 31.89 \\
        RemoteCLIP & ViT-B/32 & 19.09 & 45.32 & 23.66 & 55.33 & 20.98 & 49.60 & 15.45 & 36.74 & 19.80 & 46.75 \\
        GeoRSCLIP & ViT-B/32 & 31.69 & 64.22 & 59.59 & 90.61 & 53.84 & 85.81 & 30.80 & 63.70 & 43.98 & 76.09 \\
        \bottomrule
    \end{tabular}
    \label{table:zero}
\end{table*}

\renewcommand\arraystretch{1.1}

\begin{table*}[h]
    \centering
    \caption{Detailed overview of the OOD detection performance on each sub-dataset. 'Near' represents the average AUROC for Near-OOD datasets, 'Far' indicates the average AUROC for Far-OOD datasets, and 'Acc' denotes the Top-1 ID classification accuracy.}
    \setlength{\tabcolsep}{0.8mm} 
    \small 
    \begin{tabular}{c|ccc|ccc|ccc|ccc|ccc}
        \toprule

        \multicolumn{1}{c}{Method} & \multicolumn{3}{c}{Standard} & \multicolumn{3}{c}{Resampling Bias} & \multicolumn{3}{c}{Modal-shift (Aerial)} & \multicolumn{3}{c}{Modal-shift (MS)} & \multicolumn{3}{c}{Modal-shift (IR)} \\
\multicolumn{1}{c}{} & \multicolumn{1}{c}{Near} & \multicolumn{1}{c}{Far} & \multicolumn{1}{c}{Acc} & \multicolumn{1}{c}{Near} & \multicolumn{1}{c}{Far} & \multicolumn{1}{c}{Acc} & \multicolumn{1}{c}{Near} & \multicolumn{1}{c}{Far} & \multicolumn{1}{c}{Acc} & \multicolumn{1}{c}{Near} & \multicolumn{1}{c}{Far} & \multicolumn{1}{c}{Acc} & \multicolumn{1}{c}{Near} & \multicolumn{1}{c}{Far} & \multicolumn{1}{c}{Acc} \\
\hline
     
\hline
\multicolumn{16}{l}{\textbf{Post-hoc Methods}} \\
\hline

MSP \cite{hendrycks17baseline} 
& 87.90 & 95.25 & 92.01 
& 66.00 & 82.34 & 47.73 
& 53.32 & 60.00 & 18.41 
& 64.92 & 64.94 & 46.59 
& 58.64& 62.09 & 31.06 \\

ODIN \cite{odin18iclr} & 86.29 & 95.82 & 92.01 & 63.14 & 78.93 & 47.73 & 52.97 & 56.12 & 18.41 & 65.67 & 60.44 & 46.59 & 59.42 & 67.49 & 31.06 \\
 
MDS \cite{mahananobis18nips} 
& 89.71 & 97.66 & 92.01 
& 54.93 & 64.28 & 47.73
& 48.05 & 44.57 & 18.41 
& 58.44 & 69.67 & 46.59 
& 55.20 & 36.51 & 31.06 \\

ReAct \cite{sun2021react} & 88.89 & 96.80 & 92.01 & 62.90 & 79.98 & 47.73 & 52.60 & 59.74 & 18.41 & 64.21 & 63.39 & 46.59 & 59.15 & 63.39 & 31.06 \\
\bottomrule

\end{tabular}

\label{overview of ood}
\end{table*}

 \renewcommand\arraystretch{1.1}
\begin{table*}[h]
    \centering
         \caption{OOD detection performance of VLM-based methods with various CLIP architectures on sub-dataset 1. For CoOp, LoCoOp, SCT, and DPM with GeoRSCLIP, we report the results with softmax, while for the remaining architectures, we report the results without softmax. AUR stands for AUROC and FPR stands for FPR95.}
    \setlength{\tabcolsep}{2.8mm} 
    \begin{tabular}{ccc|cc|cc|cc|c}
        \toprule
        \multirow{2}{*}{Method} & \multicolumn{2}{c}{OOD-Easy} & \multicolumn{2}{c}{OOD-Hard} & \multicolumn{2}{c}{SUN} & \multicolumn{2}{c}{Average} & \multirow{2}{*}{acc@1} \\
        \cmidrule(lr){2-3} \cmidrule(lr){4-5} \cmidrule(lr){6-7} \cmidrule(lr){8-9}
        & AUR$\uparrow$ & FPR$\downarrow$ & AUR$\uparrow$ & FPR$\downarrow$ & AUR$\uparrow$ & FPR$\downarrow$ & AUR$\uparrow$ & FPR$\downarrow$ & \\
        \midrule
        
        \multicolumn{10}{l}{\textbf{Image Encoder: CLIP ViT-B/16}} \\
        \midrule
        MaxLogits \cite{species22icml} & 67.28 & 79.29 & 54.38 & 63.09 & 53.18 & 88.87 & 58.28 & 77.08 & 27.73 \\
        MCM \cite{ming2022delving} & 53.48 & 93.35 & 53.49 & 93.53 & 43.22 & 98.92 & 50.06 & 95.27 & 27.73 \\
        GL-MCM \cite{miyai2025zero} & 52.99 & 93.41 & 54.26 & 93.61 & 43.94 & 96.89 & 50.40 & 94.64 & 27.73 \\
        CLIPN \cite{wang2023clipn} & 53.20 & 92.04 & 49.42 & 93.88 & 56.29 & 88.70 & 54.02 & 91.54 & 28.69 \\
        NegLabel \cite{jiang2024negative} & 65.30 & 91.87 & 51.19 & 91.87 & 58.40 & 91.87 & 58.29 & 91.87 & 25.75 \\
\bottomrule
    \end{tabular}

    \label{clipood1}
\end{table*}

\renewcommand\arraystretch{1.1}
\begin{table*}[h]
    \caption{Covariate-shift OOD detection performance of VLM-based methods with various CLIP architectures on sub-dataset 2. For CoOp, LoCoOp, SCT, and DPM with GeoRSCLIP, we report the results with softmax, while for the remaining architectures, we report the results without softmax. AUR stands for AUROC and FPR stands for FPR95.}
    \centering
    \small
    \setlength{\tabcolsep}{2.0mm}
    \begin{tabular}{ccc|cc|cc|cc|cc|c}
        \toprule
        \multirow{2}{*}{Method} & \multicolumn{2}{c}{OOD-Bias} & \multicolumn{2}{c}{OOD-Easy} & \multicolumn{2}{c}{OOD-Hard} & \multicolumn{2}{c}{SUN} & \multicolumn{2}{c}{Average} & \multirow{2}{*}{acc@1} \\
        \cmidrule(lr){2-3} \cmidrule(lr){4-5} \cmidrule(lr){6-7} \cmidrule(lr){8-9} \cmidrule(lr){10-11}
        & AUR$\uparrow$ & FPR$\downarrow$ & AUR$\uparrow$ & FPR$\downarrow$ & AUR$\uparrow$ & FPR$\downarrow$ & AUR$\uparrow$ & FPR$\downarrow$ & AUR$\uparrow$ & FPR$\downarrow$ & \\
        \midrule
        
        \multicolumn{12}{l}{\textbf{Image Encoder: CLIP ViT-B/16}} \\
        \midrule
        MaxLogits \cite{species22icml} & 58.21 & 86.76 & 61.55 & 86.54 & 47.42 & 96.21 & 46.63 & 93.15 & 53.45 & 90.67 & 36.09 \\
        MCM \cite{ming2022delving} & 55.75 & 93.53 & 53.05 & 95.69 & 53.02 & 96.29 & 43.29 & 99.54 & 51.28 & 96.26 & 36.09 \\
        GL-MCM \cite{miyai2025zero} & 56.37 & 95.10 & 56.12 & 96.22 & 57.10 & 96.89 & 47.38 & 98.82 & 54.24 & 96.76 & 36.09 \\
        NegLabel \cite{jiang2024negative} & 59.74 & 91.87 & 65.30 & 91.87 & 51.19 & 91.87 & 58.40 & 91.87 & 58.66 & 91.87 & 25.75 \\
        CLIPN \cite{wang2023clipn} & 57.69 & 92.23 & 48.96 & 92.24 & 41.88 & 94.12 & 48.90 & 89.12 & 49.36 & 91.93 & 38.38 \\
\bottomrule

    \end{tabular}
    \label{clipood2}
\end{table*}

\begin{table*}[h]
    \centering
    \small
    \caption{Covariate-shift OOD detection performance of VLM-based methods with various CLIP architectures on sub-dataset 3,4,5. For CoOp, LoCoOp, SCT, and DPM with GeoRSCLIP, we report the results with softmax, while for the remaining architectures, we report the results without softmax. AUR stands for AUROC and FPR stands for FPR95.}
    \setlength{\tabcolsep}{1.2mm}
    \begin{tabular}{c|ccc|ccc|ccc}
        \toprule
        \multirow{2}{*}{Method} & \multicolumn{3}{c}{Modal-shift(Aerial)} & \multicolumn{3}{c}{Modal-shift(MSRGB)} & \multicolumn{3}{c}{Modal-shift(IR)} \\
        \cmidrule(r){2-4} \cmidrule(lr){5-7} \cmidrule(l){8-10}  

        & OOD-Bias & SUN  & acc@1 & OOD-Bias & SUN & acc@1 & OOD-Bias  & SUN  & acc@1 \\
        \hline
        
        \multicolumn{9}{l}{\textbf{Image Encoder: CLIP ViT-B/16}} \\
        \hline
        MaxLogits \cite{species22icml} & 71.23/75.27 & 69.96/78.50 & 63.41 & 55.94/94.39 & 47.02/93.53 & 24.88 & 51.18/95.23 & 36.86/94.02 & 15.94 \\
        MCM \cite{ming2022delving}& 64.95/87.38 & 64.21/87.74 & 63.41 & 51.74/94.46 & 40.56/99.20 & 24.88 & 49.43/95.16 & 34.81/99.58 & 15.94 \\
        GL-MCM \cite{miyai2025zero}& 65.20/88.22 & 63.34/87.58 & 63.41 & 52.57/95.22 & 39.59/98.34 & 24.88 & 51.39/95.41 & 33.90/99.07 & 15.94 \\
        CLIPN \cite{wang2023clipn}& 66.61/84.97 & 55.86/90.71 & 62.16 & 45.13/99.79 & 66.30/87.02 & 28.54 & 44.77/99.06 & 75.79/80.00 & 21.13 \\
        NegLabel \cite{jiang2024negative}& 57.61/91.87 & 58.40/91.87 & 61.09 & 57.39/91.87 & 58.40/91.87 & 25.75 & 71.49/57.29 & 58.40/91.87 & 14.55 \\
\bottomrule
    \end{tabular}

    \label{clipood3}
\end{table*}

\begin{table*}[h]
    \centering
    \small
        \caption{OOD detection performance of unimodal methods with ResNet-50 on sub-dataset 1.}
    \setlength{\tabcolsep}{1.5mm}
    \begin{tabular}{c|ccc|ccc|ccc}
        \toprule
        \multirow{2}{*}{Method} & \multicolumn{3}{c}{OOD-Easy} & \multicolumn{3}{c}{OOD-Hard} & \multicolumn{3}{c}{SUN} \\
        \cmidrule(r){2-4} \cmidrule(lr){5-7} \cmidrule(l){8-10}  
        & AUROC$\uparrow$ & FPR95$\downarrow$ & acc@1 & AUROC$\uparrow$ & FPR95$\downarrow$ & acc@1 & AUROC$\uparrow$ & FPR95$\downarrow$ & acc@1 \\
        \hline
        
        \multicolumn{10}{l}{\textbf{Post-hoc Methods}} \\
        \hline
 
OpenMax \cite{openmax16cvpr} &  85.52 & 35.28 & 82.71 & 47.46 & 84.40 & 19.56&84.21 & 34.10 &92.01\\
MSP \cite{hendrycks17baseline} & 89.55 & 38.32 & 86.25 & 49.37 & 95.25 & 18.21&90.35 & 35.30&92.01 \\ 
ODIN \cite{odin18iclr} &89.61 & 53.59 & 82.96 & 74.37 & 95.82 & 17.83 &89.46 & 48.60 &92.01\\
MDS \cite{mahananobis18nips}&94.28 & 28.71 & 85.14 & 51.46 & 97.66 & 8.17 &92.36 & 29.45 &92.01\\
GradNorm \cite{huang2021importance}&59.68 & 94.88 & 60.74 & 93.20 & 67.46 & 93.32 &62.63 & 93.80 &92.01\\
\bottomrule
    \end{tabular}

    \label{table:singleood1}
\end{table*}

\setlength{\tabcolsep}{1.6mm}

\begin{table*}[h]
    \centering
    \small
    \caption{ OOD detection of unimodal methods with ResNet-50 on sub-dataset 2.}
    \setlength{\tabcolsep}{1.2mm}
\begin{tabular}{c|ccccc cc c|ccc}
\toprule 

\multirow{2}{*}{Method} & \multicolumn{2}{c}{OOD-Bias} & \multicolumn{2}{c}{OOD-Easy} & \multicolumn{2}{c}{OOD-Hard} & \multicolumn{2}{c}{SUN} & \multicolumn{2}{c}{Average} & \multirow{2}{*}{acc@1} \\
\cmidrule{2-3} \cmidrule{4-5} \cmidrule{6-7} \cmidrule{8-9} \cmidrule{10-11}
 & AUR $\uparrow$ & FPR $\downarrow$ & AUR $\uparrow$ & FPR $\downarrow$ & AUR $\uparrow$ & FPR $\downarrow$ & AUR $\uparrow$ & FPR $\downarrow$ & AUR $\uparrow$ & FPR $\downarrow$ \\
\hline 
\multicolumn{12}{l}{\textbf{Post-hoc Methods}} \\
\hline 

OpenMax \cite{openmax16cvpr} & 54.62 & 82.73 & 55.89 & 80.94 & 54.54 & 85.25 & 47.30 & 66.28 & 53.09 & 78.80 & 47.73 \\
MSP \cite{hendrycks17baseline} & 65.27 & 82.05 & 69.64 & 80.79 & 63.08 & 86.24 & 82.34 & 57.78 & 70.08 & 76.72 & 47.73 \\ 
ODIN \cite{odin18iclr} & 65.97 & 84.08 & 69.51 & 81.66 & 53.93 & 91.05 & 78.93 & 69.48 & 67.09 & 81.57 & 47.73 \\
MDS \cite{mahananobis18nips} & 60.87 & 74.60 & 63.49 & 78.43 & 40.42 & 90.10 & 64.28 & 60.02 & 57.27 & 75.79 & 47.73 \\
GradNorm \cite{huang2021importance} & 54.87 & 96.62 & 53.88 & 96.17 & 54.20 & 94.73 & 62.16 & 94.79 & 56.28 & 95.58 & 47.73 \\
\bottomrule
\end{tabular}
\end{table*}

\definecolor{Ocean}{RGB}{200,200,200}
\renewcommand\arraystretch{1.1}

\begin{table*}[h]
    \centering
    \small
    \caption{Covariate-shift OOD detection of unimodal methods with ResNet-50 on sub-dataset 3,4,5.}
    \label{table:singleood2}
    \setlength{\tabcolsep}{1.2mm}
    \begin{tabular}{c|ccc|ccc|ccc}
        \toprule
        \multirow{2}{*}{Method} & \multicolumn{3}{c}{Modal-shift(Aerial)} & \multicolumn{3}{c}{Modal-shift(MSRGB)} & \multicolumn{3}{c}{Modal-shift(IR)} \\
        \cmidrule(r){2-4} \cmidrule(lr){5-7} \cmidrule(l){8-10}
        & OOD-Bias & SUN & acc@1 & OOD-Bias & SUN & acc@1 & OOD-Bias & SUN & acc@1 \\
        \hline
        
        \multicolumn{9}{l}{\textbf{Post-hoc Methods}} \\
        \hline
        OpenMax \cite{openmax16cvpr} & 48.14/92.49 & 39.71/93.73 & 18.41 & 50.98/82.82 & 52.45/83.18 & 46.59 & 49.98/87.16 & 44.10/88.77 & 31.06 \\
        MSP \cite{hendrycks17baseline} & 53.32/92.04 & 60.00/89.24 & 18.41 & 64.92/77.56 & 64.94/77.89 & 46.59 & 58.64/83.52 & 62.09/84.23 & 31.06 \\
        ODIN \cite{odin18iclr} & 52.97/92.93 & 56.12/92.00 & 18.41 & 65.67/78.19 & 60.44/85.01 & 46.59 & 59.42/82.77 & 67.49/82.42 & 31.06 \\
        MDS \cite{mahananobis18nips} & 48.05/93.82 & 44.57/92.44 & 18.41 & 48.05/93.82 & 44.57/92.44 & 46.59 & 55.20/85.61 & 36.51/92.13 & 31.06 \\
        GradNorm \cite{huang2021importance} & 52.61/94.26 & 63.26/87.64 & 18.41 & 59.71/94.04 & 53.64/94.11 & 46.59 & 55.89/91.87 & 46.42/97.00 & 31.06 \\
\bottomrule
    \end{tabular}
    \label{singleood3}
\end{table*}

\renewcommand\arraystretch{1.0}
\begin{table*}
\centering

\caption{ CIL performance of traditional methods with ResNet-18 on benchmarks \textbf{Random}, \textbf{Coarse}, and \textbf{Scale}.}
\setlength{\tabcolsep}{1.5mm}{
\begin{tabular}{cccccccccccc} 
\hline 
Method & 0 & 1 & 2 & 3 & 4 & 5 & 6 & 7 & 8 & 9 \textbf{(Last)} & \textbf{Avg} \\
\hline 
\multicolumn{12}{c}{Benchmark: \textbf{Random}} \\
\hline 
Joint & 88.50 & 84.89 & 85.59 & 82.18 & 81.68 & 82.17 & 80.72 & 80.51 & 81.63 & 80.37 & 82.82 \\
Finetune & 88.50 & 41.93 & 28.64 & 19.81 & 19.02 & 15.09 & 12.40 & 12.48 & 10.30 & 8.49 & 25.68 \\
\hline 
LwF \cite{li2017learning}  & 88.15 & 65.55 & 56.02 & 46.96 & 43.04 & 38.71 & 33..67 & 32.16 & 28.67 & 25.26 & 45.82 \\
EWC \cite{kirkpatrick2017overcoming}  & 87.28 & 50.02 & 37.28 & 28.07 & 23.47 & 21.86 & 15.67 & 15.94 & 14.87 & 12.03 & 30.65 \\
GEM \cite{lopez2017gradient}  & 91.79 & 45.55 & 33.45 & 23.29 & 20.77 & 17.78 & 13.80 & 14.44 & 13.49 & 10.95 & 28.53 \\
iCaRL \cite{rebuffi2017icarl} & 86.99 & 81.42 & 78.66 & 70.65 & 67.14 & 63.60 & 60.18 & 57.61 & 57.83 & 53.64 & 67.77 \\
\hline 
\end{tabular}}

\label{table:cil-tradition}
\end{table*}

\begin{table*}
\centering
\caption{ CIL performance of pre-trained model based methods with ViT-B/16 on benchmarks \textbf{Random}, \textbf{Coarse}, and \textbf{Scale}.}
\renewcommand\arraystretch{1.1}
\setlength{\tabcolsep}{1.5mm}{
\begin{tabular}{cccccccccccc} 

\hline 
Method & 0 & 1 & 2 & 3 & 4 & 5 & 6 & 7 & 8 & 9 \textbf{(Last)} & \textbf{Avg} \\
\hline 
\multicolumn{12}{c}{Benchmark: \textbf{Random}} \\
\hline 
Joint &95.46 & 91.91 & 91.55 & 90.04 & 88.37 & 88.69 & 87.60 & 86.50 & 85.05 & 84.75 & 88.99\\
Finetune &95.46 & 71.04 & 59.48 & 58.72 & 49.08 & 42.27 & 39.20 & 39.03 & 34.20 & 23.51 & 51.20\\
\hline 
Adam-adapter \cite{zhou2024revisiting} & 90.47 & 79.43 & 77.31 & 72.05 & 67.76 & 67.93 & 67.17 & 64.88 & 62.63 & 62.31 & 71.19 \\
Adam-ssf \cite{zhou2024revisiting} & 90.92 & 79.40 & 77.24 & 72.55 & 67.73 & 67.34 & 66.22 & 63.95 & 61.77 & 61.52 & 70.86 \\
Adam-prompt \cite{zhou2024revisiting} & 90.92 & 80.67 & 77.84 & 73.73 & 69.42 & 69.43 & 68.62 & 66.16 & 64.01 & 63.69 & 72.45 \\
L2P \cite{wang2022learning} & 90.27 & 77.89 & 70.22 & 68.00 & 62.80 & 63.40 & 62.16 & 59.25 & 57.00 & 55.79 & 66.68 \\
\hline 
\end{tabular}}

\label{table:cil-ptm}
\end{table*}




\end{appendices}

\clearpage
\clearpage
\bibliography{sn-bibliography}

\end{document}